\documentclass[runningheads]{llncs}
\usepackage{graphicx}

\usepackage[dvipsnames]{xcolor}
\usepackage{tikz}
\usepackage{pgfplots}
\usepackage{pgfplotstable}
\pgfplotsset{compat=1.17}
\usetikzlibrary{pgfplots.groupplots}
\usetikzlibrary{calc,arrows}
\usepackage{comment}
\usepackage{amsmath,amssymb} 
\usepackage{color}
\usepackage{colortbl}
\usepackage{wrapfig}
\usepackage[export]{adjustbox}

\usepackage[font=small,labelfont=bf,labelsep=period]{caption}

\pgfplotsset{ every non boxed x axis/.append style={x axis line style=-},
	every non boxed y axis/.append style={y axis line style=-}}

\usepackage{hyperref}
\usepackage{url}
\usepackage{booktabs} 
\usepackage{multirow}
\usepackage{makecell}

\hypersetup{
	colorlinks,
	linkcolor={red!80!black},
	citecolor={blue!80!black},
	urlcolor={blue!80!black}
}

\definecolor{LightCyan}{rgb}{0.88,1,1}

\newcommand{\C}[1]{C^{\text{#1}}}
\newcommand{\etal}{\textit{et al.}}

\newcommand{\mj}{$\mathcal{J}$}
\newcommand{\mf}{$\mathcal{F}$}
\newcommand{\mjf}{$\mathcal{J}\&\mathcal{F}$}

\newcommand{\mjs}{$\mathcal{J}_s$}
\newcommand{\mfs}{$\mathcal{F}_s$}
\newcommand{\mju}{$\mathcal{J}_u$}
\newcommand{\mfu}{$\mathcal{F}_u$}
\newcommand{\mg}{$\mathcal{G}$}

\newcommand{\beginsupplement}{
	\setcounter{table}{0}
	\renewcommand{\thetable}{S\arabic{table}}%
	\setcounter{figure}{0}
	\renewcommand{\thefigure}{S\arabic{figure}}%
	\setcounter{equation}{0}
	\renewcommand{\theequation}{S\arabic{equation}}
}

\begin{document}
	\pagestyle{headings}
	\mainmatter
	\def\ECCVSubNumber{568}  
	
	\title{XMem: Long-Term Video Object Segmentation with an Atkinson-Shiffrin Memory Model \vspace{-1em}}

		\titlerunning{XMem: Long-Term Video Object Segmentation}
		%
		\author{Ho Kei Cheng \and Alexander G. Schwing}
		\index{Cheng,Ho Kei}
		\index{Schwing,Alexander}
		\authorrunning{H.K. Cheng and A.G. Schwing}
		%
		\institute{University of Illinois Urbana-Champaign \\
			\email{\{hokeikc2, aschwing\}@illinois.edu}}
	\maketitle
	
	\begin{figure}[h]
	    \vspace{-1cm}
	    \centering
	    \begin{tabular}{c@{\hskip 1pt}c@{\hskip 1pt}c@{\hskip 1pt}c@{\hskip 1pt}c}
	         \includegraphics[width=0.19\linewidth]{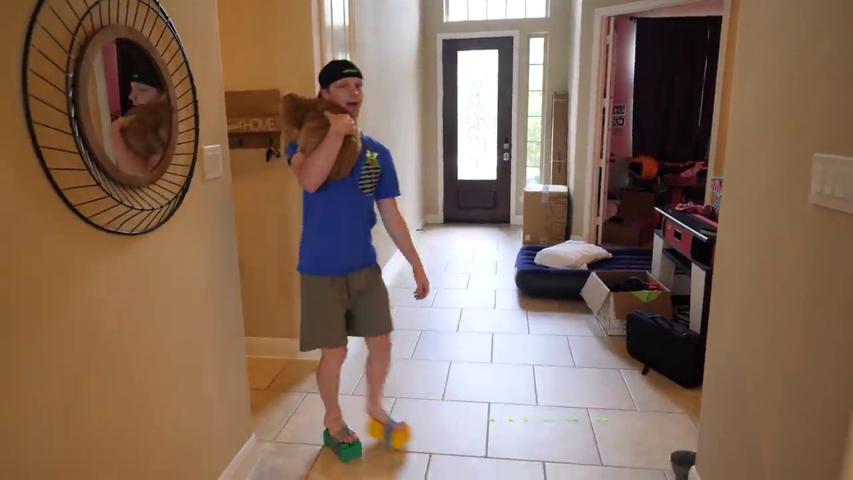} & 
	         \includegraphics[width=0.19\linewidth]{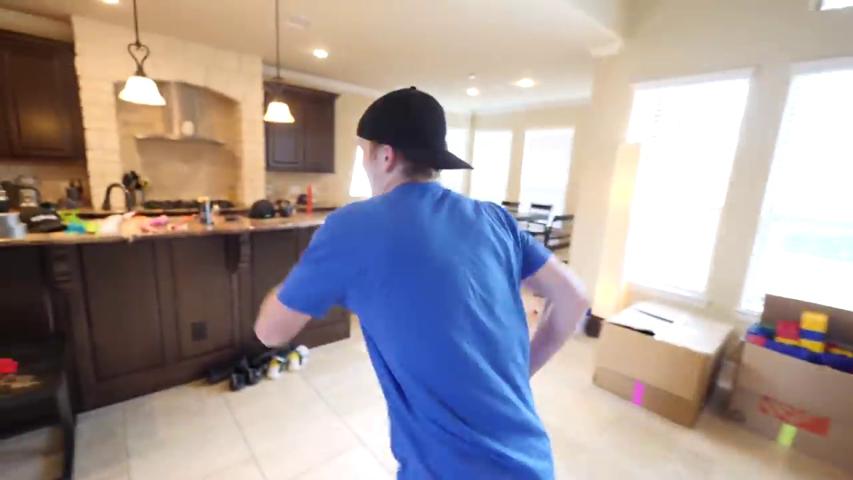} & 
	         \includegraphics[width=0.19\linewidth]{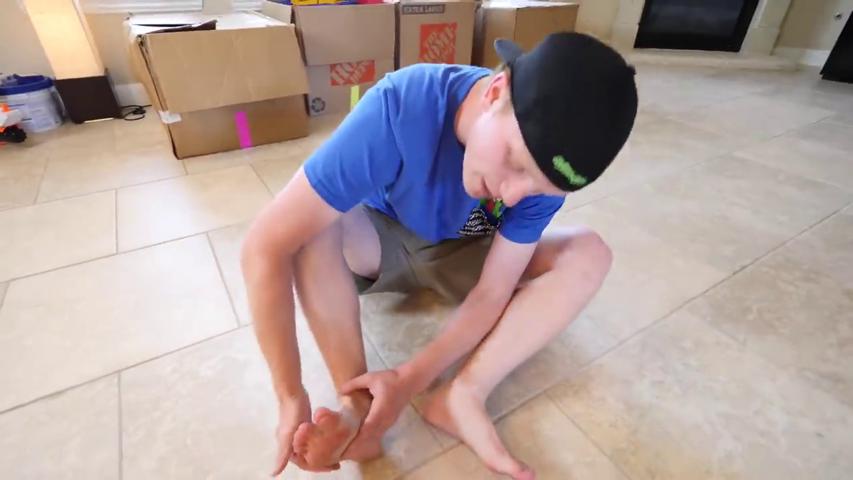} & 
	         \includegraphics[width=0.19\linewidth]{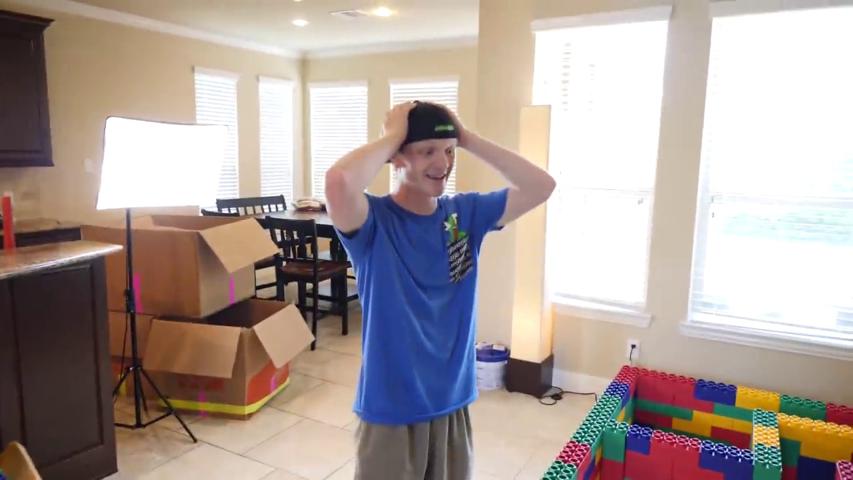} & 
	         \includegraphics[width=0.19\linewidth]{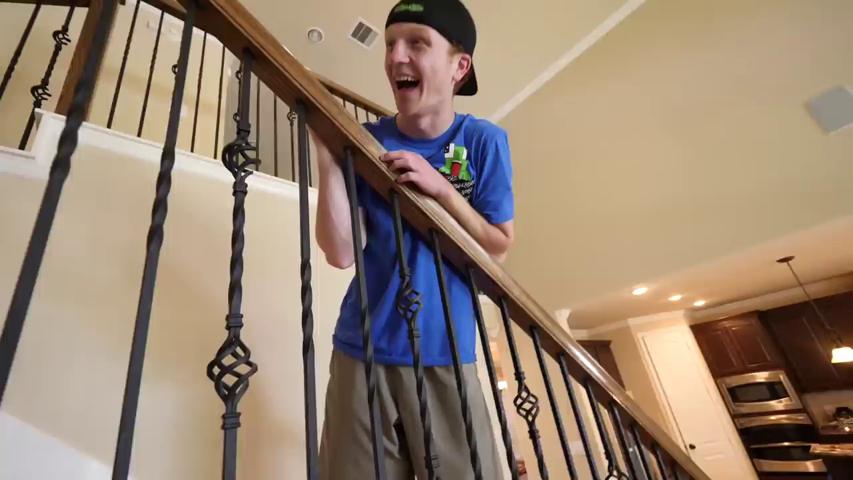} \\
	         \includegraphics[width=0.19\linewidth]{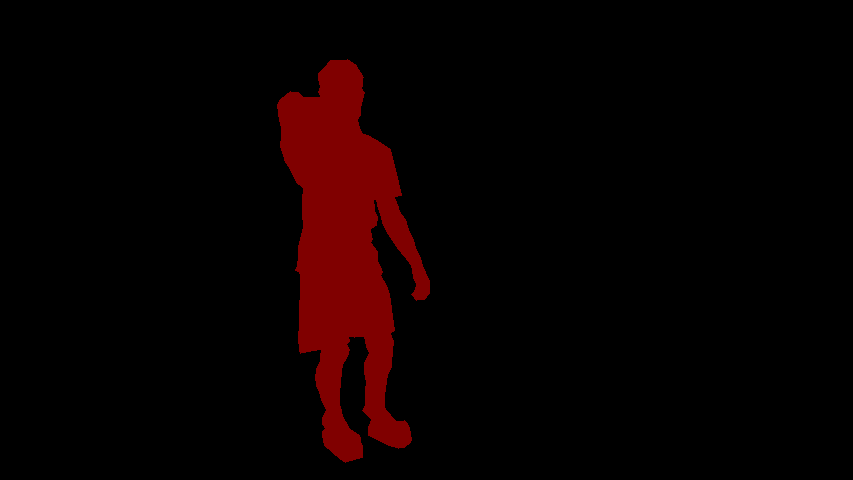} & 
	         \includegraphics[width=0.19\linewidth]{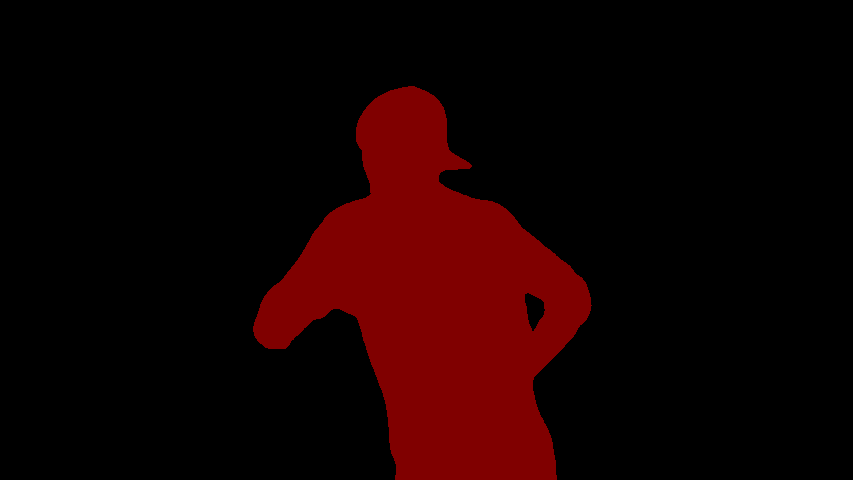} & 
	         \includegraphics[width=0.19\linewidth]{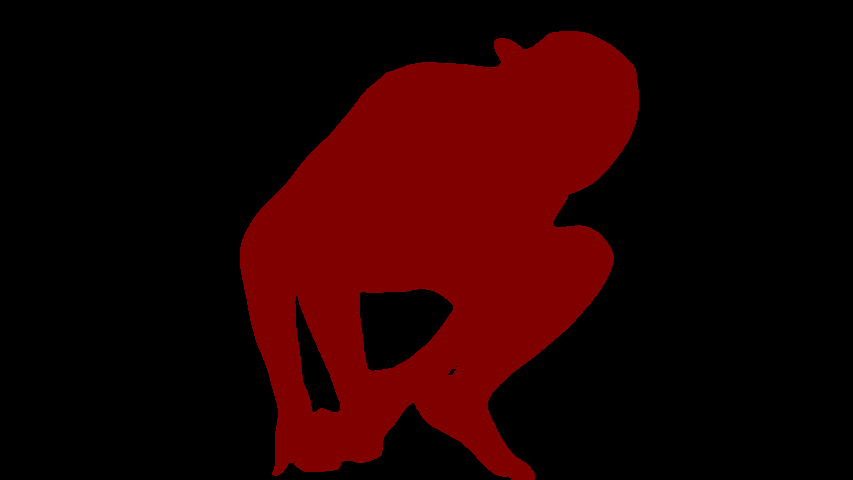} & 
	         \includegraphics[width=0.19\linewidth]{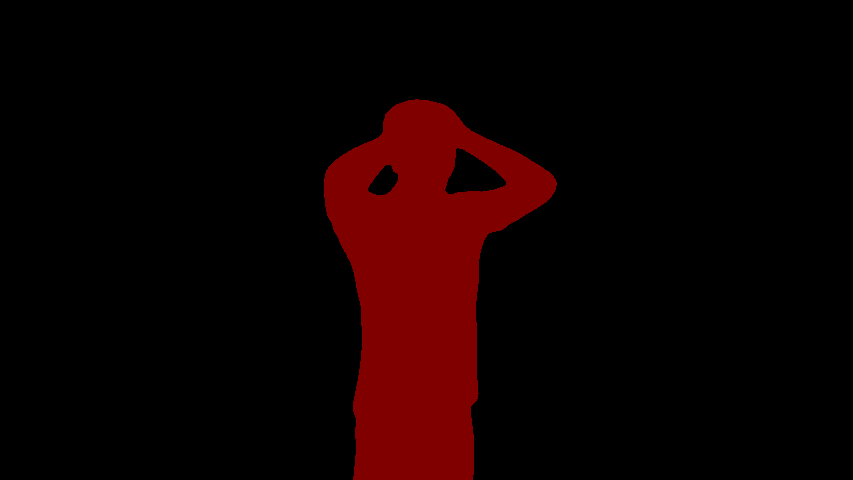} & 
	         \includegraphics[width=0.19\linewidth]{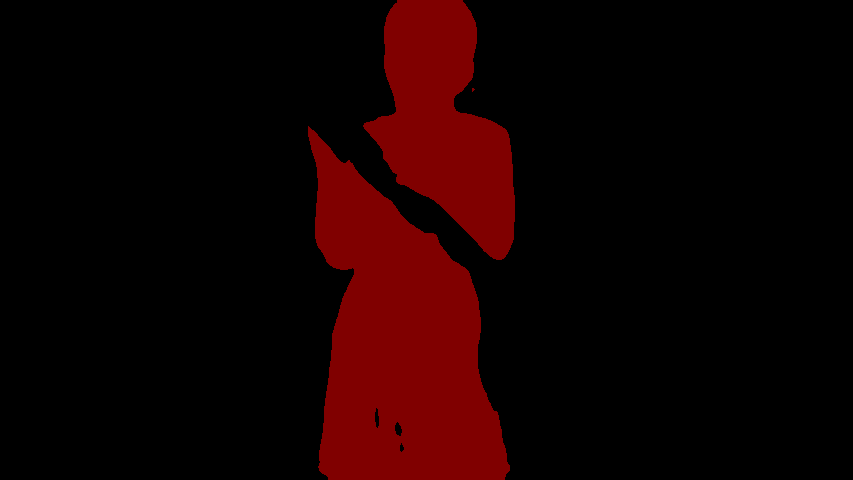} \\
	         Frame 0 (input) & 
	         Frame 295 &
	         Frame 460 & 
	         Frame 1285 & 
	         Frame 2327 
	    \end{tabular}
	    \label{fig:teaser}
	    \vspace{-1.3cm}
	\end{figure}

	\begin{abstract}
		We present XMem, a video object segmentation architecture for long videos with unified feature memory stores inspired by the Atkinson-Shiffrin memory model.
		Prior work on video object segmentation typically only uses one type of feature memory. For videos longer than a minute, a single feature memory model tightly links memory consumption and accuracy. 
		In contrast, following the Atkinson-Shiffrin model, we develop an architecture that incorporates multiple independent yet deeply-connected feature memory stores: a rapidly updated \emph{sensory memory}, a high-resolution \emph{working memory}, and a compact thus sustained \emph{long-term memory}.
		Crucially, we develop a memory potentiation algorithm that routinely consolidates actively used working memory elements into the long-term memory, which avoids memory explosion and minimizes performance decay for long-term prediction.
		Combined with a new memory reading mechanism, XMem greatly exceeds state-of-the-art performance on long-video datasets while being on par with state-of-the-art methods (that do not work on long videos) on short-video datasets.\footnote{\label{fnt:code}Code is available at \href{https://hkchengrex.github.io/XMem}{\nolinkurl{hkchengrex.github.io/XMem}}}
	\end{abstract}
	
	\vspace{-2.5em}
\section{Introduction}
\vspace{-0.5em}
	Video object segmentation (VOS) highlights specified target objects in a given video. Here, we focus on the semi-supervised setting where a first-frame annotation is provided by the user, and the method segments objects in all other frames as accurately as possible while preferably running in real-time, online, and while having a small memory footprint even when processing long videos.
	
	As information has to be propagated from the given annotation to other video frames, most VOS methods employ a \emph{feature memory} to store relevant deep-net representations of an object. 
	Online learning methods~\cite{caelles2017oneOSVOS,voigtlaender2017onlineOnAVOS,robinson2020learningTargetModel} use the weights of a network as their feature memory. This requires  training at test-time, which slows down prediction.
	Recurrent methods propagate information often from the most recent frames, either via a mask~\cite{perazzi2017learningMaskTrack} or via a hidden representation~\cite{hu2017maskrnn,ventura2019rvos}. These methods are prone to drifting and struggle with occlusions. 
	Recent state-of-the-art VOS methods use attention~\cite{oh2019videoSTM,hu2021learning,xie2021efficient,cheng2021stcn,yang2021associating} to link representations of  past frames stored in the feature memory with features extracted from the newly observed query frame which needs to be segmented. 
	Despite the high performance of these methods, they require a large amount of GPU memory to store past frame representations. In practice, they usually struggle to handle videos longer than a minute on consumer-grade hardware.
	
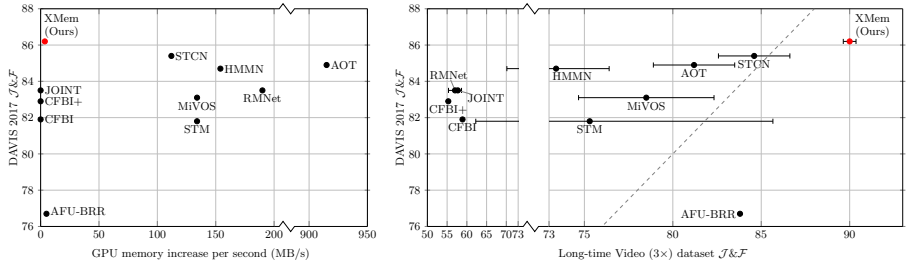
\begin{figure}[t]
		\centering
		\begin{tabular}{c@{\hspace{0pt}}c}
		\resizebox{!}{3.5cm}{
			\begin{tikzpicture}[every node/.append style={align=center}]

	\begin{groupplot}[
        group style={
            group name=mem,
            group size=2 by 1,
            yticklabels at=edge left,
            horizontal sep=0pt,
        },
        height=6cm, width=9cm,
        ymin=76, ymax=88,
        scale only axis = true, enlargelimits=false,
        grid=both,
    ]
    \nextgroupplot[
        xmin=0, xmax=200,
        xtick={0,50,100,150,200},
        ytick={76,78,80,82,84,86,88},
        axis y line=left,
        width=6.43cm,
    ] 
    \addplot[
        scatter/classes={cc={black}, gm={black}, ours={red}},
		scatter, mark=*, only marks, 
		scatter src=explicit symbolic,
		nodes near coords*={\Label},
		nodes near coords align={\Dir},
		visualization depends on={value \thisrow{label} \as \Label}, 
		visualization depends on={value \thisrow{direction} \as \Dir}, 
    ] table [meta=class] {
		x y class direction label
        3.6 86.2 ours {above=12pt, right=-5pt} {}
		5 76.7 cc {above=2pt, right=0pt} AFU-BRR
		0 83.5 cc right JOINT
		0 81.9 cc right CFBI
		0 82.9 cc right CFBI+
		134 81.8 gm below STM
		134 83.1 gm below MiVOS
		190 83.5 gm below RMNet
		154 84.7 gm right HMMN
		112 85.4 gm right STCN
	};
    \node[] at (axis cs: 18.0,87.5) {XMem};
    \node[] at (axis cs: 18.0,86.7) {(Ours)};

    \nextgroupplot[
        xmin=870, xmax=950,
        xtick={900,950},
        ytick={78,80,82,84,86},
        extra y tick style={grid},
        axis y line=right,
        axis x discontinuity=crunch,
        width=2.57cm,
    ]
    \addplot[
            scatter/classes={cc={black}, gm={black}, ours={red}},
    		scatter, mark=*, only marks, 
    		scatter src=explicit symbolic,
    		nodes near coords*={\Label},
    		nodes near coords align={\Dir},
    		nodes near coords style={align=center},
    		visualization depends on={value \thisrow{label} \as \Label}, 
    		visualization depends on={value \thisrow{direction} \as \Dir}, 
        ] table [meta=class] {
			x y class direction label
			915 84.9 gm right AOT
		};
    \end{groupplot}
    
    \node[anchor=north] (title-x) at ($(mem c1r1.south)-(-1.2cm,0.5cm)$) {GPU memory increase per second (MB/s)};
    \node[anchor=south, rotate=90] (title-y) at ($(mem c1r1.south west)!0.5!(mem c1r1.north west)-(0.5cm,0cm)$) {DAVIS 2017 $\mathcal{J}\&\mathcal{F}$};
    
\end{tikzpicture}
		} &
		\resizebox{!}{3.5cm}{
			\begin{tikzpicture}[every node/.append style={align=center}]
	
\tikzset{
	every pin/.style={fill=white,rectangle,rounded corners=3pt,font=},
	every pin edge/.style={draw=black!50!white,line width=0.3pt},
	small dot/.style={fill=black,circle,scale=0.1},
}

	\begin{groupplot}[
		group style={
			group name=longshort,
			group size=3 by 1,
			yticklabels at=edge left,
			horizontal sep=0pt,
		},
			height=6cm,
			ymin=76, ymax=88,
			scale only axis = true, enlargelimits=false,
			grid=both,
		]
		
		\nextgroupplot[
			xmin=50, xmax=73,
			xtick={50,55,60,65,70,73},
			ytick={76,78,80,82,84,86,88},
			axis y line=left,
			width=2.5cm,
		] 
		\addplot[
			scatter/classes={cc={black}, gm={black}, ours={red}},
			scatter, mark=*, only marks, 
			scatter src=explicit symbolic,
			nodes near coords*={\Label},
			nodes near coords align={\Dir},
			visualization depends on=\thisrow{ex} \as \erroshift,
			visualization depends on={value \thisrow{label} \as \Label}, 
			visualization depends on={value \thisrow{direction} \as \Dir}, 
		]plot [error bars/.cd, x dir = both, x explicit] 
			table [meta=class, x=x, y=y, x error=ex] {
				x y ex class direction label
				58.9 81.9 0.00 cc below CFBI
				55.3 82.9 0.00 cc below CFBI+
				57.0 83.5 1.65 gm {left=4pt, above=0pt} {}
				57.7 83.5 0.24 cc {right=6pt, below=0pt} {}
				90.1 86.2 0.44 ours {above=12pt, right=0pt} {XMem \vspace{0.5mm} (Ours)} 
				83.8 76.7 0.00 cc left AFU-BRR
				75.3 81.8 13.04 gm below STM
				78.5 83.1 4.53 gm below MiVOS
				73.4 84.7 3.26 gm below HMMN
				81.2 84.9 2.53 gm below AOT
				84.6 85.4 1.93 gm below STCN
			};
		\node[small dot,pin={[pin distance=1em,anchor=south]120,inner sep=0pt:{RMNet}}] at (57.0,83.5) {};
		\node[small dot,pin={[pin distance=1em,anchor=west]320,inner sep=0pt:{JOINT}}] at (57.7,83.5) {};
		
		\nextgroupplot[
		xmin=73, xmax=73,
		xtick={0},
		ytick={78,80,82,84,86},
		ytick style={draw=none},
		axis y line=none,
		axis x discontinuity=crunch,
		width=0.85cm,
		]
		\addplot[
		scatter/classes={cc={black}, gm={black}, ours={red}},
		scatter, mark=*, only marks, 
		scatter src=explicit symbolic,
		nodes near coords*={\Label},
		nodes near coords align={\Dir},
		nodes near coords style={align=center},
		visualization depends on=\thisrow{ex} \as \erroshift,
		visualization depends on={value \thisrow{label} \as \Label}, 
		visualization depends on={value \thisrow{direction} \as \Dir}, 
		]plot [error bars/.cd, x dir = both, x explicit] 
		table [meta=class, x=x, y=y, x error=ex] {
			x y ex class direction label
		};
		
		\nextgroupplot[
			xmin=73, xmax=93,
			xtick={73,75,80,85,90},
			ytick={76,78,80,82,84,86},
			extra y tick style={grid},
			axis y line=right,
			width=9.74cm,
		]
		\addplot[
			scatter/classes={cc={black}, gm={black}, ours={red}},
			scatter, mark=*, only marks, 
			scatter src=explicit symbolic,
			nodes near coords*={\Label},
			nodes near coords align={\Dir},
			nodes near coords style={align=center},
			visualization depends on=\thisrow{ex} \as \erroshift,
			visualization depends on={value \thisrow{label} \as \Label}, 
			visualization depends on={value \thisrow{direction} \as \Dir}, 
		]plot [error bars/.cd, x dir = both, x explicit] 
			table [meta=class, x=x, y=y, x error=ex] {
				x y ex class direction label
				90.0 86.2 0.36 ours {above=8pt, right=0pt} {}
				83.8 76.7 0.00 cc left AFU-BRR
				75.3 81.8 10.36 gm below STM
				78.5 83.1 3.83 gm below MiVOS
				73.4 84.7 3.00 gm {below=6pt, right=-6pt} HMMN
				81.2 84.9 2.30 gm below AOT
				84.6 85.4 2.02 gm below STCN
			};
		\node[] at (axis cs: 91.3,87.5) {XMem};
        \node[] at (axis cs: 91.3,86.7) {(Ours)};
		\addplot[gray,dashed] coordinates {(74,74)(88,88)};
	\end{groupplot}
	\node[anchor=north] (title-x) at ($(mem c1r1.south)-(-3.0cm,0.5cm)$) {Long-time Video (3$\times$) dataset $\mathcal{J}\&\mathcal{F}$};
	\node[anchor=south, rotate=90] (title-y) at ($(mem c1r1.south west)!0.5!(mem c1r1.north west)-(0.5cm,0cm)$) {DAVIS 2017 $\mathcal{J}\&\mathcal{F}$};
\end{tikzpicture}
		}
		\end{tabular}
		\vspace{-0.5em}
    	\caption{Do state-of-the-art VOS algorithms scale well? 
    		\textbf{Left}: Memory scaling with respect to short-term segmentation quality.
    		\textbf{Right}: Segmentation quality scaling from standard short videos (y-axis) to long videos (x-axis) -- the dashed line indicates a 1:1 performance ratio. Error bars show standard deviations in memory sampling if applicable. See Section~\ref{sec:expr-long-vid} for details.
    		\label{fig:scaling}
    	}
	    \vspace{-1.5em}
    \end{figure}

	Methods that are specifically designed for VOS in long videos exist~\cite{Liang2020AFBURR,li2020fastGlobalContext}. However, they often sacrifice segmentation quality. Specifically,  these methods reduce the size of the representation during feature memory insertion by merging new features with those already stored in the feature memory. 
	As high-resolution features are compressed right away, they produce less accurate segmentations.
	Figure~\ref{fig:scaling} shows the relation between GPU memory consumption and segmentation quality in short/long video datasets (details are given in Section~\ref{sec:expr-long-vid}).
	
	We think this undesirable connection of performance and GPU memory consumption is a direct consequence of using a single feature memory type. 
	To address this limitation we propose a unified memory architecture, dubbed XMem. 
	Inspired by the Atkinson–Shiffrin memory model~\cite{atkinson1968human} which hypothesizes that the human memory consists of three components, 
	XMem maintains three independent yet deeply-connected feature memory stores: a rapidly updated \emph{sensory memory}, a high-resolution \emph{working memory}, and a compact thus sustained \emph{long-term memory}. 
	In XMem, the sensory memory 
	corresponds to the  hidden representation of a GRU~\cite{cho2014propertiesGRU} which is updated every frame. It provides temporal smoothness but fails for long-term prediction due to representation drift.
	To complement, the working memory 
	is agglomerated from a subset of historical frames and considers them equally~\cite{oh2019videoSTM,cheng2021stcn} without drifting over time. 
	To control the size of the working memory, XMem routinely consolidates its representations into the long-term memory, inspired by the consolidation mechanism in the human memory~\cite{squire2015memory}.
	XMem stores long-term memory as a set of highly compact prototypes. For this, we develop a memory potentiation algorithm that aggregates richer information into these prototypes to prevent aliasing due to sub-sampling. 
	To read from the working and long-term memory, we  devise a  space-time memory reading operation. 
	The three feature memory stores combined permit handling long videos with high accuracy while keeping GPU memory usage low. 
	
	We find XMem to greatly exceed prior state-of-the-art results on the Long-time Video dataset~\cite{Liang2020AFBURR}. Importantly, XMem is also on par with current state-of-the-art  (that cannot handle long videos) on short-video datasets~\cite{Pont-Tuset_arXiv_2017,xu2018youtubeVOS}. In summary:
	\vspace{-0.6em}
	\begin{itemize}
		\item We devise XMem. Inspired by the Atkinson–Shiffrin memory model~\cite{atkinson1968human}, we introduce memory stores with different temporal scales and equip them with a  memory reading operation for high-quality video object segmentation on both long and short videos.
		\item We develop a memory consolidation algorithm that selects representative prototypes from the working memory, and a memory potentiation algorithm that enriches these prototypes into a compact yet powerful representation for long-term memory storage.
	\end{itemize}
	
	\vspace{-1.4em}
\section{Related Works}
\vspace{-0.6em}
\textbf{General VOS Methods.}
Most VOS methods employ a \emph{feature memory} to store information given in the first frame and to segment any new frames. 
Online learning approaches either train or fine-tune their networks at test-time and are therefore typically slow in inference~\cite{caelles2017oneOSVOS,voigtlaender2017onlineOnAVOS,maninis2018videoOSVOSS}. 
Recent improvements are more efficient~\cite{meinhardt2020make,robinson2020learningTargetModel,park2021learning,bhat2020learningLWL}, but they still require online adaptation which is sensitive to the input and has diminishing gains when more training data is available.
In contrast, tracking-based approaches~\cite{perazzi2017learningMaskTrack,wang2019fastSiamMask,cheng2018fastTrackingParts,jang2017onlineTrident,chen2020stateAwareTracker,oh2018fastRGMP,zhang2019fast,hu2018motion,ventura2019rvos,hu2017maskrnn,xu2019spatiotemporal} perform frame-to-frame propagation and are thus efficient at test-time. They however lack long-term context and often lose track after object occlusions.
While some methods~\cite{voigtlaender2019feelvos,yang2020collaborativeCFBI,wang2019ranet,johnander2019generative,li2018video,chen2018blazinglyFast} also include the first reference frame for global matching, the context is still limited and it becomes harder to match as the video progresses.
To address the context limitation, recent state-of-the-art methods use more past frames as feature memory~\cite{oh2019videoSTM,Duarte2019Capsule,zhang2020transductive,huang2020fastTemporalAggregation,liang2021video,xu2021reliable,ge2021video}.
Particularly, Space-Time Memory (STM)~\cite{oh2019videoSTM} is popular and has been extended by many follow-up works~\cite{seong2020kernelizedMemory,cheng2021mivos,hu2021learning,xie2021efficient,wang2021swiftnet,lu2020videoGraphMem,cheng2021stcn,seong2021hierarchical,mao2021joint}.
Among these extensions, we use STCN~\cite{cheng2021stcn} as our working memory backbone as it is simple and effective. However, most variants cannot handle long videos due to the ever-expanding feature memory bank of STM. 
AOT~\cite{yang2021associating} is a recent work that extends the attention mechanism to transformers but does not solve the GPU memory explosion problem. 
Some methods~\cite{mao2021joint,duke2021sstvos} employ a local feature memory window that fails to consider long-term context outside of this window.
In contrast, \emph{XMem} uses multiple memory stores to capture different temporal contexts while keeping the GPU memory usage strictly bounded due to our long-term memory and consolidation.

\noindent\textbf{Methods that Specialize in Handling Long Videos.}
Liang \etal~\cite{Liang2020AFBURR} propose AFB-URR which selectively uses exponential moving averages to merge a given  memory element with existing ones if they are close, or to add it as a new element otherwise. A least-frequently-used-based mechanism is employed to remove unused features when the feature memory reaches a predefined limit. 
Li \etal~\cite{li2020fastGlobalContext} propose the global context module. It averages all past memory into a single representation, thus having zero GPU memory increase over time.
However, both of these methods \emph{eagerly} compress new high-resolution feature memory into a compact representation, thus sacrificing segmentation accuracy. Our multi-store feature memory avoids eager compression and achieves much higher accuracy in both short-term and long-term predictions.

	\vspace{-0.1em}
\section{XMem}
\vspace{-0.1em}
	\begin{figure}[t]
		\centering
		\includegraphics[width=.85\linewidth]{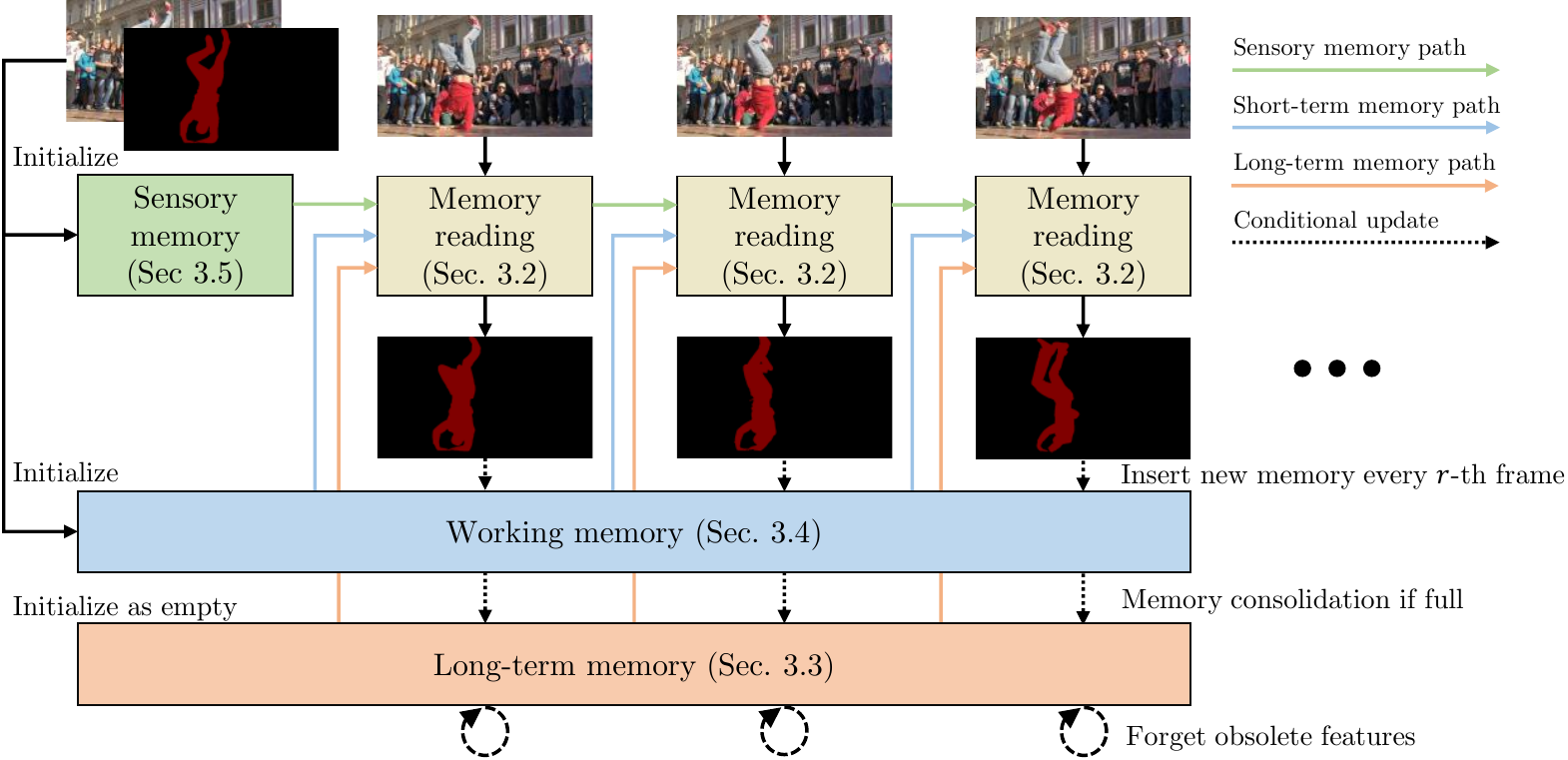}
		\vspace{-0.5em}
		\caption{
			Overview of XMem. The memory reading operation extracts relevant features from all three memory stores and uses those features to produce a mask. To incorporate new memory, the sensory memory is updated every frame while the working memory is only updated every $r$-th frame. The working memory is consolidated into the long-term memory in a compact form when it is full, and the long-term memory will forget obsolete features over time.
		}
		\label{fig:overview}
		\vspace{-1.5em}
	\end{figure}
	\vspace{-0.1em}
	\subsection{Overview}
	Figure~\ref{fig:overview} provides an overview of XMem.
	For readability, we consider a single target object. However, note that XMem is implemented to deal with multiple objects, which is straightforward.
	Given the image and target object mask at the first frame (top-left of Figure~\ref{fig:overview}), XMem tracks the object and generates corresponding masks for subsequent query frames. 
	For this, we first initialize the different feature memory stores using the inputs. 
	For each subsequent query frame, we perform memory reading (Section~\ref{sec:mem_reading}) from long-term memory (Section~\ref{sec:lt_memory}), working memory (Section~\ref{sec:working_memory}), and sensory memory (Section~\ref{sec:sensory_memory}) respectively.
	The readout features are used to generate a segmentation mask. 
	Then, we update each of the feature memory stores at different frequencies. We update the sensory memory every frame and insert features into the working memory at every $r$-th frame. 
	When the working memory reaches a pre-defined maximum of $T_{\max}$ frames, we consolidate features from the working memory into the long-term memory in a highly compact form.
	When the long-term memory is also full (which only happens after processing thousands of frames), we discard obsolete features to bound the maximum GPU memory usage.
	These feature memory stores work in conjunction to provide high-quality features with low GPU memory usage even for very long videos.
	
	XMem consists of three end-to-end trainable convolutional networks as shown in Figure~\ref{fig:one_query}: a \emph{query encoder} that extracts query-specific image features, a \emph{decoder} that takes the output of the memory reading step to generate an object mask, and a \emph{value encoder} that combines the image with the object mask to extract new memory features.
	See Section~\ref{sec:implementation} for details of these networks.
	In the following, we will first describe the memory reading operation before discussing each feature memory store in detail.
	
	\vspace{-0.3em}
	\subsection{Memory Reading}\label{sec:mem_reading}
	\vspace{-0.1em}
	\begin{figure}[t]
		\centering
		\includegraphics[width=.99\linewidth]{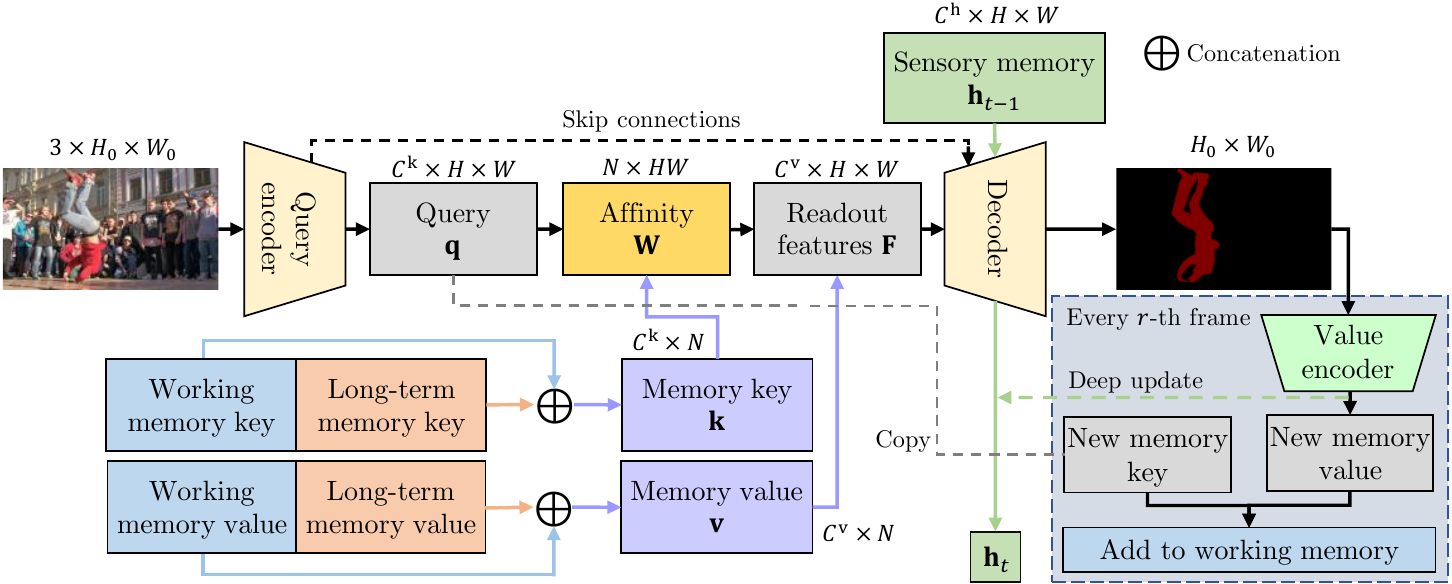}
		\caption{
			Process of memory reading and mask decoding of a single query frame. We extract query $\mathbf{q}$ from the image and perform attention-based memory reading from the working/long-term memory to obtain features $F$. Together with the sensory memory, it is fed into the decoder to generate a mask. For every $r$-th frame, we store new features into the working memory and perform a deep update to the sensory memory.
		}
		\label{fig:one_query}
		\vspace{-1.5em}
	\end{figure}
	
	Figure~\ref{fig:one_query} illustrates the process of memory reading and mask generation for a single frame. 
	The mask is computed via the decoder which uses as input the short-term sensory memory $\mathbf{h}_{t-1}\in\mathbb{R}^{\C{h}\times H\times W}$ and a feature $\mathbf{F}\in\mathbb{R}^{\C{v}\times H\times W}$ representing information stored in both the long-term and the working memory.

	The feature $\mathbf{F}$ representing information stored in both the long-term and the working memory is computed via the readout operation 
	\begin{equation}
	\vspace{-0.1em}
		\mathbf{F} = \mathbf{v}\mathbf{W(\mathbf{k}, \mathbf{q})}.
		\label{eq:readout}
	\vspace{-0.1em}
	\end{equation}
	Here,  $\mathbf{k}\in\mathbb{R}^{\C{k}\times N}$ and  $\mathbf{v}\in\mathbb{R}^{\C{v}\times N}$ are $\C{k}$- and $\C{v}$-dimensional keys and values for a total of $N$  memory elements which are stored in both the long-term and working memory.
	Moreover, $\mathbf{W(\mathbf{k}, \mathbf{q})}$ is an affinity matrix of size $N\times HW$, representing a readout operation that is controlled by the key $\mathbf{k}$ and a query $\mathbf{q}\in\mathbb{R}^{\C{k}\times HW}$ obtained from the query frame through the query encoder.
	The readout operation maps every query element to a distribution over all $N$ memory elements and correspondingly aggregates their values $\mathbf{v}$.
	
\begin{figure}[t]
		\centering
		\scriptsize
		\begin{tabular}{cccc}
			\includegraphics[width=0.19\linewidth]{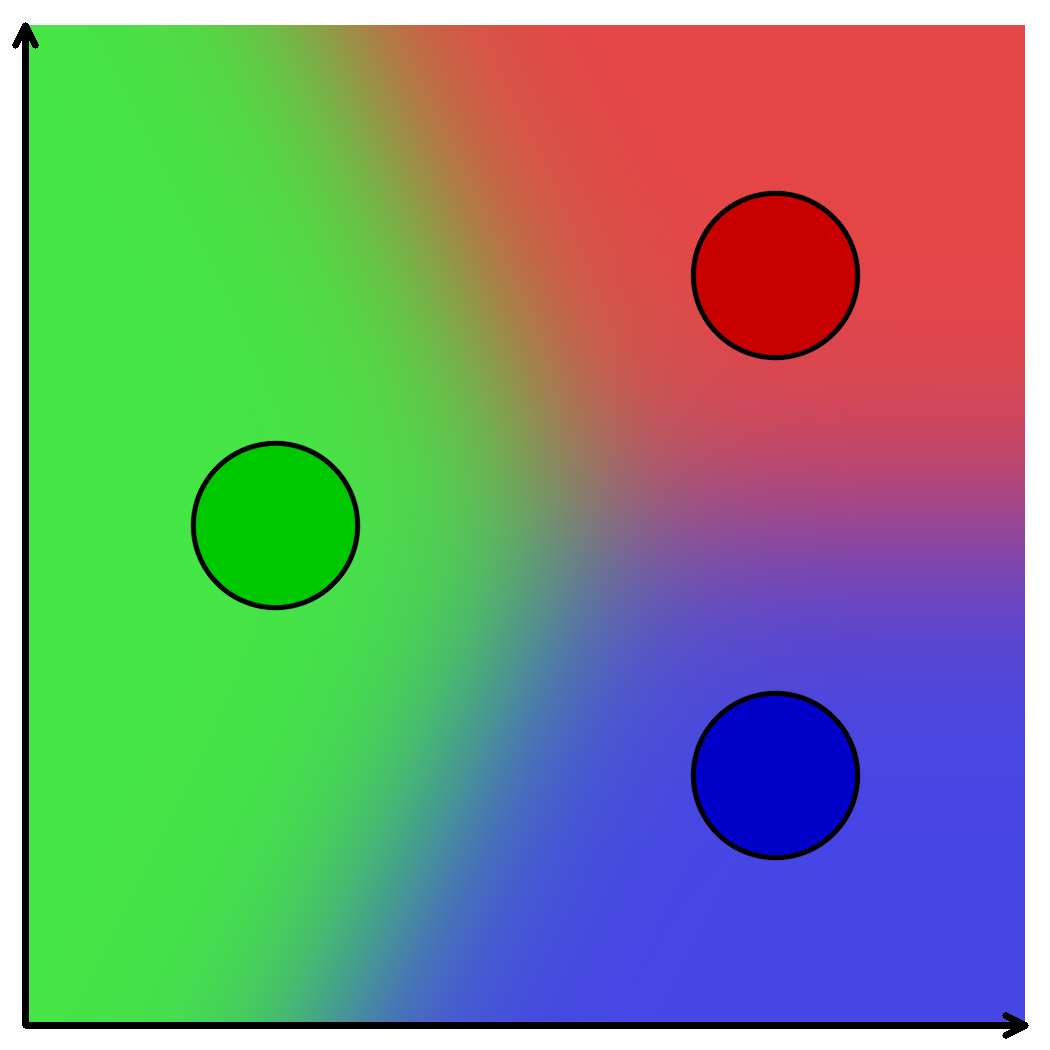}
			& \includegraphics[width=0.19\linewidth]{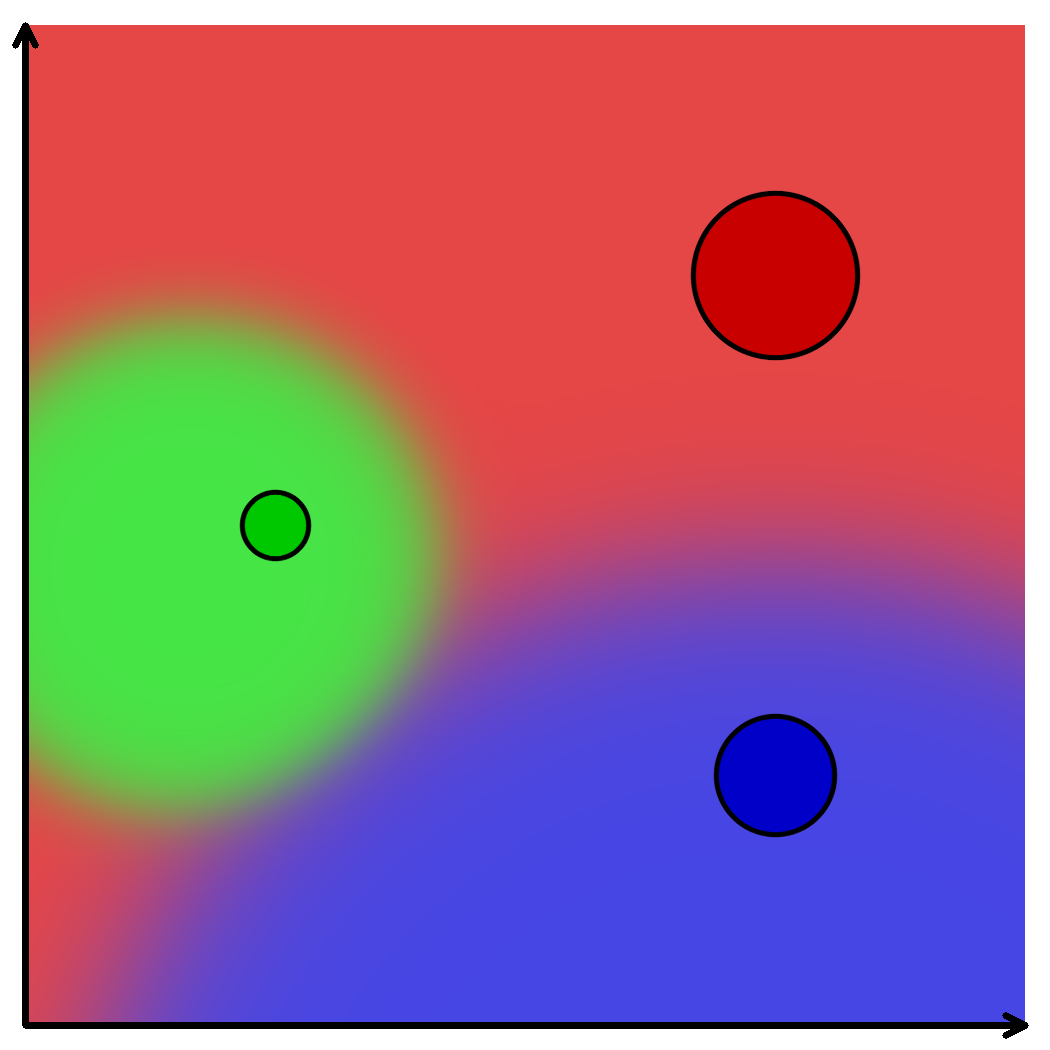}
			& \includegraphics[width=0.19\linewidth]{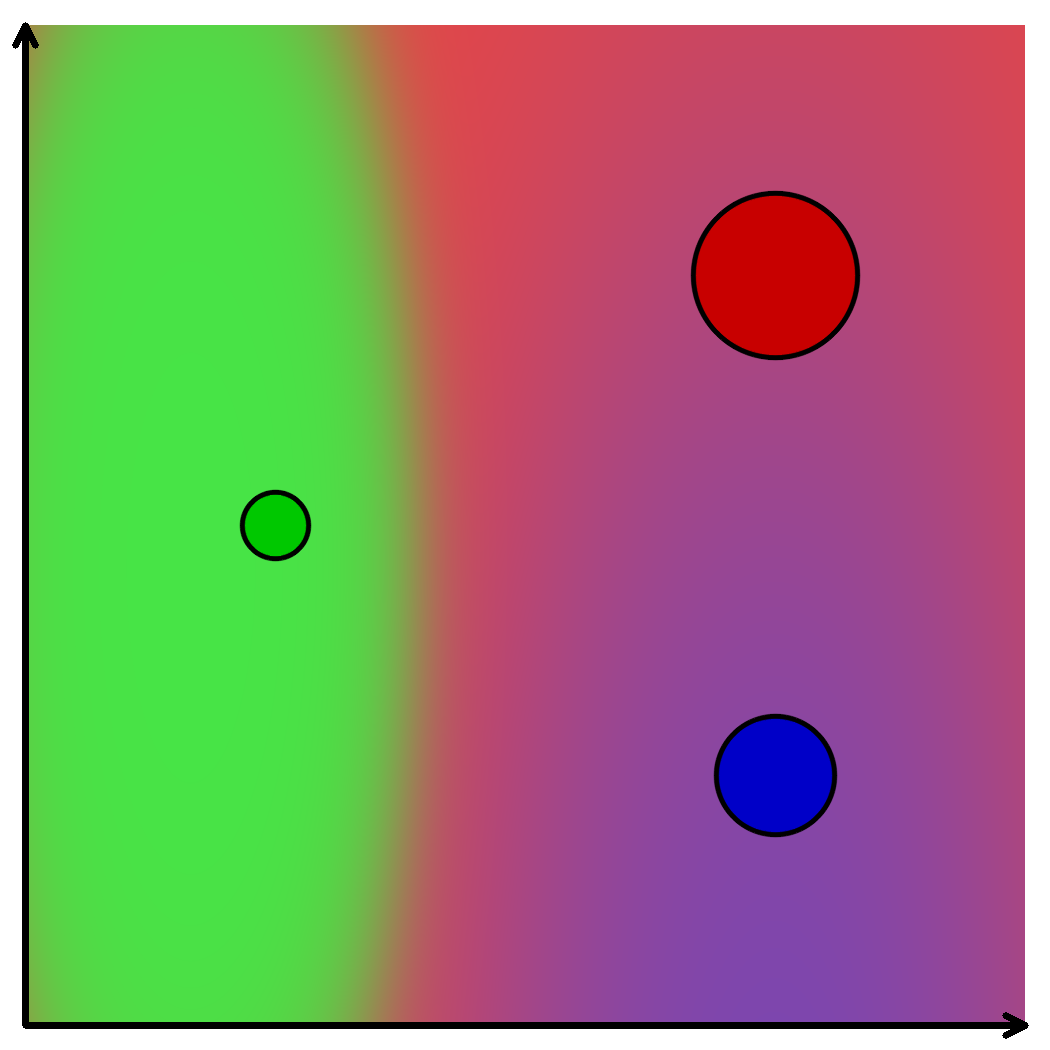}
			& \includegraphics[width=0.19\linewidth]{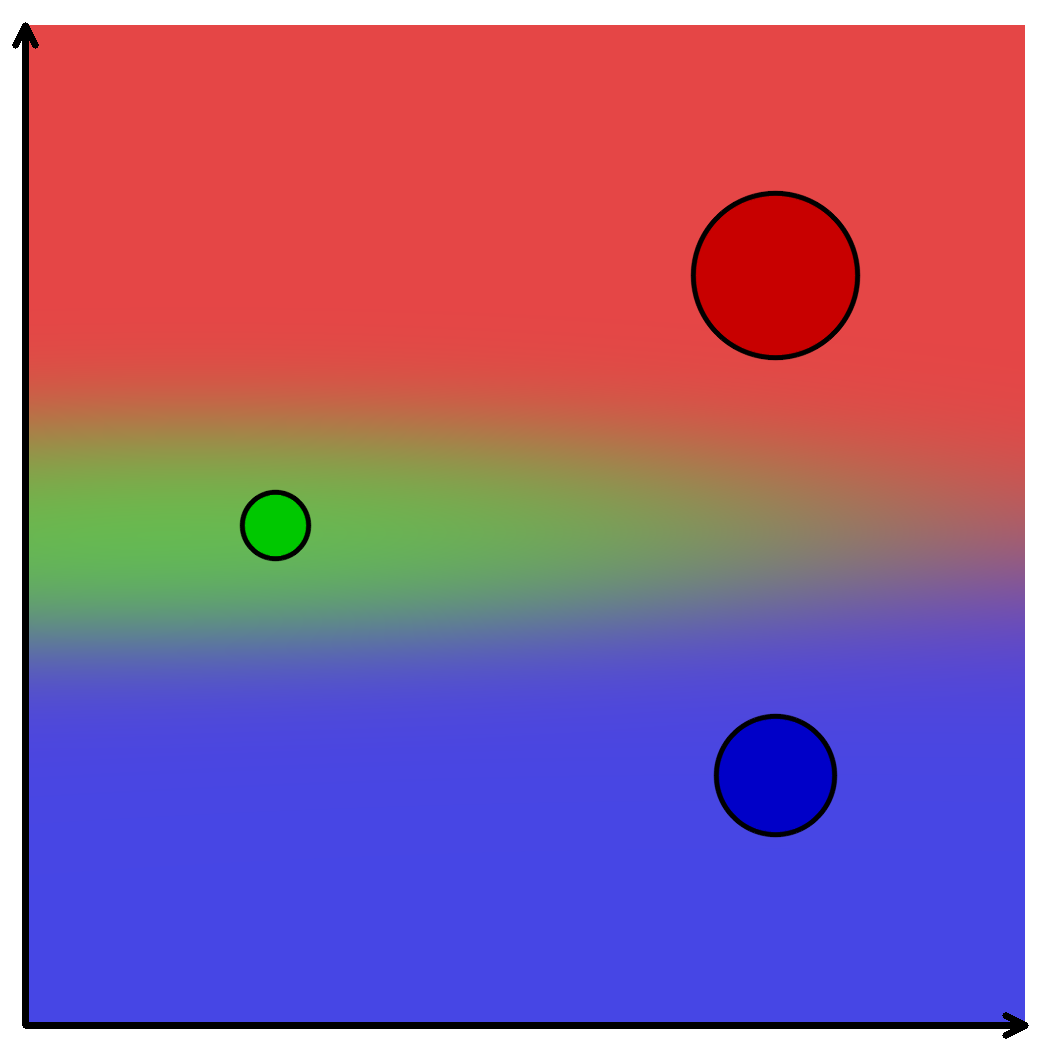}\\
			(a) L2 similarity
			& (b) With shrinkage
			& \makecell[c]{(c) With both \\(query 1)}
			& \makecell[c]{(d) With both \\(query 2)} \\
		\end{tabular}
		\vspace{-0.5em}
		\caption{
			Visualization of similarity functions in 2D with the background color showing the influence of each memory element (RGB). 
			L2 similarity~(a)~\cite{cheng2021stcn} considers all memory elements uniformly.
			The shrinkage term~(b) allows encoding element-level confidence (visualized by the size of dots) that accounts for the area of influence and sharpness of the mixing weights.
			The selection term allows query-specific interpretation of the memory -- (c) and (d) show its effect with two different queries that focus on the vertical and horizontal dimension respectively.
			(b) can be seen as a case where the selection term is isotropic.
			When combined, we can model more complex similarity relations.
		}
		\label{fig:two_terms}
		\vspace{-1.5em}
	\end{figure}

	The affinity matrix 
	$\mathbf{W}(\mathbf{k}, \mathbf{q})$ is obtained by applying a softmax on the memory dimension (rows) of a similarity matrix $\mathbf{S(\mathbf{k}, \mathbf{q})}$ which contains the pairwise similarity between every key element and every query element. 
	For computing the similarity matrix we note that the L2 similarity proposed in STCN~\cite{cheng2021stcn} is more stable than the dot product~\cite{oh2019videoSTM}, but it is less expressive, e.g., it cannot encode the confidence level of a memory element.
	To overcome this, we propose a new similarity function (\emph{anisotropic L2}) by introducing two new scaling terms that  break the symmetry between key and query. Figure~\ref{fig:two_terms} visualizes their effects.
	
	Concretely, the key is associated with a \textbf{\underline{s}}hrinkage term $\mathbf{s}\in[1, \infty)^N$ and the query is associated with a s\textbf{\underline{e}}lection term $\mathbf{e}\in[0, 1]^{\C{k}\times HW}$.
	Then, the similarity between the $i$-th key element and the $j$-th query element is computed via 
	\vspace{-0.2em}
	\begin{equation}
		\mathbf{S}(\mathbf{k}, \mathbf{q})_{ij} = -\mathbf{s}_i \sum_{c}^{\C{k}}{ \mathbf{e}_{cj} \left( \mathbf{k}_{ci} - \mathbf{q}_{cj} \right)^2 }, 
		\label{eq:similarity}
		\vspace{-0.2em}
	\end{equation}
	which equates to the original L2 similarity~\cite{cheng2021stcn} if $\mathbf{s}_i=\mathbf{e}_{cj}=1$ for all $i$, $j$, and $c$. 
	The \textbf{\underline{s}}hrinkage term $\mathbf{s}$ directly scales the similarity and explicitly encodes confidence -- a high shrinkage represents low confidence and leads to a more local influence. 
	Note that even low-confidence keys can have a high contribution if the query happens to coincide with it -- thus avoiding the memory domination problem of the dot product, as discussed in~\cite{cheng2021stcn}.
	Differently, the s\textbf{\underline{e}}lection term $\mathbf{e}$ controls the relative importance of each channel  in the key space such that attention is given to the more discriminative channels.

	The selection term $\mathbf{e}$ is generated together with the query $\mathbf{q}$ 
	by the query encoder. The shrinkage term $\mathbf{s}$ is collected together with the key $\mathbf{k}$ and the value $\mathbf{v}$ from the working and the long-term memory.\footnote{
	For brevity, we omit the handling of these two scaling terms in memory updates for the rest of the paper. They are updated in the same way as the value.}
	The collection is simply implemented as a concatenation in the last dimension: $\mathbf{k} = \mathbf{k}^\text{w} \oplus \mathbf{k}^\text{lt}$ and $\mathbf{v} = \mathbf{v}^\text{w} \oplus \mathbf{v}^\text{lt}$, where superscripts `w' and `lt' denote working and long-term memory respectively. 
	The working memory consists of key $\mathbf{k}^\text{w}\in\mathbb{R}^{\C{k}\times THW}$ and value $\mathbf{v}^\text{w} \in\mathbb{R}^{\C{v}\times THW}$, where $T$ is the number of working memory frames.
	The long-term memory similarly consists of keys $\mathbf{k}^\text{lt}\in\mathbb{R}^{\C{k}\times L}$ and values $\mathbf{v}^\text{lt} \in\mathbb{R}^{\C{v}\times L}$, where $L$ is the number of long-term memory prototypes.
	Thus, the total number of elements in the working/long-term memory is $N=THW+L$.
	
	Next, we discuss the feature memory stores in detail.

	\subsection{Long-Term Memory}\label{sec:lt_memory}
	\begin{figure}[t]
		\centering
		\includegraphics[width=0.9\linewidth]{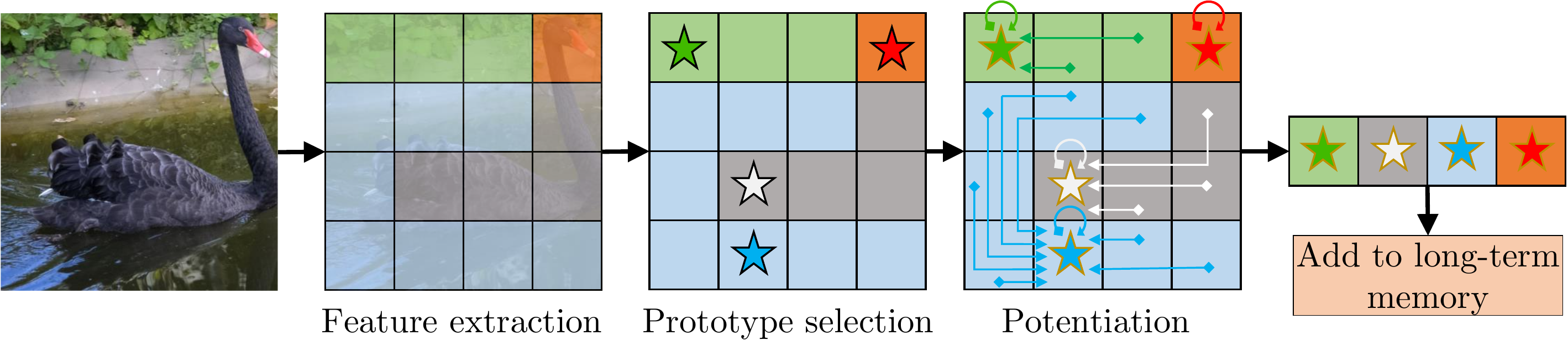}
		\caption{Memory consolidation procedure. Given an image, we extract features as memory keys (image stride exaggerated). We visualize these features with colors.
		For memory consolidation, we first select prototype keys (stars) from the candidates (all grids).
		Then, we invoke potentiation which non-locally aggregates values from all the candidates to generate more representative prototype values (golden outline).
		The resultant prototype keys and values are added to the long-term memory.
		Only one frame is shown here -- in practice multiple frames are used in a single consolidation.}
		\label{fig:lt_procedure}
		\vspace{-1em}
	\end{figure}
	
	\vspace{-0.2em}
	\subsubsection{Motivation.}
	A long-term memory is crucial for handling long videos. With the goal of storing a set of compact (consume little GPU memory) yet representative (lead to high segmentation quality) memory features, we design a \emph{memory consolidation} procedure that selects \emph{prototypes} from the working memory and enriches them with a \emph{memory potentiation} algorithm, as illustrated in Figure~\ref{fig:lt_procedure}.
	
	We perform memory consolidation when the working memory reaches a pre-defined size $T_{\text{max}}$.
	The first frame (with user-provided ground-truth) and the most recent $T_{\text{min}}-1$ memory frames will be kept in the working memory as a high-resolution buffer while the remainder ($T_{\max}-T_{\min}$ frames) are \textit{candidates} for being converted into long-term memory representations. 
	We refer to the keys and values of these candidates as $\mathbf{k}^{\text{c}}\subset\mathbf{k}^{\text{w}}$ and $\mathbf{v}^{\text{c}}\subset\mathbf{v}^{\text{w}}$ respectively. 
	In the following, we describe the prototype selection process that picks a compact set of prototype keys $\mathbf{k}^\text{p}\subset\mathbf{k}^{\text{c}}$, and the memory potentiation algorithm that generates enriched prototype values $\mathbf{v}^\text{p}$ associated with these prototype keys.
	Finally, these prototype keys and values are appended to the long-term memory $\mathbf{k}^{\text{lt}}$ and $\mathbf{v}^{\text{lt}}$.
	
	\vspace{-0.6em}
	\subsubsection{Prototype Selection.}\label{sec:key_selection}
	In this step, we sample a small representative subset $\mathbf{k}^\text{p}\subset\mathbf{k}^{\text{c}}$ from the candidates as \emph{prototypes}.
	It is essential to pick only a small number of prototypes, as their amount is directly proportional to the size of the resultant long-term memory. 
	Inspired by human memory which moves frequently accessed or studied patterns to a long-term store,  we pick candidates with high \emph{usage}.
	Concretely, we pick the top-$P$ frequently used candidates as prototypes. 
	``Usage'' of a memory element is defined by its cumulative total affinity (probability mass) in the affinity matrix $\mathbf{W}$ (Eq.~\eqref{eq:readout}), and normalized by the duration that each candidate is in the working memory. 
	Note that the duration for each candidate is at least $r\cdot(T_{\min}-1)$, leading to stable usage statistics.
	We obtain the keys of these prototypes as $\mathbf{k}^\text{p}\in\mathbb{R}^{\C{k}\times P}$.
	
	
	\vspace{-0.7em}
	\subsubsection{Memory Potentiation.}
	Note that, so far, our sampling of prototype keys $\mathbf{k}^\text{p}$ from the candidate keys $\mathbf{k}^\text{c}$ is both \emph{sparse} and \emph{discrete}.
	If we were to sample the prototypes values $\mathbf{v}^\text{p}$ in the same manner, the resultant prototypes would inevitably under-represent other candidates and would be prone to \emph{aliasing}. 
	The common technique to prevent aliasing is to apply an anti-aliasing (e.g., Gaussian) filter~\cite{forsyth2011computer}. 
	Similarly motivated, we perform filtering and aggregate more information into every sampled prototype. 
	While standard filtering can be easily performed on the image plane (2D) or the spatial-temporal volume (3D), it leads to blurry features -- especially near object boundaries. 
	To alleviate, we instead construct the neighbourhood for the filtering in the high dimensional ($C^{\text{k}}$) key space, such that the highly expressive adjacency information given by the keys $\mathbf{k}^\text{p}$ and $\mathbf{k}^\text{c}$ is utilized. As these keys have to be computed and stored for memory reading anyway, it is also economical in terms of run-time and memory consumption.
	
	Concretely, for each prototype, we aggregate values from all the value candidates $\mathbf{v}^\text{c}$ via a weighted average. The weights are computed using a softmax over the key-similarity. 
	For this, we conveniently re-use Eq.~\eqref{eq:similarity}.
	By substituting the memory key $\mathbf{k}$ with the candidate key $\mathbf{k}^\text{c}$, and the query $\mathbf{q}$ with the prototype keys $\mathbf{k}^\text{p}$, we obtain the similarity matrix $\mathbf{S}(\mathbf{k}^\text{c}, \mathbf{k}^\text{p})$.
	As before, we use a softmax to obtain the affinity matrix $\mathbf{W}(\mathbf{k}^\text{c}, \mathbf{k}^\text{p})$ (where every prototype corresponds to a distribution over candidates). Then, we compute the prototype values $\mathbf{v}^{\text{p}}$ via
	\begin{equation}
		\mathbf{v}^{\text{p}} = \mathbf{v}^{\text{c}}\mathbf{W}(\mathbf{k}^\text{c}, \mathbf{k}^\text{p}).
	\end{equation}
	Finally, $\mathbf{k}^{\text{p}}$ and $\mathbf{v}^{\text{p}}$ are appended to the long-term memory $\mathbf{k}^\text{lt}$ and $\mathbf{v}^\text{lt}$ respectively -- concluding the memory consolidation process. 
	Note, similar prototypical approximations have been used in transformers~\cite{xiong2021nystr,patrick2021keeping}. Differently, our approach uses a novel prototype selection scheme suitable for video object segmentation. 
	
	\vspace{-0.5em}
	\subsubsection{Removing Obsolete Features.}
	Although the long-term memory is extremely compact with a high ($>6000\%$) compression ratio, memory can still overflow since we are continuously appending new features. Empirically, with a 6GB memory budget (e.g., a consumer-grade mid-end GPU), we can process up to 34,000 frames before running into any memory issues.
	To handle even longer videos, we introduce a least-frequently-used (LFU) eviction algorithm similar to~\cite{Liang2020AFBURR}. 
	Unlike~\cite{Liang2020AFBURR}, our ``usage'' (as defined in Section~\ref{sec:key_selection}, Prototype Selection) is defined by the cumulative affinity after top-$k$ filtering~\cite{cheng2021mivos} which circumvents the introduction of an extra threshold hyperparameter.
	Long-term memory elements with the least usage  will be evicted when a pre-defined memory limit is reached.
	
	\vspace{0.5em}
	The long-term memory is key to enabling efficient and accurate segmentation of long videos.
	Next, we discuss the working memory, which is crucial for accurate short-term prediction. It acts as the basis for the long-term memory.
	
	\subsection{Working Memory}\label{sec:working_memory}
	The working memory stores high-resolution features in a temporary buffer. It facilitates accurate matching in the temporal context of a few seconds. It also acts as a gateway into the long-term memory, as the importance of each memory element is estimated by their usage frequency in the working memory.
	
	In our multi-store feature memory design, we find that a classical instantiation of the working memory is sufficient for good results. 
	We largely employ a baseline STCN-style~\cite{cheng2021stcn} feature memory bank as our working memory, which we will briefly describe for completeness. We refer readers to~\cite{cheng2021stcn} for details. However, note that our memory reading step (Section~\ref{sec:mem_reading}) differs significantly.
	The working memory consists of keys $\mathbf{k}^\text{w} \in \mathbb{R}^{\C{k}\times THW}$ and values $\mathbf{v}^\text{w} \in\mathbb{R}^{\C{v}\times THW}$, where $T$ is the number of working memory frames. 
	The key is encoded from the image and resides in the same embedding space as the query $\mathbf{q}$ while the value is encoded from both the image and the mask.
	Bottom-right of Figure~\ref{fig:one_query} illustrates the working memory update process. 
	At every $r$-th frame, we 
	1) copy the query as a new  key; and
	2) generate a new value by feeding the image and the predicted mask into the value encoder.
	The new key and value are appended to the working memory and are later used in memory reading for subsequent frames.
	To avoid memory explosion, we limit the number of frames in the working memory $T$: $T_{\min} \leq T < T_{\max}$ by consolidating extra frames into the long-term memory store as discussed in Section~\ref{sec:lt_memory}.

	\vspace{-1em}
	\subsection{Sensory Memory}\label{sec:sensory_memory}
	
	The sensory memory focuses on the short-term and retains low-level information such as object location which nicely complements the lack of temporal locality in the working/long-term memory.
	Similar to the working memory, we find a classical baseline to work well. 
	
	\begin{wrapfigure}{r}{0.45\textwidth}
		\vspace{-2em}
		\centering
		\includegraphics[width=0.99\linewidth]{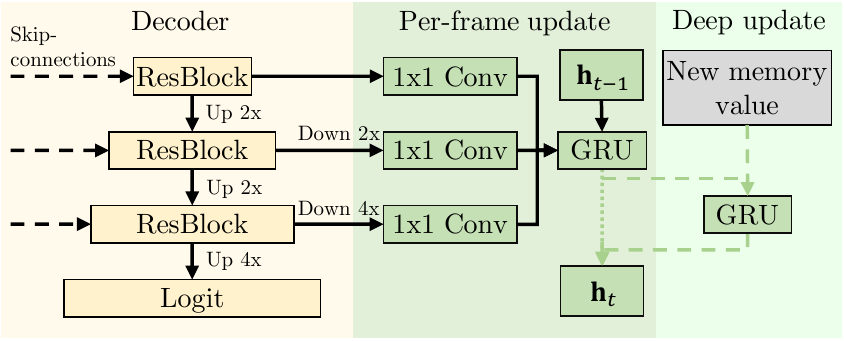}
		\caption{
			Sensory memory update overview. Multi-scale features from the decoder are downsampled and concatenated as inputs to a GRU. 
			In the deep update, a separate GRU is additionally used to refresh the sensory memory.}
		\vspace{-2em}
		\label{fig:sensory}
	\end{wrapfigure}
	Concretely, the sensory memory stores a hidden representation $\mathbf{h}_{t}\in\mathbb{R}^{\C{h}\times H\times W}$, initialized as the zero vector, and propagated by a Gated Recurrent Unit (GRU)~\cite{cho2014propertiesGRU} as illustrated in Figure~\ref{fig:sensory}. 
	This sensory memory is updated every frame using multi-scale features of the decoder.
	At every $r$-th frame, whenever a new working memory frame is generated, we  perform a \emph{deep update}. Features from the value encoder are used to refresh the sensory memory with another GRU. This allows the sensory memory to 
	1) discard redundant information that has already been saved to the working memory, and 
	2) receive updates from a deep network (i.e., the value encoder) with minimal overhead as we are reusing existing features.
	
	\vspace{-1em}
	\subsection{Implementation Details}\label{sec:implementation}
	Here, we describe some key implementation details. To fully reproduce both training and inference, please see our open-source implementation (footnote~\ref{fnt:code}).
	
	\vspace{-1.1em}
	\subsubsection{Networks.}
	Following common practice~\cite{oh2019videoSTM,seong2020kernelizedMemory,Liang2020AFBURR,cheng2021stcn}, we adopt ResNets~\cite{he2016deepResNet} as the feature extractor, removing the classification head and the last convolutional stage. This results in features with stride 16. The query encoder is based on a ResNet-50 and the value encoder is based on a ResNet-18, following~\cite{cheng2021stcn}. 
	To generate the query $\mathbf{q}$, the shrinkage term $\mathbf{s}$, and the selection term $\mathbf{e}$, we apply separate $3\times3$ convolutional projections to the query encoder feature output. Note that both the query and the shrinkage term are used for the current query frame, while the selection term is copied to the memory (along the copy path in Figure~\ref{fig:one_query}) for later use if and only if we are inserting new working memory.
	We set $\C{k}=64$, $\C{v}=512$ following~\cite{cheng2021stcn}, and $\C{h}=64$.
	To control the range of the shrinkage factor to be in $[1, \infty)$, we apply $(\cdot)^2+1$, and to control the range of the selection factor to be in $[0, 1]$, we apply a sigmoid.
	
	The decoder concatenates hidden representation $\mathbf{h}_{t-1}$ and readout feature $\mathbf{F}$. It then iteratively upsamples by $2\times$ at a time until stride 4 while fusing skip-connections from the query encoder at every level, following STM~\cite{oh2019videoSTM}. The stride 4 feature map is projected to a single channel logit via a $3\times3$ convolution, and is bilinearly upsampled to the input resolution. 
	In the multi-object scenario, we use soft-aggregation~\cite{oh2019videoSTM} to fuse the final logits from different objects. Note that the bulk of the computation (i.e., query encoder, affinity $\mathbf{W}$) can be shared between different objects as they are only conditioned on the image~\cite{cheng2021stcn}.
	
	\vspace{-1.1em}
	\subsubsection{Training.}
	Following~\cite{oh2019videoSTM,seong2020kernelizedMemory,Liang2020AFBURR,cheng2021stcn}, we first pretrain our network on synthetic sequences of length three generated by deforming static images. We adopt the open-source implementation of STCN~\cite{cheng2021stcn} without modification, which trains on~\cite{shi2015hierarchicalECSSD,wang2017DUTS,FSS1000,zeng2019towardsHRSOD,cheng2020cascadepsp}.
	Next, we perform the main training on YouTubeVOS~\cite{xu2018youtubeVOS} and DAVIS~\cite{Pont-Tuset_arXiv_2017} with curriculum sampling~\cite{oh2019videoSTM}.
	We note that the default sequence length of three is insufficient to train the sensory memory as it would be heavily dependent on the initial state.
	Thus, we instead sample sequences of length eight. To reduce training time and for regularization, a maximum of three (instead of all) past frames are randomly selected to be the working memory for any query in training time.
	The entire training process takes around 35 hours on two RTX A6000 GPUs.
	Deep updates are performed with a probability of $0.2$, which is $1/r$ as we use $r=5$ by default following~\cite{cheng2021stcn}.
	Optionally, we also pretrain on BL30K~\cite{cheng2021mivos,denninger2019blenderproc,shapenet2015} which gives a further boost in accuracy. We label any method that uses BL30K with an asterisk ($^\ast$).
	
	We use bootstrapped cross entropy loss and dice loss with equal weighting following~\cite{yang2021associating}.
	For optimization, we use AdamW~\cite{kingma2015adam,loshchilov2017decoupled} with a learning rate of 1e-5 and a weight decay of 0.05, for 150K iterations with batch size 16 in static image pretraining, and for 110K iterations with batch size 8 in main training. We drop the learning rate by a factor of 10 after the first 80K iterations. 
	For a fair comparison, we also retrain the STCN~\cite{cheng2021stcn} baseline with the above setting. There is no significant difference in performance for STCN (see appendix).
	
	\section{Experiments}
Unless otherwise specified, we use $T_{\min}=5$, $T_{\max}=10$, and $P=128$, resulting in a compression ratio of $6328\%$ from  working memory to  long-term memory.
We set the maximum number of long-term memory elements to be 10,000 which means XMem never consumes more than 1.4GB of GPU memory, possibly enabling applications even on mobile devices. 
We use top-$k$ filtering~\cite{cheng2021mivos} with $k=30$. 480p videos are used by default.
To evaluate we use standard metrics (higher is better)~\cite{perazzi2016benchmark}: Jaccard index \mj, contour accuracy \mf, and their average \mjf. For YouTubeVOS~\cite{xu2018youtubeVOS}, \mj~and \mf~are computed for ``seen'' and ``unseen'' classes separately, denoted by subscripts $S$ and $U$ respectively. \mg~is averaged \mjf~for both seen and unseen classes.
For AOT~\cite{yang2021associating}, we compare with their R50 variant which has the same ResNet backbone as ours.

\vspace{-0.2em}
	\subsection{Long-Time Video Dataset}\label{sec:expr-long-vid}
	To evaluate long-term performance, we test models on the Long-time Video dataset~\cite{Liang2020AFBURR} which contains three videos with more than 7,000 frames in total. 
	We also synthetically extend it to even longer variants by playing the video back and forth. 
	$n\times$ denotes a variant that has $n$ times the number of frames.
	For comparison, we select state-of-the-art methods with available implementation as we need to re-run their models. 
	Most SOTA methods cannot handle long videos natively.
	We first measure their GPU memory increase per frame by averaging the memory consumption difference between the 100-th and 200-th frame in 480p.\footnote{We make sure to exclude any caching or input buffering overhead.} 
	Figure~\ref{fig:scaling} (left) shows our findings, assuming 24FPS. 
	For methods with prohibitive memory usage on long videos, we limit their feature memory insertion frequency accordingly, using 50 memory frames in STM as a baseline following~\cite{Liang2020AFBURR}.
	Our method uses less memory than this baseline.
	We note that a low memory insertion frequency leads to high variances in performance, thus we run these experiments with 5 evenly-spaced offsets to the memory insertion routine and show ``mean $\pm$ standard deviation'' if applicable.
    In this dataset, we use $r=10$. We do not find BL30K~\cite{cheng2021mivos} pretraining to help here.
	
	\begin{table}
		\centering
		\caption{Quantitative comparisons on the Long-time Video dataset~\cite{Liang2020AFBURR}. }
		\begin{tabular}{l@{\hspace{1pt}}c@{\hspace{0pt}}l@{\hspace{3pt}}c@{\hspace{0pt}}l@{\hspace{3pt}}c@{\hspace{0pt}}l@{\hspace{3pt}}c@{\hspace{0pt}}l@{\hspace{3pt}}c@{\hspace{0pt}}l@{\hspace{3pt}}c@{\hspace{0pt}}l@{\hspace{3pt}}c}
	\toprule
	& \multicolumn{6}{c}{Long-time Video (1$\times$)}  & \multicolumn{6}{c}{Long-time Video (3$\times$)} & $\mathrm{\Delta}_{\text{1$\times$}\to \text{3$\times$}}$\\
	\cmidrule(lr){2-7} \cmidrule(lr){8-13} \cmidrule(lr){14-14}
	Method & \multicolumn{2}{c}{\mjf} & \multicolumn{2}{c}{\mj} & \multicolumn{2}{c}{\mf} & \multicolumn{2}{c}{\mjf} & \multicolumn{2}{c}{\mj} & \multicolumn{2}{c}{\mf} & \mjf \\
	\midrule
	CFBI+~\cite{yang2020CFBIP} & 50.9 &  & 47.9 &  & 53.8 &  & 55.3 &  & 54.0 &  & 56.5 &  & \textcolor{gray}{4.4} \\ 
	RMNet~\cite{xie2021efficient} & 59.8 & {\scriptsize$\pm$3.9} & 59.7 & {\scriptsize$\pm$8.3} & 60.0 & {\scriptsize$\pm$7.5} & 57.0 & {\scriptsize$\pm$1.6} & 56.6 & {\scriptsize$\pm$1.5} & 57.3 & {\scriptsize$\pm$1.8} & \textcolor{red}{-2.8} \\ 
	JOINT~\cite{mao2021joint} & 67.1 & {\scriptsize$\pm$3.5} & 64.5 & {\scriptsize$\pm$4.2} & 69.6 & {\scriptsize$\pm$3.9} & 57.7 & {\scriptsize$\pm$0.2} & 55.7 & {\scriptsize$\pm$0.3} & 59.7 & {\scriptsize$\pm$0.2} & \textcolor{red}{-9.4} \\ 
	CFBI~\cite{yang2020collaborativeCFBI} & 53.5 &  & 50.9 &  & 56.1 &  & 58.9 &  & 57.7 &  & 60.1 &  & \textcolor{gray}{5.4} \\ 
	HMMN~\cite{seong2021hierarchical} & 81.5 & {\scriptsize$\pm$1.8} & 79.9 & {\scriptsize$\pm$1.2} & 83.0 & {\scriptsize$\pm$1.5} & 73.4 & {\scriptsize$\pm$3.3} & 72.6 & {\scriptsize$\pm$3.1} & 74.3 & {\scriptsize$\pm$3.5} & \textcolor{red}{-8.1} \\ 
	STM~\cite{oh2019videoSTM} & 80.6 & {\scriptsize$\pm$1.3} & 79.9 & {\scriptsize$\pm$0.9} & 81.3 & {\scriptsize$\pm$1.0} & 75.3 & {\scriptsize$\pm$13.0} & 74.3 & {\scriptsize$\pm$13.0} & 76.3 & {\scriptsize$\pm$13.1} & \textcolor{red}{-5.3} \\ 
	MiVOS$^{\ast}$~\cite{cheng2021mivos} & 81.1 & {\scriptsize$\pm$3.2} & 80.2 & {\scriptsize$\pm$2.0} & 82.0 & {\scriptsize$\pm$3.1} & 78.5 & {\scriptsize$\pm$4.5} & 78.0 & {\scriptsize$\pm$3.7} & 79.0 & {\scriptsize$\pm$5.4} & \textcolor{red}{-2.6} \\ 
	AOT~\cite{yang2021associating} & 84.3 & {\scriptsize$\pm$0.7} & 83.2 & {\scriptsize$\pm$3.2} & 85.4 & {\scriptsize$\pm$3.3} & 81.2 & {\scriptsize$\pm$2.5} & 79.6 & {\scriptsize$\pm$3.0} & 82.8 & {\scriptsize$\pm$2.1} & \textcolor{red}{-3.1} \\ 
	AFB-URR~\cite{Liang2020AFBURR} & 83.7 &  & 82.9 &  & 84.5 &  & 83.8 &  & 82.9 &  & 84.6 &  & \textcolor{gray}{0.1} \\ 
	STCN~\cite{cheng2021stcn} & 87.3 & {\scriptsize$\pm$0.7} & 85.4 & {\scriptsize$\pm$1.1} & 89.2 & {\scriptsize$\pm$1.1} & 84.6 & {\scriptsize$\pm$1.9} & 83.3 & {\scriptsize$\pm$1.7} & 85.9 & {\scriptsize$\pm$2.2} & \textcolor{red}{-2.7} \\ 
	XMem~(Ours) & \textbf{89.8} & {\scriptsize$\pm$0.2} & \textbf{88.0} & {\scriptsize$\pm$0.2} & \textbf{91.6} & {\scriptsize$\pm$0.2} & \textbf{90.0} & {\scriptsize$\pm$0.4} & \textbf{88.2} & {\scriptsize$\pm$0.3} & \textbf{91.8} & {\scriptsize$\pm$0.4} & \textcolor{gray}{0.2} \\ 
	\midrule
	\bottomrule
\end{tabular}
		\label{tab:long-time}
	\end{table}

	Table~\ref{tab:long-time} tabulates the quantitative results, and Figure~\ref{fig:scaling} (right) plots the short-term performance against the long-term performance.
	Methods that use a temporally local feature window (CFBI(+)~\cite{yang2020collaborativeCFBI,yang2020CFBIP}, JOINT~\cite{mao2021joint}) have a constant memory cost but fail when they lose track of the context.
	Methods with a fast-growing memory bank (e.g., STM~\cite{oh2019videoSTM}, AOT~\cite{yang2021associating}, STCN~\cite{cheng2021stcn}) are forced to use a low feature memory insertion frequency and do not scale well to long videos.
	Figure~\ref{fig:linear-reg} shows the scaling behavior of STCN vs.\ XMem in more detail.
	

	AFB-URR~\cite{Liang2020AFBURR} is designed to handle long videos and scales well with no degradation -- but  due to eager feature compression it has relatively low performance in the short term compared to other methods.
	In contrast, XMem not only holds up well in scaling to longer videos but also performs well in the short-term as shown in the next section.
	We provide qualitative comparisons in the appendix.
	
	\subsection{Short Video Datasets}
	\vspace{-1em}
	Table~\ref{tab:y18-d17} and Table~\ref{tab:d17-td} tabulate our result on YouTubeVOS~\cite{xu2018youtubeVOS} 2018 validation, DAVIS~\cite{perazzi2016benchmark} 2016/2017 validation, and DAVIS 2017~\cite{Pont-Tuset_arXiv_2017} test-dev. Results on YouTubeVOS~\cite{xu2018youtubeVOS} 2019 validation can be found in the appendix.
	The test set for YouTubeVOS is closed at the time of writing. We use $r=5$ for these datasets.
	Following standard practice~\cite{oh2019videoSTM,yang2020collaborativeCFBI,cheng2021stcn}, we report single/multi-object FPS on DAVIS 2016/2017 validation.
	We additionally report FPS on YouTubeVOS 2018 validation which has longer videos on average.
	We measure FPS on a V100 GPU. For a fair comparison, we re-time prior works that report FPS on a slower GPU if possible and label this with a $\dagger$. 
	We note that some methods (not ours) are faster on a 2080Ti than on a V100. 
	In these cases, we always give competing methods the benefit.
	Our speed-up solely comes from the use of long-term memory -- a compact feature memory representation is faster to read from.
	
	\vspace{-2em}

	\begin{table}[h]
		\centering
		\scriptsize
		\caption{Quantitative comparisons on three commonly used short-term datasets. $^\ast$~denotes BL30K~\cite{cheng2021mivos} pretraining. Bold and underline denote the best and the second-best respectively in each column. $\dagger$~denotes FPS re-timed on our hardware.
		On YouTubeVOS, we re-run AOT with all input frames (improving its performance) for a fair comparison.}
		\begin{tabular}{l@{\hspace{3pt}} c@{\hspace{3pt}}c@{\hspace{3pt}}c@{\hspace{3pt}}c@{\hspace{3pt}}c@{\hspace{3pt}}c@{\hspace{10pt}} c@{\hspace{3pt}}c@{\hspace{3pt}}c@{\hspace{3pt}}c@{\hspace{10pt}} c@{\hspace{3pt}}c@{\hspace{3pt}}c@{\hspace{3pt}}c}
	\toprule
	& \multicolumn{6}{c}{YT-VOS 2018 val~\cite{xu2018youtubeVOS}}  & \multicolumn{4}{c}{DAVIS 2017 val~\cite{Pont-Tuset_arXiv_2017}} & \multicolumn{4}{c}{DAVIS 2016 val~\cite{perazzi2016benchmark}} \\
	\cmidrule(lr{\dimexpr 4\tabcolsep+5pt}){2-7} \cmidrule(lr{\dimexpr 4\tabcolsep+5pt}){8-11} \cmidrule(lr){12-15}
	Method & \mg & \mjs & \mfs & \mju & \mfu & FPS & \mjf & \mj & \mf & FPS & \mjf & \mj & \mf & FPS \\
	\midrule
	STM~\cite{oh2019videoSTM} & 79.4 & 79.7 & 84.2 & 72.8 & 80.9 & - & 81.8 & 79.2 & 84.3 & 11.1$\dagger$ & 89.3 & 88.7 & 89.9 & 14.0$\dagger$ \\
	AFB-URR~\cite{Liang2020AFBURR} & 79.6 & 78.8 & 83.1 & 74.1 & 82.6 & - & 76.9 & 74.4 & 79.3 & 6.8$\dagger$ & - & - & - & - \\
	CFBI~\cite{yang2020collaborativeCFBI} & 81.4 & 81.1 & 85.8 & 75.3 & 83.4 & 3.4 & 81.9 & 79.1 & 84.6 & 5.9 & 89.4 & 88.3 & 90.5 & 6.2 \\
	RMNet~\cite{xie2021efficient} & 81.5 & 82.1 & 85.7 & 75.7 & 82.4 & - & 83.5 & 81.0 & 86.0 & 4.4$\dagger$ & 88.8 & 88.9 & 88.7 & 11.9 \\
	HMMN~\cite{seong2021hierarchical} & 82.6 & 82.1 & 87.0 & 76.8 & 84.6 & - & 84.7 & 81.9 & 87.5 & 9.3$\dagger$ & 90.8 & 89.6 & 92.0 & 13.0$\dagger$ \\
	MiVOS$^{\ast}$~\cite{cheng2021mivos} & 82.6 & 81.1 & 85.6 & 77.7 & 86.2 & - & 84.5 & 81.7 & 87.4 & 11.2 & 91.0 & 89.6 & 92.4 & 16.9 \\
	STCN~\cite{cheng2021stcn} & 83.0 & 81.9 & 86.5 & 77.9 & 85.7 & \underline{13.2}$\dagger$ & 85.4 & 82.2 & 88.6 & \underline{20.2}$\dagger$ & 91.6 & \textbf{90.8} & 92.5 & \underline{26.9}$\dagger$ \\
	JOINT~\cite{mao2021joint} & 83.1 & 81.5 & 85.9 & 78.7 & 86.5 & - & 83.5 & 80.8 & 86.2 & 6.8$\dagger$ & - & - & - & - \\
	STCN$^{\ast}$~\cite{cheng2021stcn} & 84.3 & 83.2 & 87.9 & 79.0 & 87.3 & \underline{13.2}$\dagger$ & 85.3 & 82.0 & 88.6 & \underline{20.2}$\dagger$ & \underline{91.7} & 90.4 & \underline{93.0} & \underline{26.9}$\dagger$ \\
	AOT~\cite{yang2021associating} & 85.5 & 84.5 & \underline{89.5} & 79.6 & 88.2 & 6.4 & 84.9 & 82.3 & 87.5 & 18.0 & 91.1 & 90.1 & 92.1 & 18.0 \\
	XMem~(Ours) & \underline{85.7} & \underline{84.6} & 89.3 & \underline{80.2} & \underline{88.7} & \textbf{22.6} & \underline{86.2} & \underline{82.9} & \underline{89.5} & \textbf{22.6} & 91.5 & 90.4 & 92.7 & \textbf{29.6} \\
	XMem$^{\ast}$~(Ours) & \textbf{86.1} & \textbf{85.1} & \textbf{89.8} & \textbf{80.3} & \textbf{89.2} & \textbf{22.6} & \textbf{87.7} & \textbf{84.0} & \textbf{91.4} & \textbf{22.6} & \textbf{92.0} & \underline{90.7} & \textbf{93.2} & \textbf{29.6} \\
	\midrule
	\bottomrule
\end{tabular}

		\label{tab:y18-d17}
		\vspace{-0.5em}
	\end{table}

	\begin{minipage}{\textwidth}
	\vspace{-0.5em}
		\hspace{-15pt}
		\begin{minipage}[c]{0.59\textwidth}
			\centering
			\resizebox{!}{0.5\textwidth}{
				\begin{tikzpicture}
	\pgfplotsset{
		width=11cm,
		height=5cm,
		compat=1.13,
		legend style={font=\footnotesize}}
	\begin{axis}[
		scale only axis = true, enlargelimits=false,
		xmin=2, xmax=37,
		ymin=70,ymax=95,
		grid=both,
		xlabel={Number of frames (K)},
		ylabel={\mjf},
		legend cell align=left,
		legend pos=south west]
		
		\addplot[only marks,mark=*,color=red,mark size=1.5pt] table[row sep=\\]{
			X Y\\
			2.406 90.3\\
			1.416 88\\
			3.589 91.1\\
			4.812 90.6\\
			2.832 88.4\\
			7.178 91\\
			7.218 90.9\\
			4.248 85\\
			10.767 86.1\\
			9.624 90\\
			5.664 90\\
			14.356 91.1\\
			12.03 90.9\\
			7.08 85.2\\
			17.945 91.2\\
			14.436 90.9\\
			8.496 89.9\\
			21.534 91.1\\
			16.842 90.9\\
			9.912 90.4\\
			25.123 91.2\\
			19.248 90.8\\
			11.328 90\\
			28.712 91.1\\
			21.654 90.9\\
			12.744 90.3\\
			32.301 91.2\\
			24.06 90.8\\
			14.16 90.1\\
			35.89 91.1\\
		};
		\addlegendentry{XMem (Ours)}
		\addplot+[mark=None,color=red,forget plot] table[row sep=\\,
		y={create col/linear regression={y=Y}}] 
		{
			X Y\\
			2.406 90.3\\
			1.416 88\\
			3.589 91.1\\
			4.812 90.6\\
			2.832 88.4\\
			7.178 91\\
			7.218 90.9\\
			4.248 85\\
			10.767 86.1\\
			9.624 90\\
			5.664 90\\
			14.356 91.1\\
			12.03 90.9\\
			7.08 85.2\\
			17.945 91.2\\
			14.436 90.9\\
			8.496 89.9\\
			21.534 91.1\\
			16.842 90.9\\
			9.912 90.4\\
			25.123 91.2\\
			19.248 90.8\\
			11.328 90\\
			28.712 91.1\\
			21.654 90.9\\
			12.744 90.3\\
			32.301 91.2\\
			24.06 90.8\\
			14.16 90.1\\
			35.89 91.1\\
		};
		\xdef\slopeA{\pgfplotstableregressiona} 
		\xdef\interceptA{\pgfplotstableregressionb}
		\addplot+[no marks,red,domain=0:1.5,forget plot]{\slopeA*x+\interceptA};
		\addplot+[no marks,red,domain=35:38,forget plot]{\slopeA*x+\interceptA};
		
		\addplot[only marks,mark=diamond*,color=blue] table[row sep=\\]{
			X Y\\
			2.406 88\\
			1.416 77.2\\
			3.589 84\\
			4.812 86.8\\
			2.832 80.2\\
			7.178 83.3\\
			7.218 86.4\\
			4.248 78.5\\
			10.767 85.1\\
			9.624 86.4\\
			5.664 75.4\\
			14.356 83.5\\
			12.03 73.6\\
			7.08 87.1\\
			17.945 82.6\\
			14.436 85.7\\
			8.496 80.9\\
			21.534 82\\
			16.842 45.9\\
			9.912 81.7\\
			25.123 81.9\\
			19.248 83.9\\
			11.328 78.6\\
			28.712 70.3\\
			21.654 80.1\\
			12.744 74\\
			32.301 77.7\\
			24.06 85\\
			14.16 78.6\\
			35.89 83.8\\
		};
		\addlegendentry{STCN}
		\addplot+[mark=None,color=blue,forget plot] table[row sep=\\,
		y={create col/linear regression={y=Y}}] 
		{
			X Y\\
			2.406 88\\
			1.416 77.2\\
			3.589 84\\
			4.812 86.8\\
			2.832 80.2\\
			7.178 83.3\\
			7.218 86.4\\
			4.248 78.5\\
			10.767 85.1\\
			9.624 86.4\\
			5.664 75.4\\
			14.356 83.5\\
			12.03 73.6\\
			7.08 87.1\\
			17.945 82.6\\
			14.436 85.7\\
			8.496 80.9\\
			21.534 82\\
			16.842 45.9\\
			9.912 81.7\\
			25.123 81.9\\
			19.248 83.9\\
			11.328 78.6\\
			28.712 70.3\\
			21.654 80.1\\
			12.744 74\\
			32.301 77.7\\
			24.06 85\\
			14.16 78.6\\
			35.89 83.8\\
		};
	\xdef\slopeB{\pgfplotstableregressiona} 
	\xdef\interceptB{\pgfplotstableregressionb}
	\addplot+[no marks,blue,domain=0:1.5,forget plot]{\slopeB*x+\interceptB};
	\addplot+[no marks,blue,domain=35:38,forget plot]{\slopeB*x+\interceptB};
	\end{axis}
\end{tikzpicture}

			}
			\vspace{-2.2em}
			\captionof{figure}{Least-square fits of performance over video length for XMem and STCN~\cite{cheng2021stcn} on variants of the Long-time Video dataset~\cite{Liang2020AFBURR} from $1\times$ to $10\times$.
			In longer videos, STCN decays due to missing context while ours stabilizes as we gain sufficient context.}
			\label{fig:linear-reg}
		\end{minipage}
		\hspace{5pt}
		\begin{minipage}[c]{0.30\textwidth}
			\vspace{0em}
			\centering
			\scriptsize
			\captionof{table}{Results on DAVIS 2017 test-dev. $\ddagger$:~uses 600p videos.}
			\vspace{-1em}
			\begin{tabular}{lccc}
	\toprule
	& \multicolumn{3}{c}{DAVIS 2017 td} \\
	\cmidrule(lr{\dimexpr 4\tabcolsep+5pt}){2-4} 
	Method & \mjf & \mj & \mf \\
	\midrule
	STM$\ddagger$~\cite{oh2019videoSTM} & 72.2 & 69.3 & 75.2 \\
	RMNet~\cite{xie2021efficient} & 75.0 & 71.9 & 78.1 \\
	STCN~\cite{cheng2021stcn} & 76.1 & 73.1 & 80.0 \\
	CFBI+$\ddagger$~\cite{yang2020CFBIP} & 78.0 & 74.4 & 81.6 \\
	HMMN~\cite{seong2021hierarchical} & 78.6 & 74.7 & 82.5 \\
	MiVOS$^{\ast}$~\cite{cheng2021mivos} & 78.6 & 74.9 & 82.2 \\
	AOT~\cite{yang2021associating} & 79.6 & 75.9 & 83.3 \\
	STCN$^{\ast}$~\cite{cheng2021stcn} & 79.9 & 76.3 & 83.5 \\
	XMem~(Ours) & {81.0} & {77.4} & {84.5} \\
	XMem$^{\ast}$~(Ours) & \underline{81.2} & \underline{77.6} & \underline{84.7} \\
	XMem$^{\ast}$$\ddagger$~(Ours) & \textbf{82.5} & \textbf{79.1} & \textbf{85.8} \\
	\midrule
	\bottomrule
\end{tabular}
			\label{tab:d17-td}
		\end{minipage}
	\end{minipage}
	
	\begin{minipage}{\textwidth}
		\vspace{0em}
		\hspace{-15pt}
		\begin{minipage}{0.55\textwidth}
			\scriptsize
			\centering
			\captionof{table}{Ablation on our memory stores. Standard deviations for $L_{1\times}$ are omitted.}
			\vspace{-11pt}
			\begin{tabular}{lccccc}
	\toprule
	Setting & Y$_{\text{18}}$ & D$_{\text{17}}$ & L$_{1\times}$ & FPS$_{\text{D17}}$ & FPS$_{\text{Y18}}$\\
	\midrule
	\rowcolor{LightCyan}
	All memory stores & 85.7 & 86.2 & \textbf{89.8} & 22.6 & 22.6 \\
	No sensory memory & 84.4 & 85.1 & 87.9 & 23.1 & 23.1 \\
	No working memory & 72.7 & 77.6 & 38.7 & \textbf{31.8} & \textbf{28.1} \\
	No long-term memory & \textbf{85.9} & \textbf{86.3} & n/a & 17.6 & 10.0 \\
	\midrule
	\bottomrule
\end{tabular}
			\label{tab:diff-memory}
		\end{minipage}
		\hspace{10pt}
		\begin{minipage}{0.38\textwidth}
			\scriptsize
			\centering
			\captionof{table}{Ablation on the two scaling terms in memory reading.}
			\vspace{-11pt}
			\begin{tabular}{lcc}
	\toprule
	Setting & ~Y$_{\text{18}}$~ & D$_{\text{17}}$ \\
	\midrule
	\rowcolor{LightCyan}
	With both terms & \textbf{85.7} & \textbf{86.2} \\
	With shrinkage $\mathbf{s}$ only & 85.1 & 85.6 \\
	With selection $\mathbf{e}$ only & 84.8 & 84.8 \\
	With neither & 85.0 & 85.1 \\
	\midrule
	\bottomrule
\end{tabular}
			\label{tab:diff-mem-read}
		\end{minipage}
	\vspace{-0.8em}
	\end{minipage}
	
	\subsection{Ablations}
	We perform ablation studies on validation sets of YouTubeVOS 2018~\cite{xu2018youtubeVOS} (Y$_\text{18}$), DAVIS 2017~\cite{Pont-Tuset_arXiv_2017} (D$_{\text{17}}$), and Long-time Video ($n\times$)~\cite{Liang2020AFBURR} (L$_{n\times}$). We report the most representative metric (\mg~for YouTubeVOS, \mjf~for DAVIS/Long-time Video).
	FPS is measured on DAVIS 2017 validation unless otherwise specified.
	We highlight our final configuration with \colorbox{LightCyan}{cyan}.
	
	\noindent\textbf{Memory Stores.}
	Table~\ref{tab:diff-memory} tabulates the performance of XMem without any one of the memory stores.
	If the working memory is removed, long-term memory cannot function and it becomes ``sensory memory only'' with a constant memory cost.
	If the long-term memory is removed, all the memory frames are stored in the working memory. Although it has a slightly better performance due to its higher resolution feature, it cannot handle long videos and is slower.
	
	\noindent\textbf{Memory Reading.}
	Table~\ref{tab:diff-mem-read} shows the importance of the two scaling terms in the anisotropic L2 similarity. 
	Interestingly, the selection term $\mathbf{e}$ alone does not help. 
	We hypothesize that the selection term allows attention on a different subset of memory elements for every query, thus increasing the relative importance of each memory element. The shrinkage term $\mathbf{s}$ allows element-level modulation of confidence, thus avoiding too much emphasis on less confident elements.
	There is a synergy between the two terms, and our final model benefits from both.
	
	\noindent\textbf{Long-term Memory Strategies.}
	Table~\ref{tab:diff-long-term} compares different prototype selection strategies and shows the importance of potentiation.
	We run all algorithms 5 times with evenly-spaced memory insertion offsets and show  standard deviations.
	We choose the usage-based selection scheme with $P=128$ for a balance between performance and memory compression.
	Table~\ref{tab:diff-discard} compares additional strategies used by prior works, employed on our model. 
	Eager compression is inspired by AFB-URR~\cite{Liang2020AFBURR}. We set $T_{\min}=1$ and $T_{\max}=2$. Note, since we cannot compute usage statistics in this setting, we use random prototype selection with the same compression ratio.
	Sparse insertion follows our treatment to methods with a growing memory bank~\cite{oh2019videoSTM,cheng2021stcn}. We set the maximum number of memory frames to be 50 following~\cite{Liang2020AFBURR}.
	Local window follows~\cite{yang2020collaborativeCFBI,duke2021sstvos,mao2021joint}, where we simply discard the oldest memory frame when the memory bank reaches its capacity. We always keep the first reference frame and set the memory bank capacity to be 50.
	Our memory consolidation algorithm is the most effective among these.

	\noindent\textbf{Deep Update.}
	Table~\ref{tab:diff-deep-update} shows different configurations of the deep update.
	Employing deep update every $r$-th frame results in a performance boost, with no noticeable speed drop (recall that we have to use the value encoder every $r$-th frame for our working memory anyway).
	However, using deep updates more often requires extra invocations of the value encoder and leads to a  slowdown.
	
	\begin{table}[t]
		\vspace{-1em}
		\hspace{-0pt}
		\begin{minipage}{0.5\textwidth}
			\scriptsize
			\centering
			\captionof{table}{Comparisons between different memory consolidation methods. }
			\begin{tabular}{l@{\hspace{1pt}}lc@{\hspace{1pt}}lc}
	\toprule
	& & & & Compress\\
	Setting & & \multicolumn{2}{c}{L$_{3\times}$} & ratio \\
	\midrule
	Random & $P=64$ & 89.5 &$\pm0.8$ & \textbf{12625\%} \\
	K-means centroid & $P=64$ & 89.5 &$\pm0.5$ & \textbf{12625\%}\\
	Usage-based & $P=64$ & \textbf{89.6}&$\pm0.4$ & \textbf{12625\%}\\
	\midrule
	Random & $P=128$ & 89.7&$\pm0.7$ & 6328\% \\
	K-means centroid & $P=128$ & 82.4&$\pm10.3$ & 6328\%\\
	\rowcolor{LightCyan}
	Usage-based & $P=128$ & \textbf{90.0}&$\pm0.4$ & 6328\%\\
	\midrule
	Random & $P=256$ & 89.8&$\pm0.7$ &  3164\% \\
	K-means centroid & $P=256$ & 74.5&$\pm17.0$ & 3164\%\\
	Usage-based & $P=256$ & \textbf{90.1}&$\pm0.4$ & 3164\%\\
	\bottomrule
	No potentiation & & 87.9&$\pm0.2$ & \\
	\rowcolor{LightCyan}
	With potentiation & & \textbf{90.0}&$\pm0.4$ & \\
	\midrule
	\bottomrule
\end{tabular}
			\label{tab:diff-long-term}
		\end{minipage}
		\begin{minipage}{0.5\textwidth}
			\scriptsize
			\centering
			\captionof{table}{Comparisons between different strategies for handling long videos.}			\begin{tabular}{lc@{\hspace{1pt}}lc@{\hspace{1pt}}lc}
	\toprule
	Setting & \multicolumn{2}{c}{L$_{1\times}$} & \multicolumn{2}{c}{L$_{3\times}$} & $\mathrm{\Delta}_{\text{1$\times$}\to \text{3$\times$}}$\\
	\midrule
	\rowcolor{LightCyan}
	Consolidation & \textbf{89.8}&$\pm0.2$ & \textbf{90.0}&$\pm0.4$ & \textcolor{gray}{0.2}\\
	Eager compression & 87.8&$\pm0.3$ & 87.3&$\pm1.3$ & \textcolor{red}{-0.5}\\
	Sparse insertion & \textbf{89.8}&$\pm0.4$ & 87.3&$\pm1.0$ & \textcolor{red}{-2.5}\\
	Local window & 86.2&$\pm1.5$ & 85.5&$\pm0.9$ & \textcolor{red}{-0.7}\\
	\midrule
	\bottomrule
\end{tabular}
			\label{tab:diff-discard}
			
			\vspace{4pt}
			
			\captionof{table}{Ablation on the deep update frequency of sensory memory.}
			\vspace{-8pt}
			\begin{tabular}{lccc}
	\toprule
	Setting & ~Y$_{\text{18}}$~ & D$_{\text{17}}$ & FPS\\
	\midrule
	\rowcolor{LightCyan}
	Every $r$-th frame & \textbf{85.7} & \textbf{86.2} & \textbf{22.6} \\
	Every single frame & 85.5 & 86.1 & 18.5 \\
	No deep update & 85.3 & 85.4 & \textbf{22.6} \\
	\midrule
	\bottomrule
\end{tabular}
			\label{tab:diff-deep-update}
		\end{minipage}
	\vspace{-2em}
	\end{table}

	\noindent\textbf{Pretraining.}
	There are prior works that do not use static image pretraining~\cite{yang2020CFBIP,bhat2020learningLWL,duke2021sstvos,mao2021joint}.
	We provide our results without pretraining in the appendix.
	
	\subsection{Limitations}
	Our method sometimes fails when the target object moves too quickly or has severe motion blur as even the fastest updating sensory memory cannot catch up. See the appendix for examples.
	We think a sensory memory with a large receptive field that is more powerful than our baseline instantiation could help.

	\section{Conclusion}
	We present XMem -- to our best knowledge the first multi-store feature memory model used for video object segmentation.
	XMem achieves excellent performance with minimal GPU memory usage for both long and short videos.
	We believe XMem is a good step toward accessible VOS on mobile devices, and
	we hope to draw attention to the more widely-applicable long-term VOS task.
	
	\noindent\textbf{Acknowledgment.} 
	Work supported in part by NSF under Grants 1718221, 2008387, 2045586, 2106825, MRI 1725729, and NIFA award 2020-67021-32799.

	%
	%
	\bibliographystyle{splncs04}
	\bibliography{xmem}
	
	\newpage
	\begin{center}
	\vspace{1em}
		\Large
		\textbf{Appendix}
	\vspace{1em}
	\end{center}
	\noindent{}The appendix is structured as follows:
	\appendix
\beginsupplement

\begin{itemize}
    \item We first provide a more detailed analysis of the memory consolidation process (Sec.~\ref{sec:app:consol}).
    \item We then provide qualitative results, comparing the proposed XMem to baselines (Sec.~\ref{sec:app:qual}).
    \item We demonstrate failure cases (Sec.~\ref{sec:app:failure}).
    \item We compare different limits on the size of the long-term memory and illustrate the processing rate over the number of processed frames (Sec.~\ref{sec:app:scaling}).
    \item We give quantitative results when retraining the STCN baseline in our training setting (Sec.~\ref{sec:app:retrain}).
    \item We provide results on the YouTubeVOS 2019 validation set (Sec.~\ref{sec:app:yv-2019}).
    \item We provide detailed quantitative results when XMem is trained on different datasets (Sec.~\ref{sec:app:no-pretrain}).
    \item We explain our multi-scale (MS) evaluation methodology and provide the corresponding enhanced performance of XMem (Sec.~\ref{sec:app:multiscale}).
    \item We outline an efficient implementation of the proposed anisotropic L2 similarity function (Sec.~\ref{sec:app:similarity}).
\end{itemize}

\section{Visualizing Memory Consolidation}
\label{sec:app:consol}
\begin{figure}
	\vspace{-1em}
	\centering
	\begin{tabular}{c@{\hspace{.5mm}}c@{\hspace{.5mm}}c@{\hspace{.5mm}}c@{\hspace{.5mm}}c}
	\includegraphics[width=0.19\linewidth]{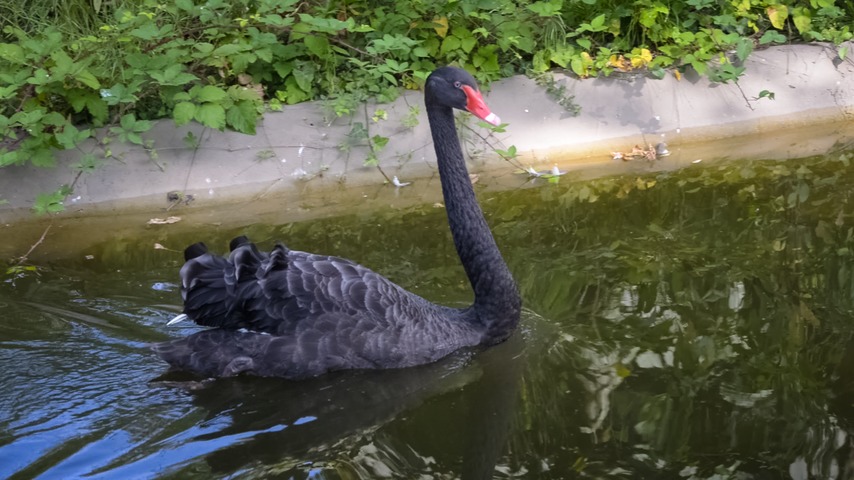} &
	\includegraphics[width=0.19\linewidth]{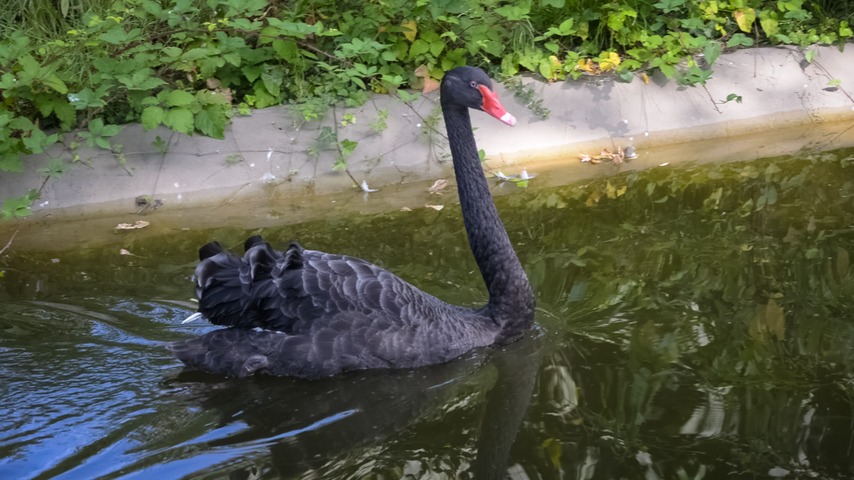} & 
	\includegraphics[width=0.19\linewidth]{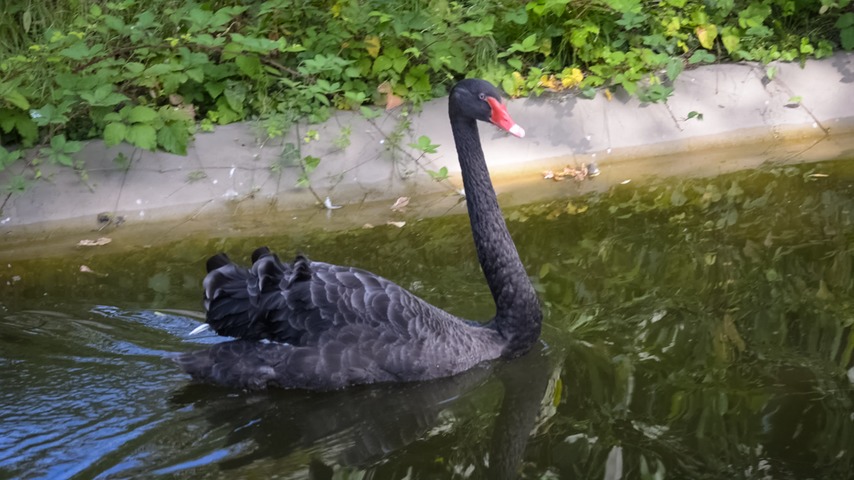} & 
	\includegraphics[width=0.19\linewidth]{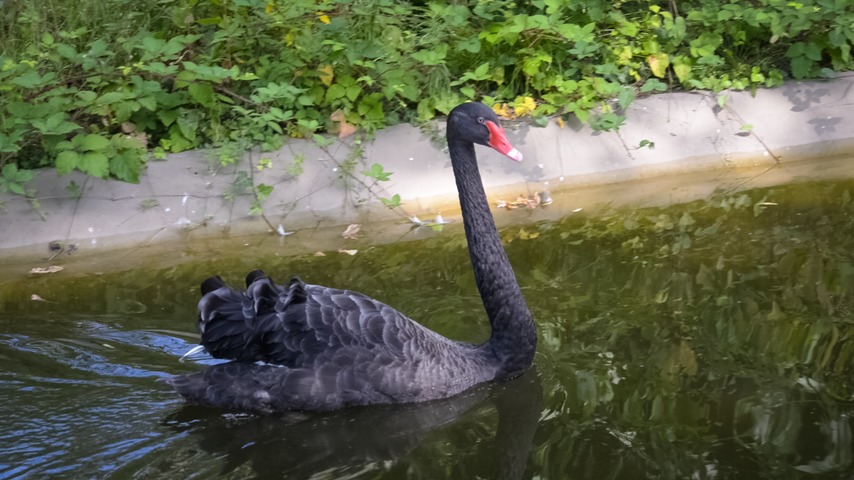} & 
	\includegraphics[width=0.19\linewidth]{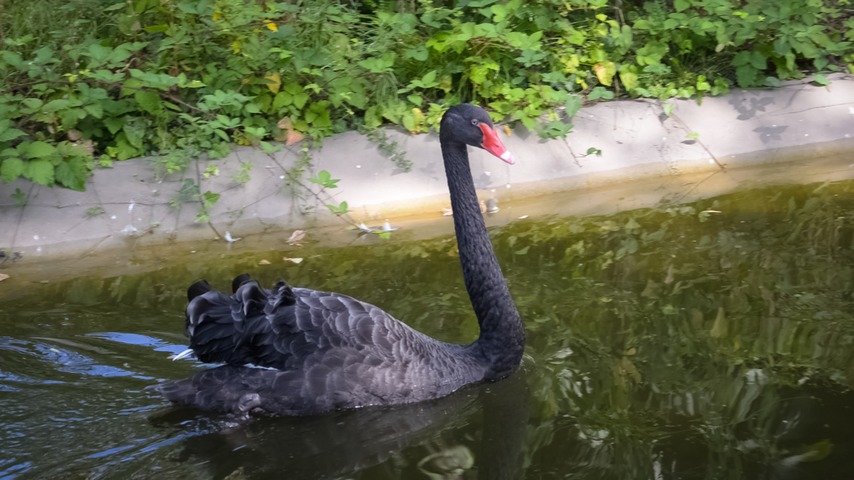} \\
	\vspace{-4.5mm}&&&&\\
	\includegraphics[width=0.19\linewidth,cfbox=red 1pt 0pt]{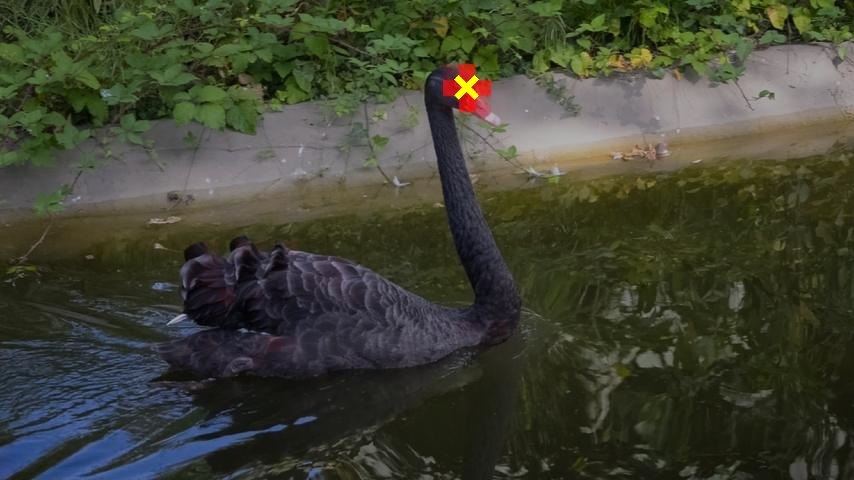} &
	\includegraphics[width=0.19\linewidth]{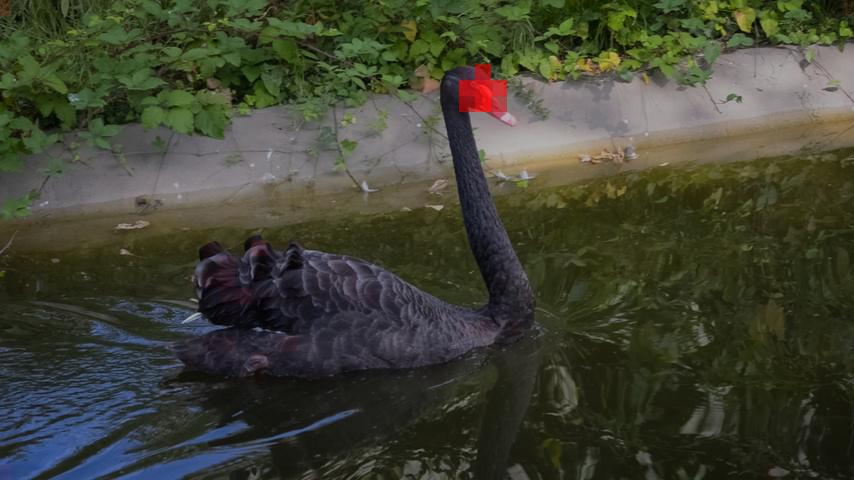} & 
	\includegraphics[width=0.19\linewidth]{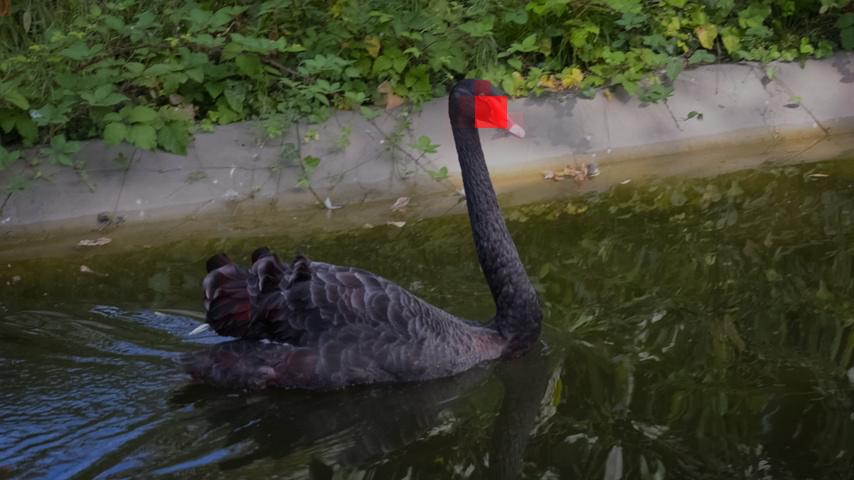} & 
	\includegraphics[width=0.19\linewidth]{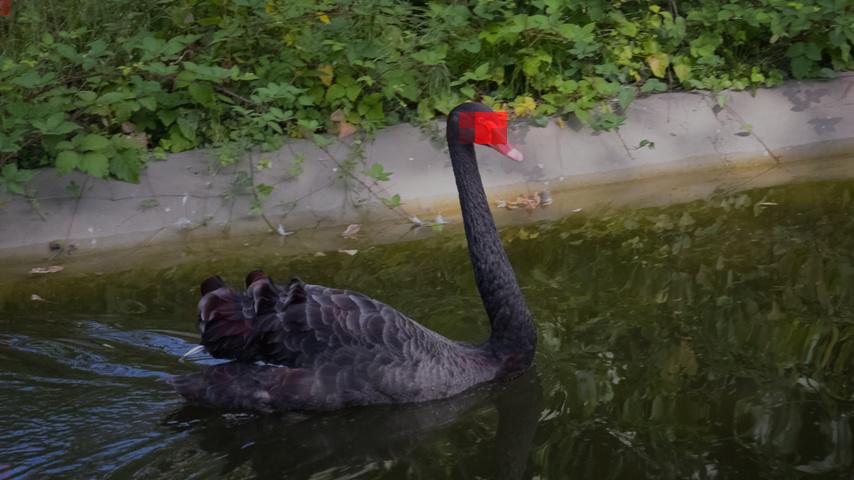} & 
	\includegraphics[width=0.19\linewidth]{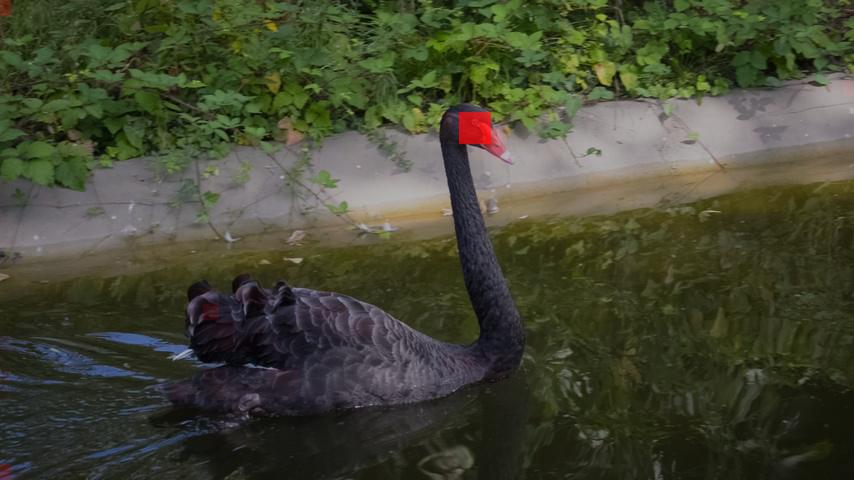} \\
	\vspace{-4.5mm}&&&&\\
	\includegraphics[width=0.19\linewidth]{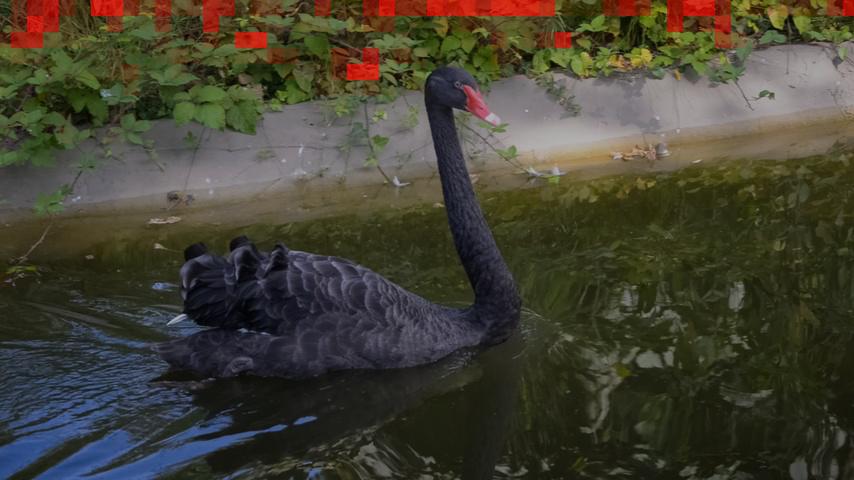} &
	\includegraphics[width=0.19\linewidth]{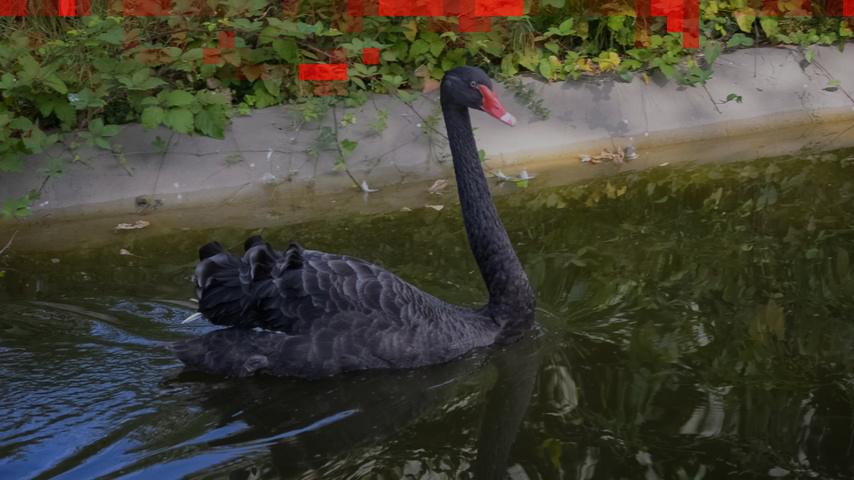} & 
	\includegraphics[width=0.19\linewidth]{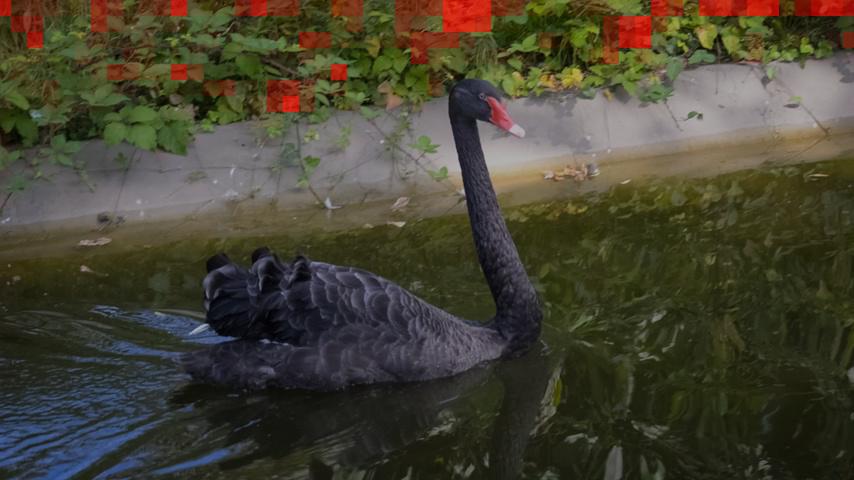} & 
	\includegraphics[width=0.19\linewidth,cfbox=red 1pt 0pt]{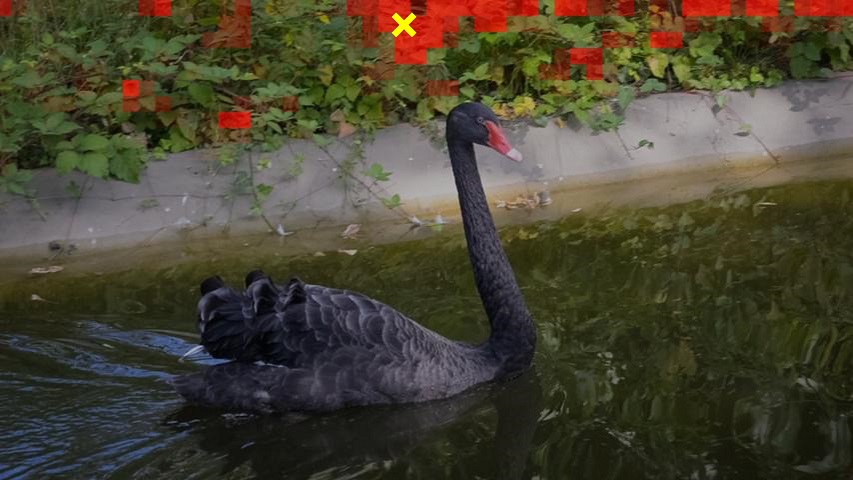} & 
	\includegraphics[width=0.19\linewidth]{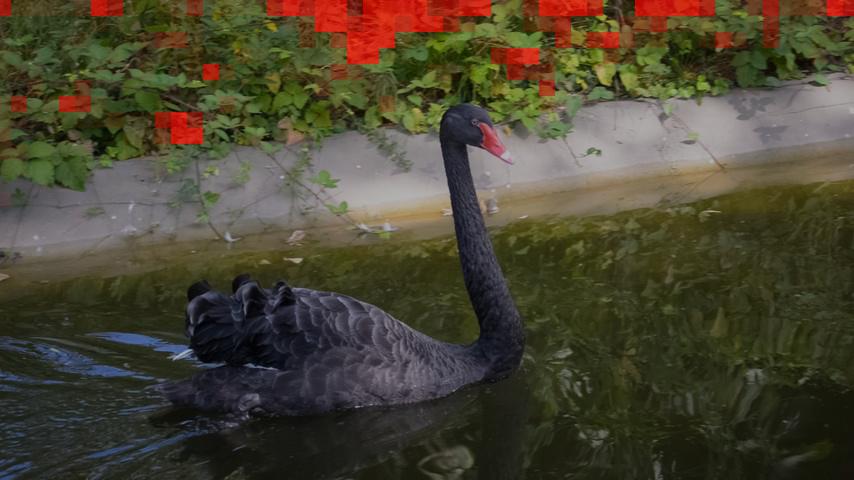} \\
	\vspace{-4.5mm}&&&&\\
	\includegraphics[width=0.19\linewidth]{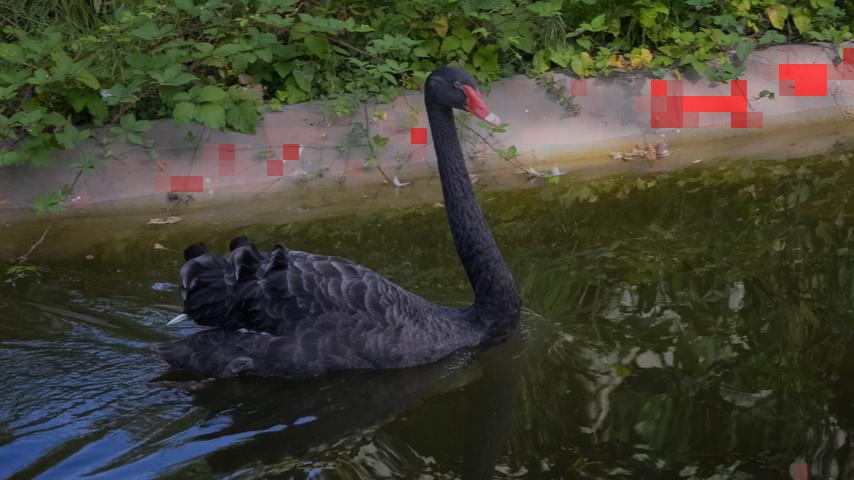} &
	\includegraphics[width=0.19\linewidth]{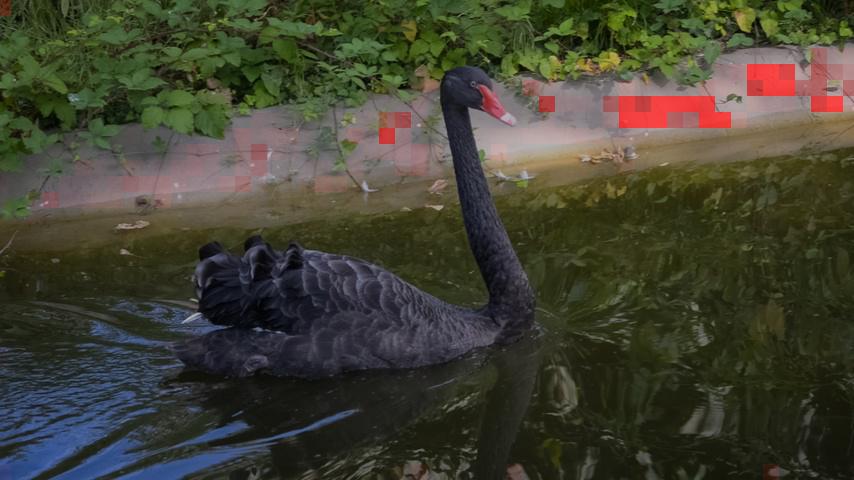} & 
	\includegraphics[width=0.19\linewidth]{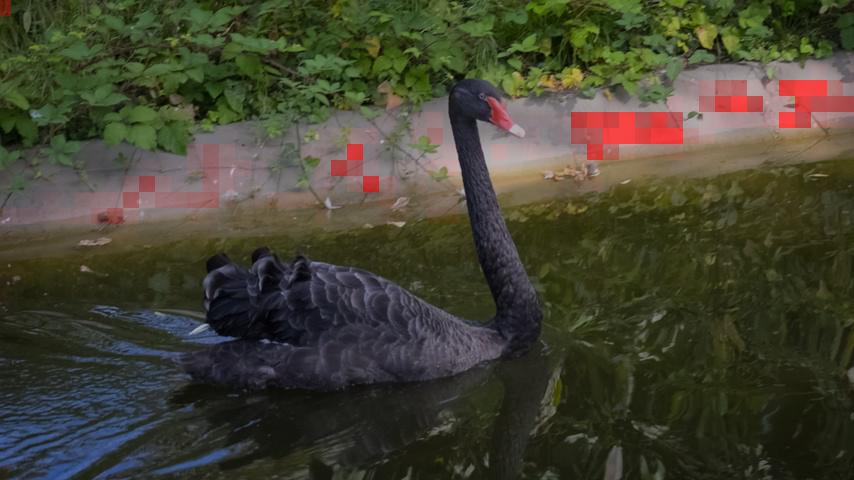} & 
	\includegraphics[width=0.19\linewidth]{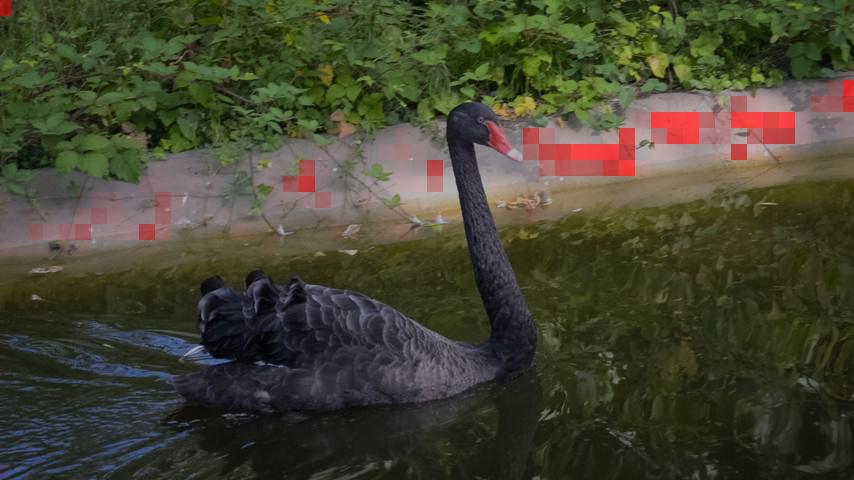} & 
	\includegraphics[width=0.19\linewidth,cfbox=red 1pt 0pt]{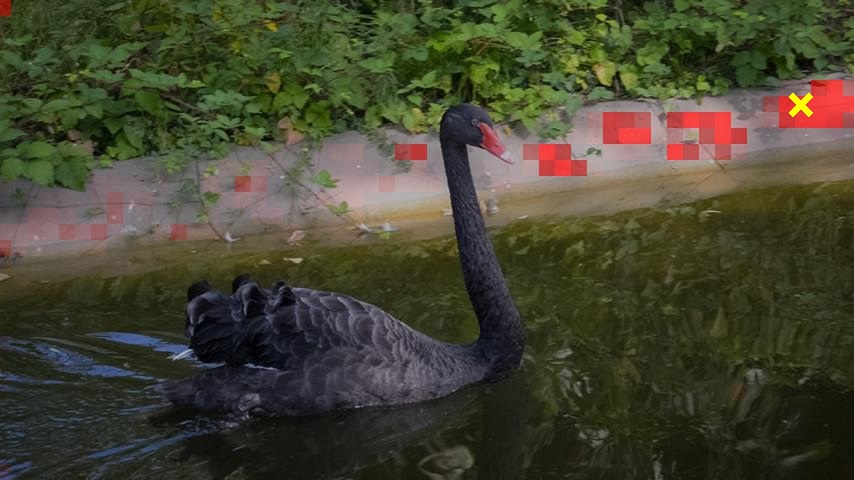} \\
	\vspace{-4.5mm}&&&&\\
	\includegraphics[width=0.19\linewidth]{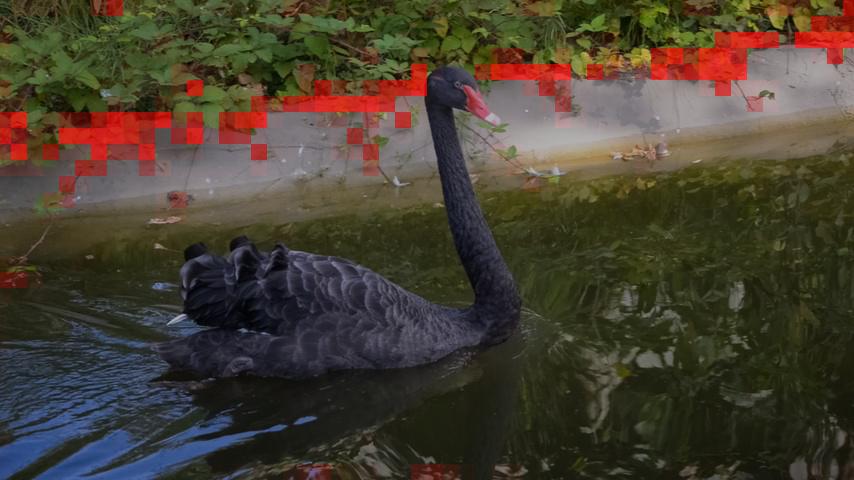} &
	\includegraphics[width=0.19\linewidth]{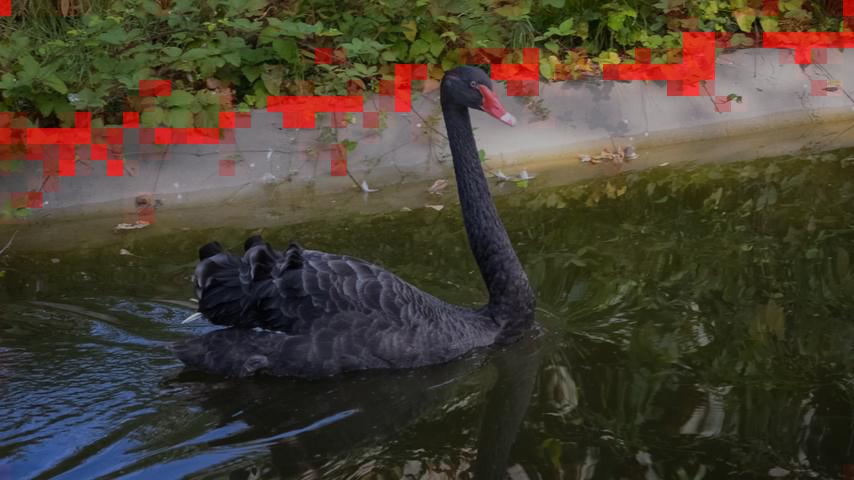} & 
	\includegraphics[width=0.19\linewidth]{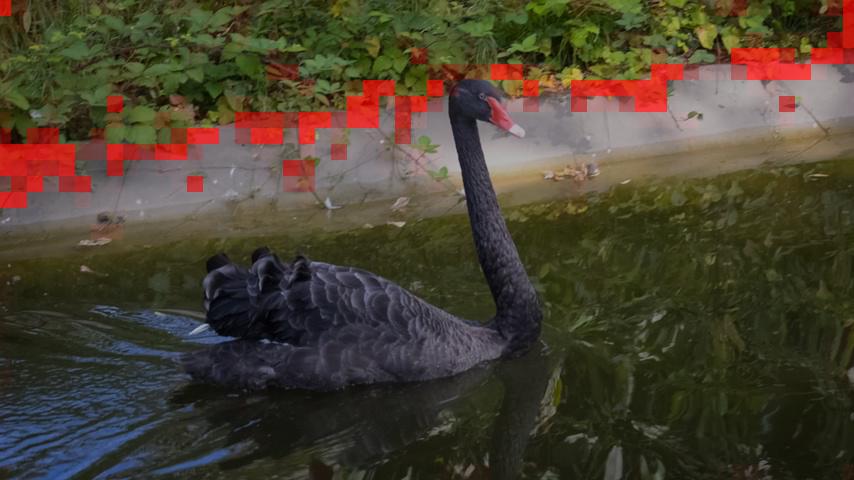} & 
	\includegraphics[width=0.19\linewidth]{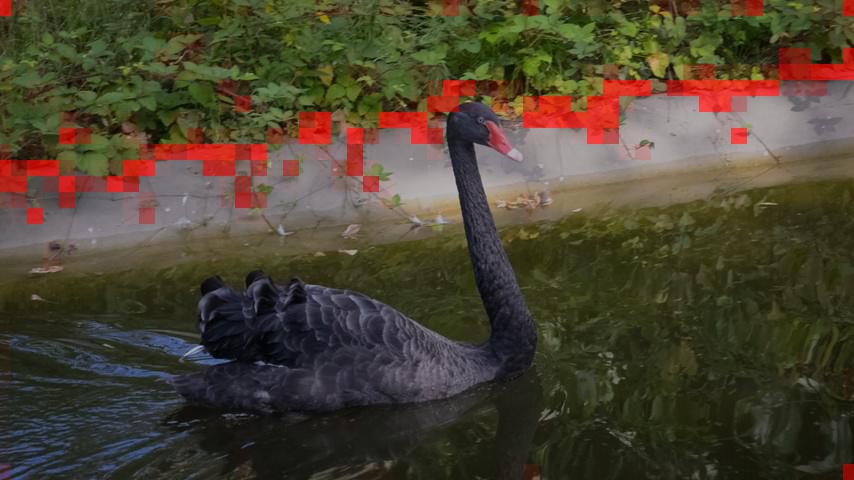} & 
	\includegraphics[width=0.19\linewidth,cfbox=red 1pt 0pt]{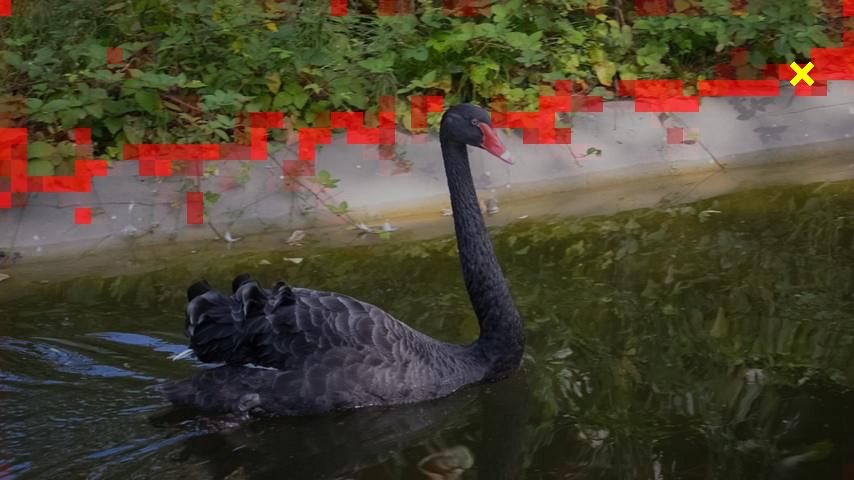} \\
	\vspace{-4.5mm}&&&&\\
	\includegraphics[width=0.19\linewidth]{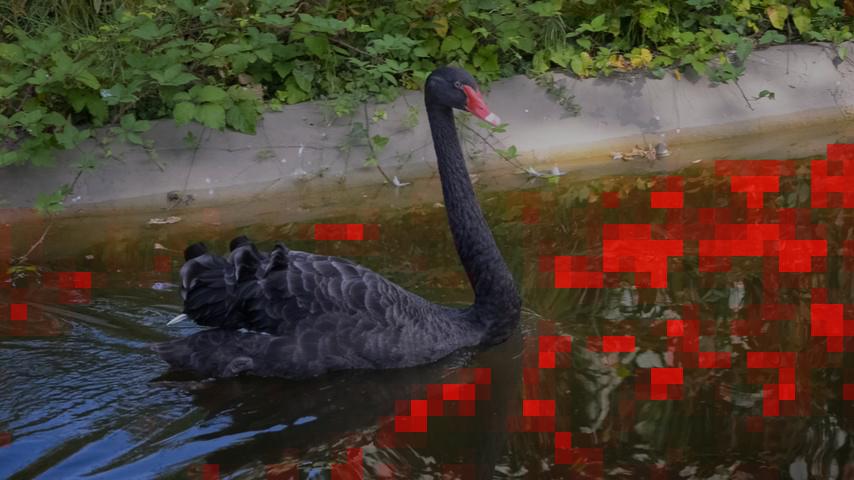} &
	\includegraphics[width=0.19\linewidth]{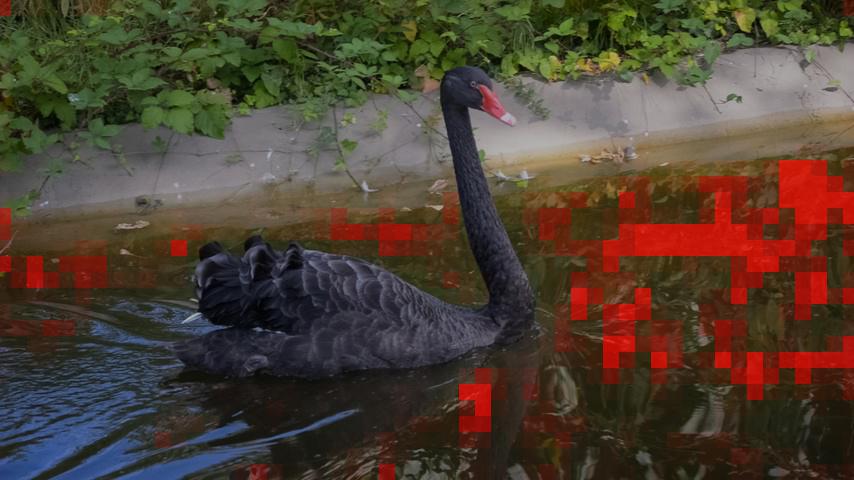} & 
	\includegraphics[width=0.19\linewidth]{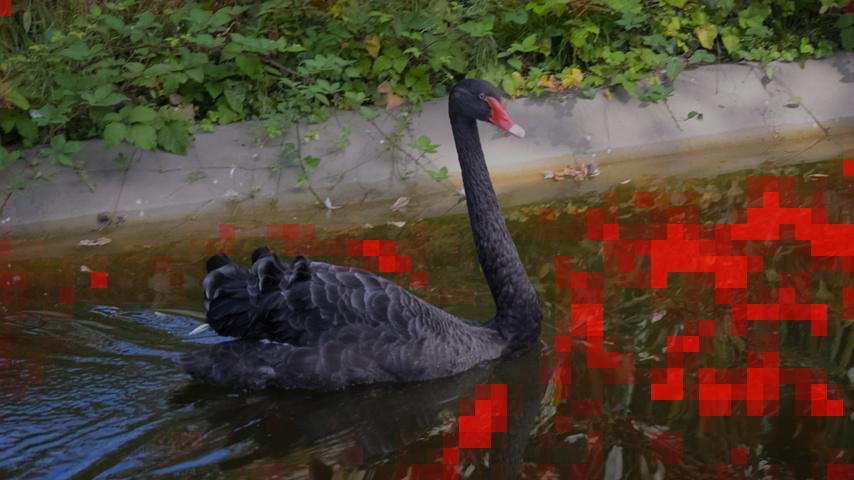} & 
	\includegraphics[width=0.19\linewidth]{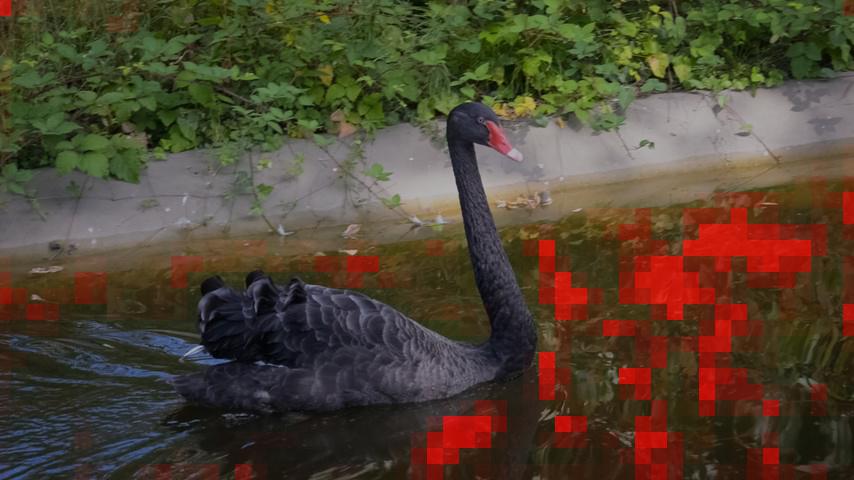} & 
	\includegraphics[width=0.19\linewidth,cfbox=red 1pt 0pt]{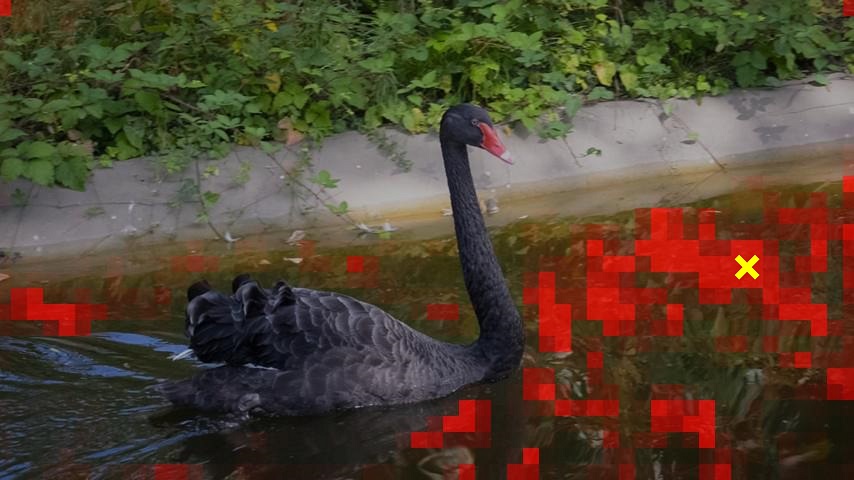} \\
\end{tabular}

	\vspace{-1em}
	\caption{Visualization of memory consolidation. The first row shows the candidate frames to be converted into long-term memory. 
	Each of the following rows show a prototype position  (indicated by a yellow cross), and the corresponding aggregation weights (visualized as a red overlay). 
	Frames that contain a prototype are framed in red.
	The consolidation process aggregates information from semantically meaningful regions (top-to-bottom): the swan's beak, part of the vegetation, part of the riverbank, the transition between vegetation and river bank, and part of the water surface.
	}
	\label{fig:vis-mem-a}
	\vspace{-1em}
\end{figure}

\begin{figure}
	\vspace{-1em}
	\centering
	\begin{tabular}{c@{\hspace{.5mm}}c@{\hspace{.5mm}}c@{\hspace{.5mm}}c@{\hspace{.5mm}}c}
	\includegraphics[width=0.19\linewidth]{appendix/pk/00004.jpg} &
	\includegraphics[width=0.19\linewidth]{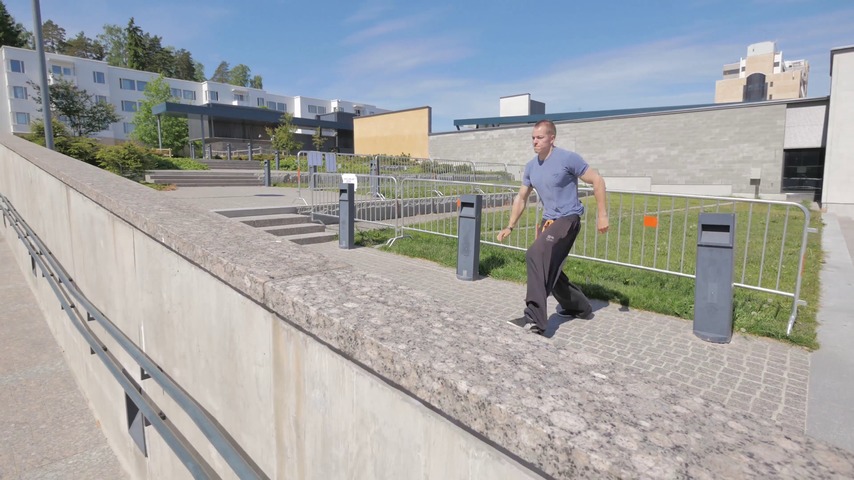} & 
	\includegraphics[width=0.19\linewidth]{appendix/pk/00014.jpg} & 
	\includegraphics[width=0.19\linewidth]{appendix/pk/00019.jpg} & 
	\includegraphics[width=0.19\linewidth]{appendix/pk/00024.jpg} \\
	\vspace{-4.5mm}&&&&\\
	\includegraphics[width=0.19\linewidth]{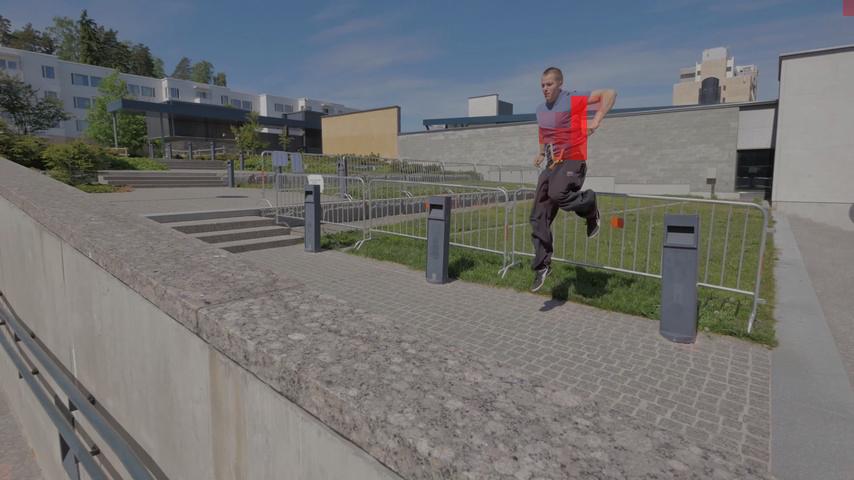} &
	\includegraphics[width=0.19\linewidth,cfbox=red 1pt 0pt]{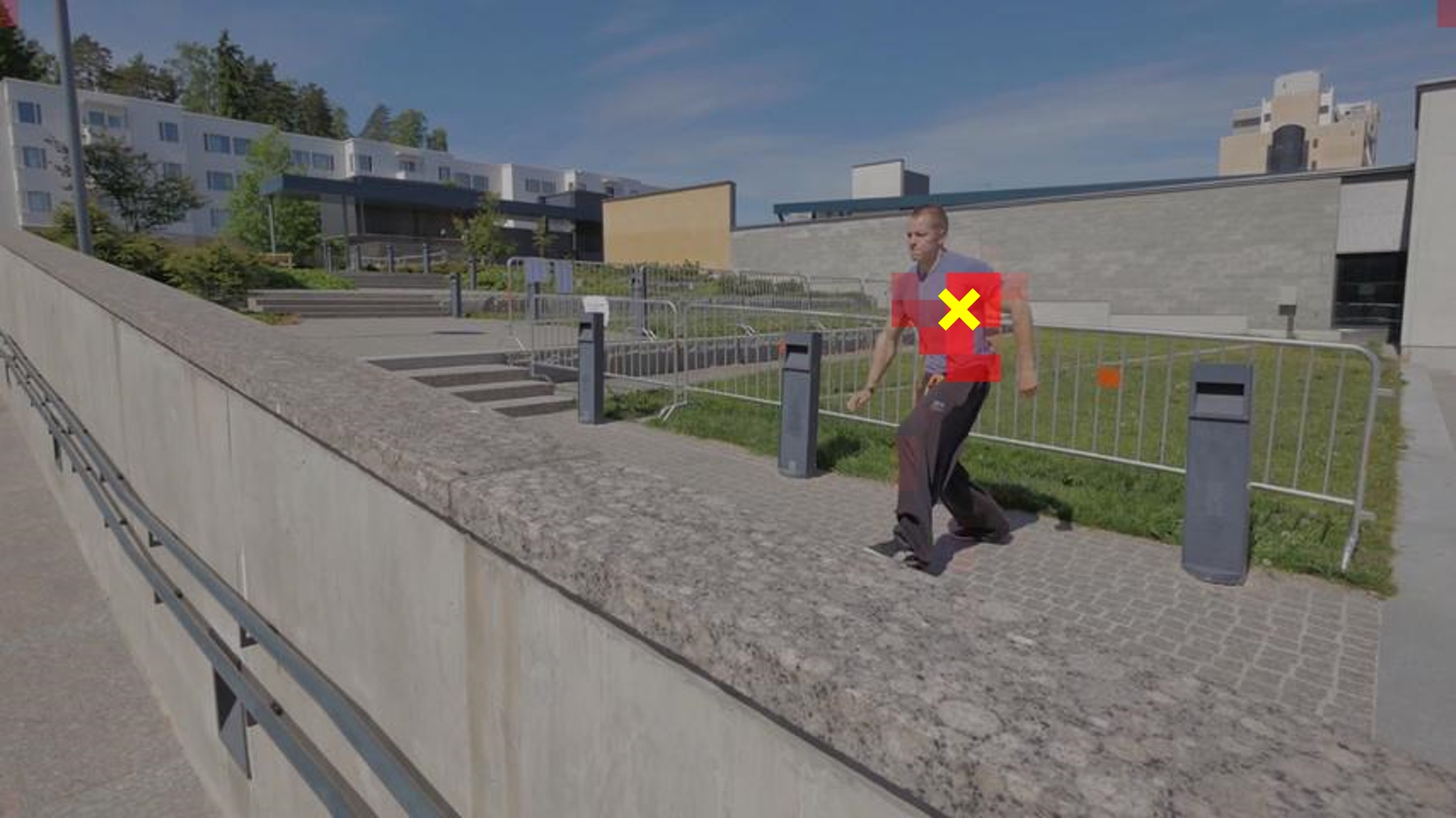} & 
	\includegraphics[width=0.19\linewidth]{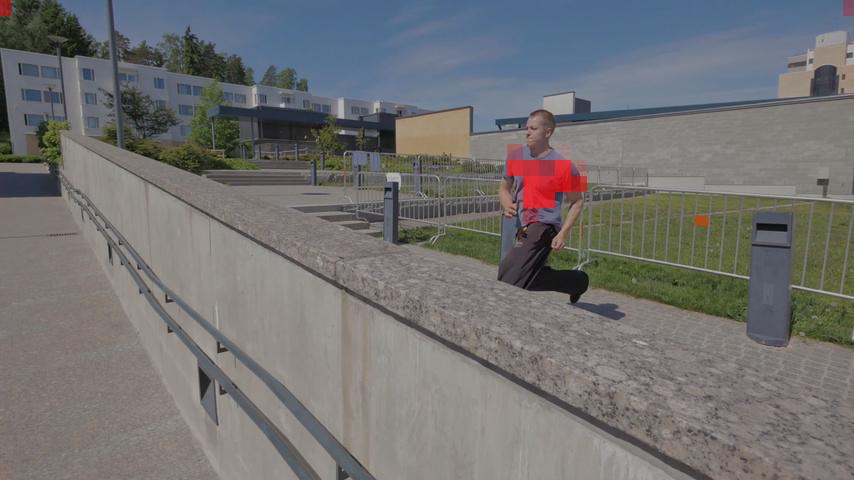} & 
	\includegraphics[width=0.19\linewidth]{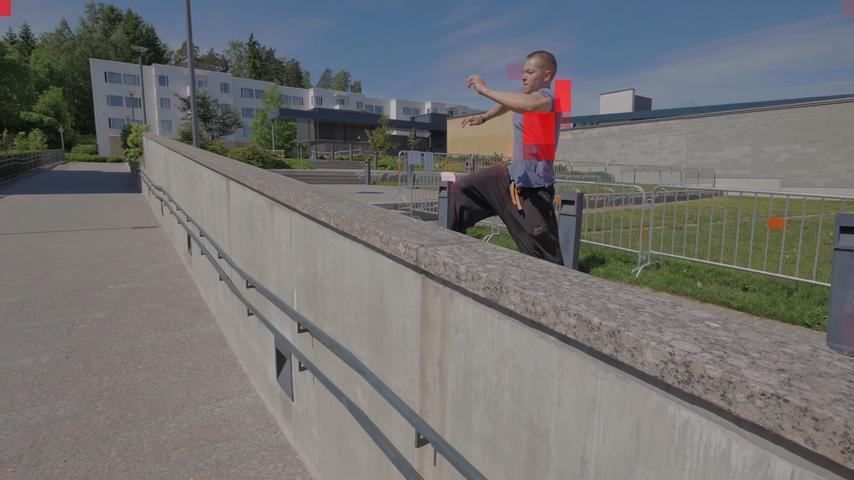} & 
	\includegraphics[width=0.19\linewidth]{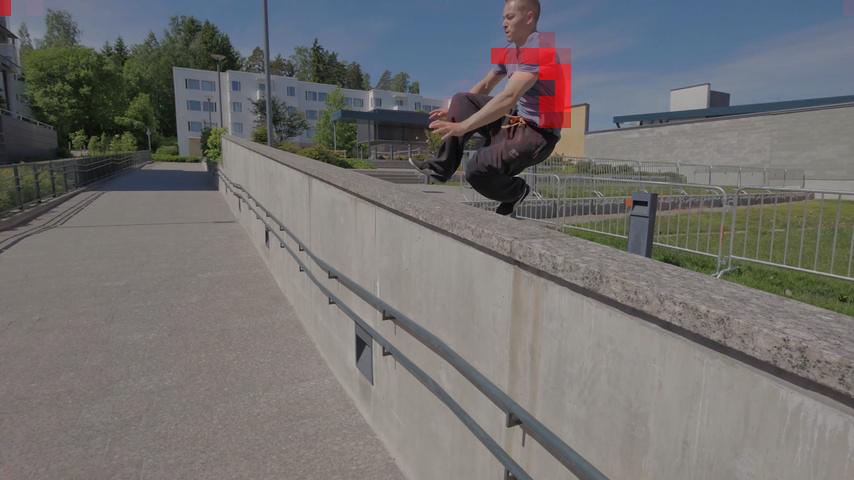} \\
	\vspace{-4.5mm}&&&&\\
	\includegraphics[width=0.19\linewidth]{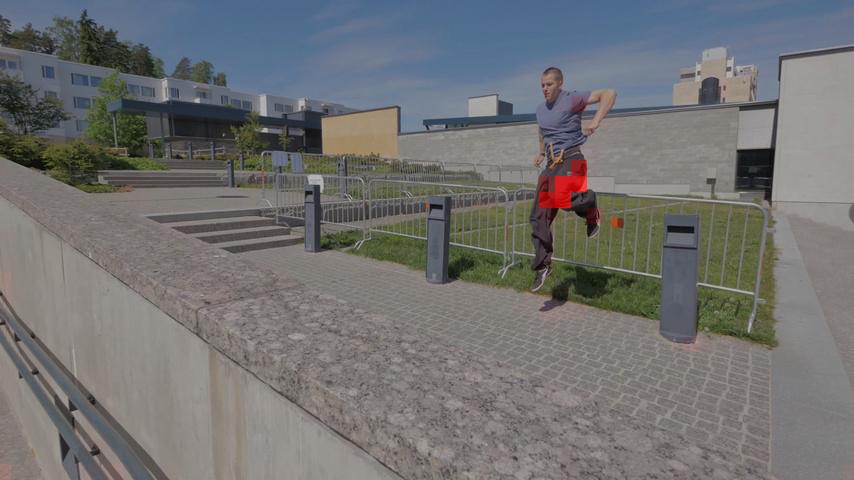} &
	\includegraphics[width=0.19\linewidth,cfbox=red 1pt 0pt]{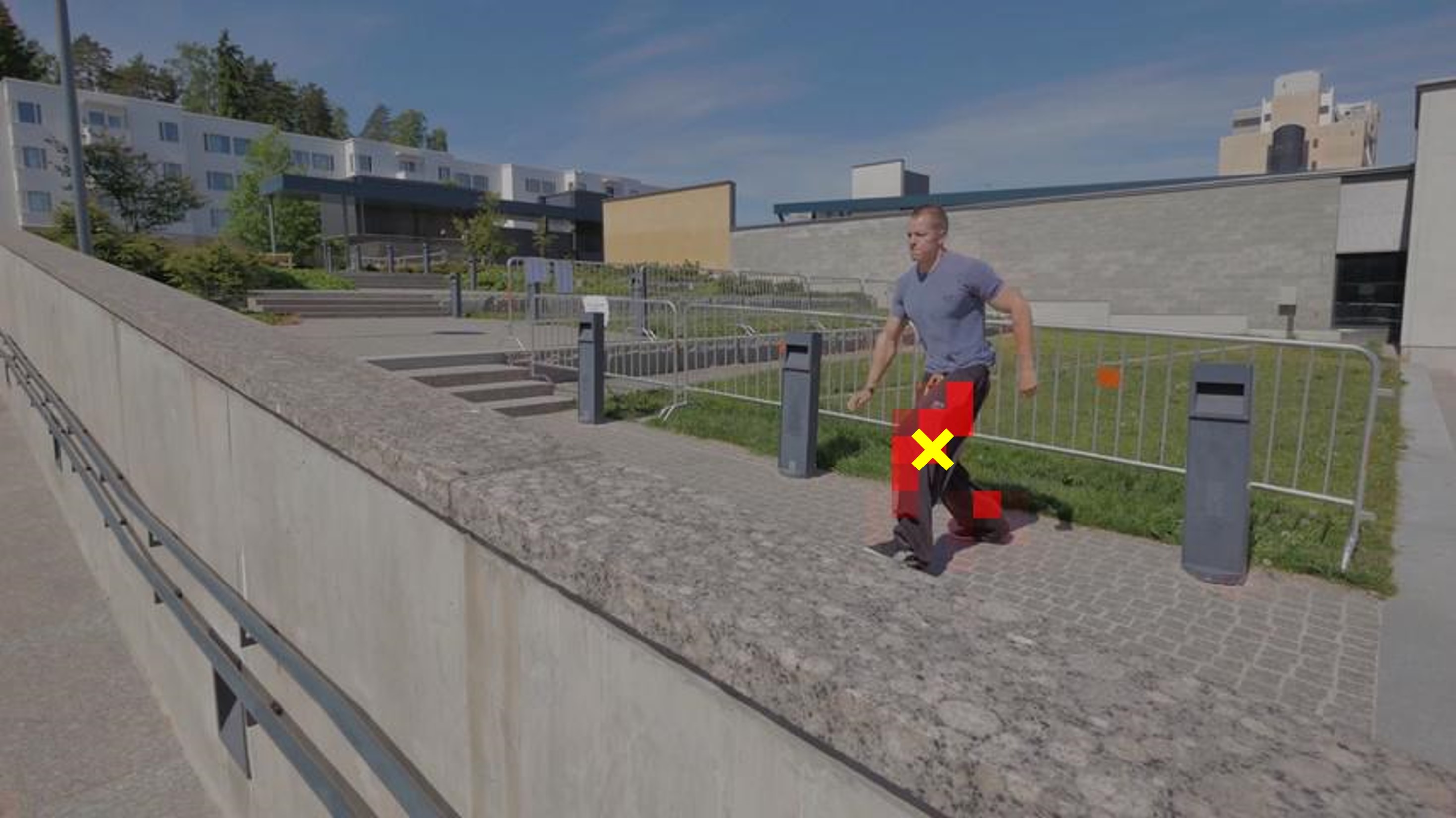} & 
	\includegraphics[width=0.19\linewidth]{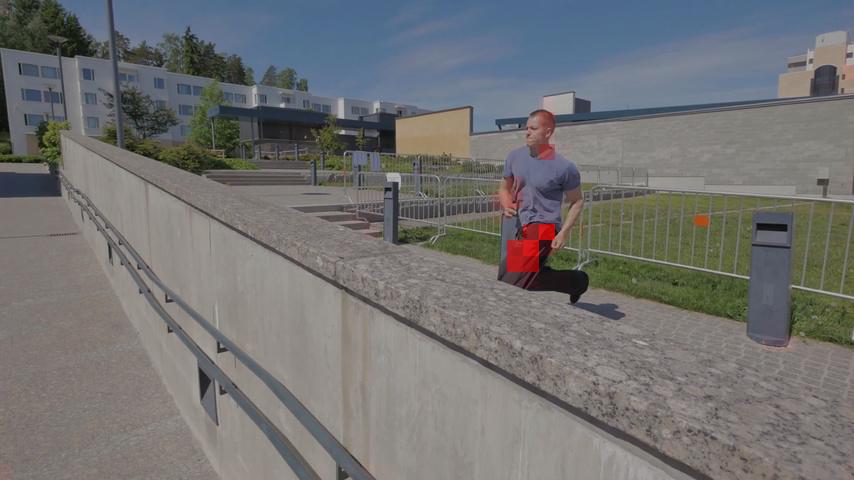} & 
	\includegraphics[width=0.19\linewidth]{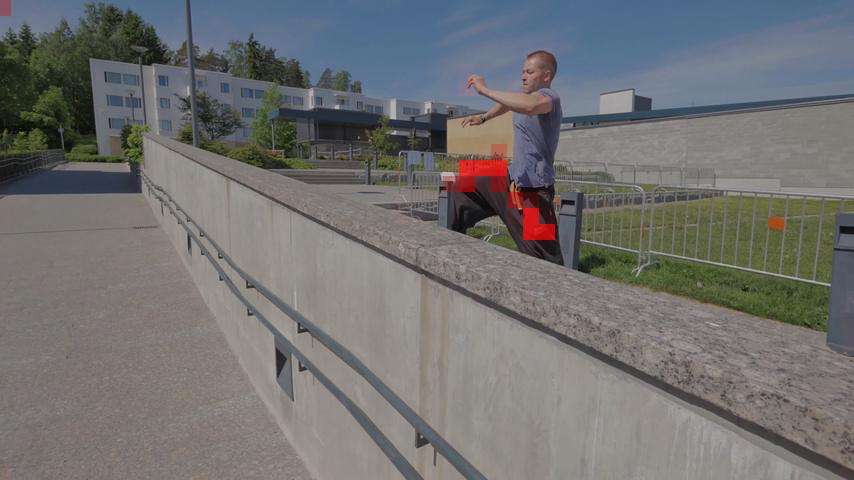} & 
	\includegraphics[width=0.19\linewidth]{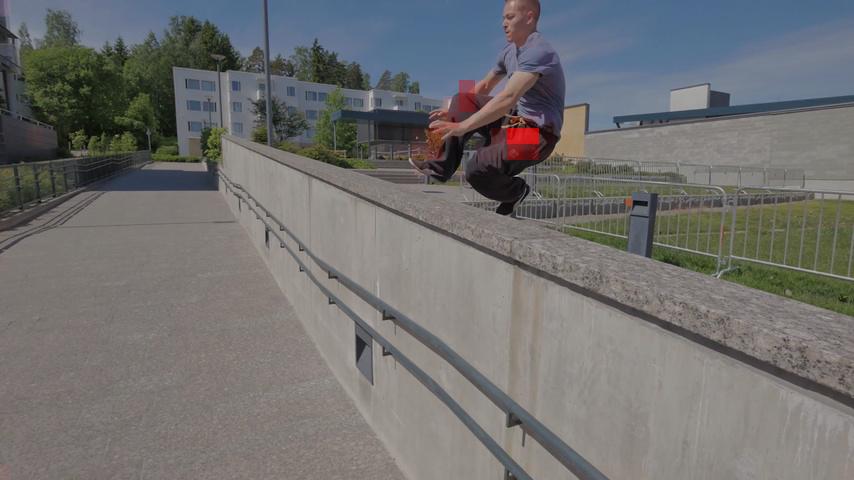} \\
	\vspace{-4.5mm}&&&&\\
	\includegraphics[width=0.19\linewidth]{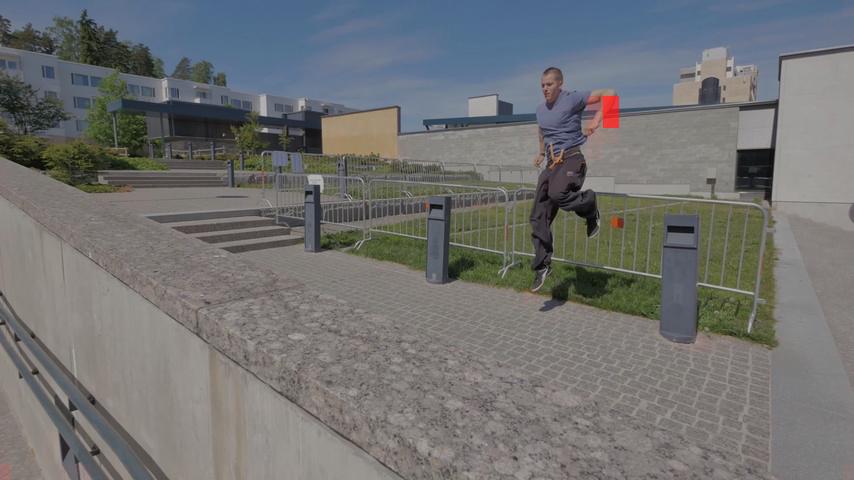} &
	\includegraphics[width=0.19\linewidth]{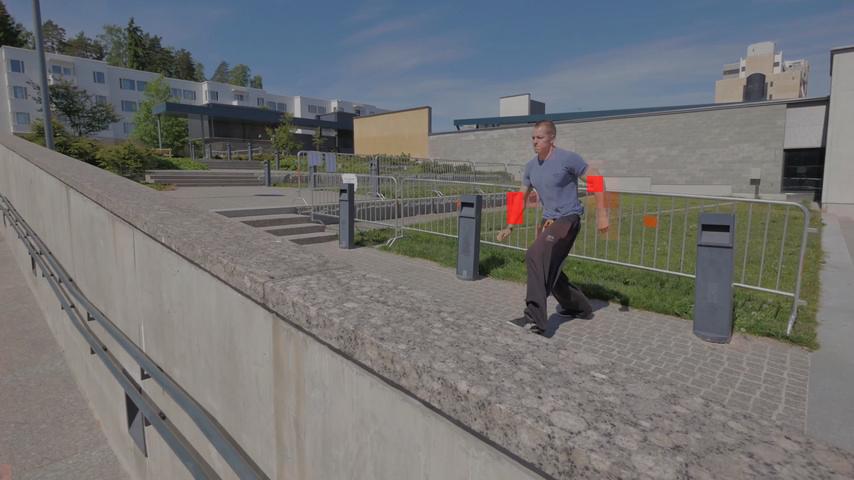} & 
	\includegraphics[width=0.19\linewidth,cfbox=red 1pt 0pt]{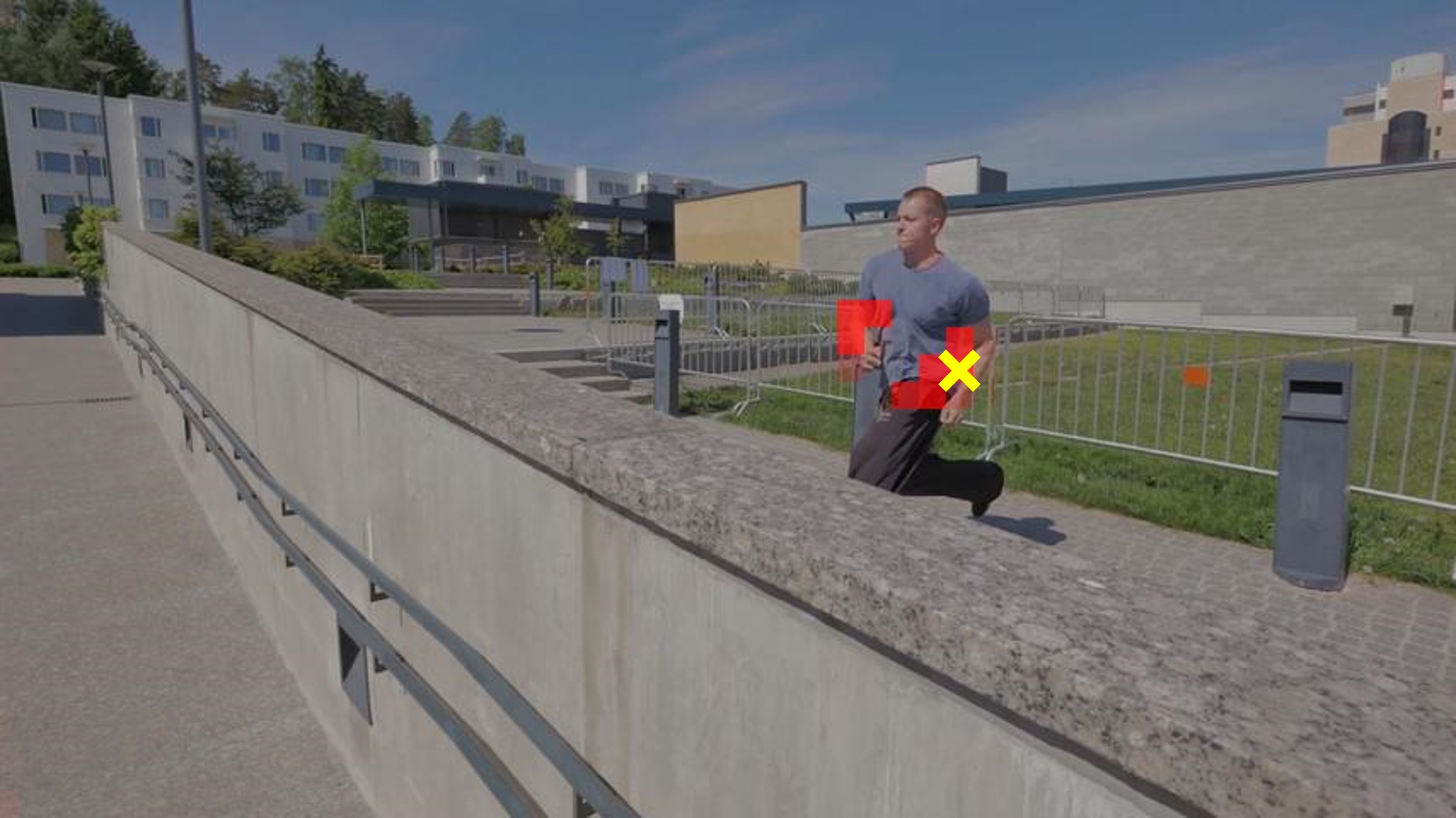} & 
	\includegraphics[width=0.19\linewidth]{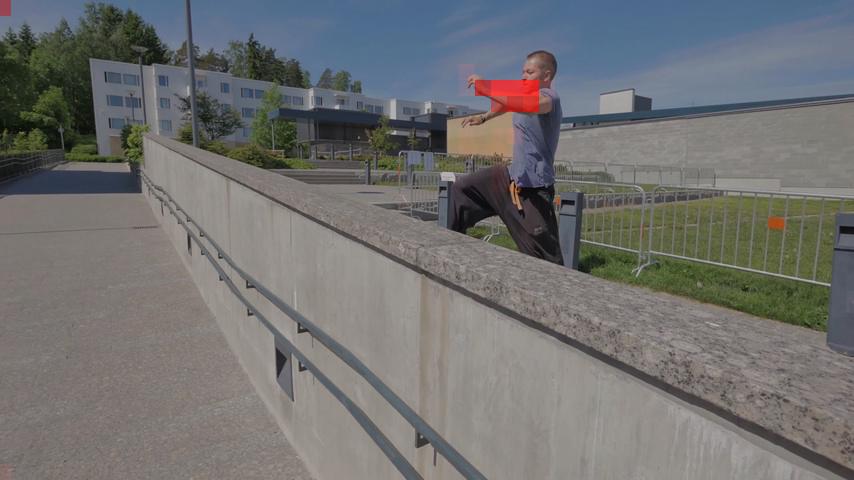} & 
	\includegraphics[width=0.19\linewidth]{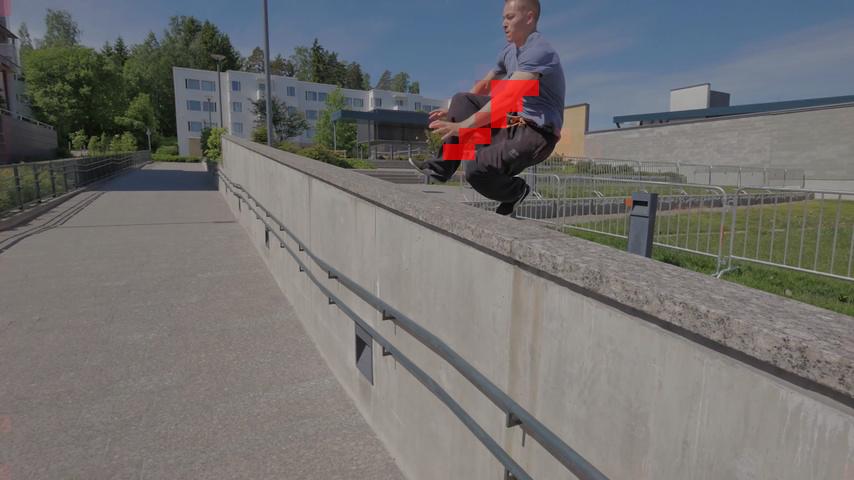} \\
	\vspace{-4.5mm}&&&&\\
	\includegraphics[width=0.19\linewidth]{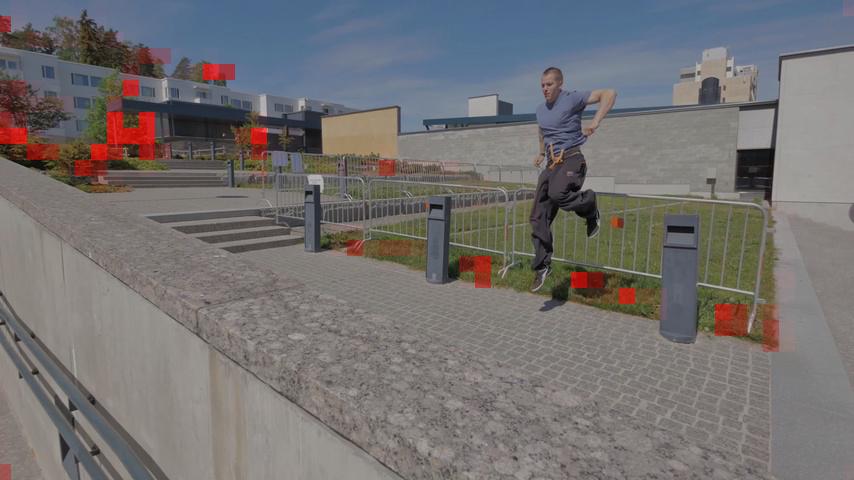} &
	\includegraphics[width=0.19\linewidth]{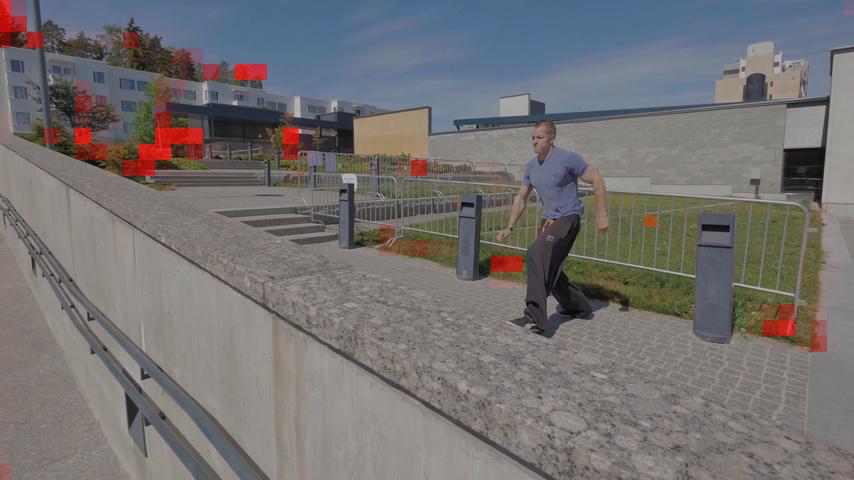} & 
	\includegraphics[width=0.19\linewidth]{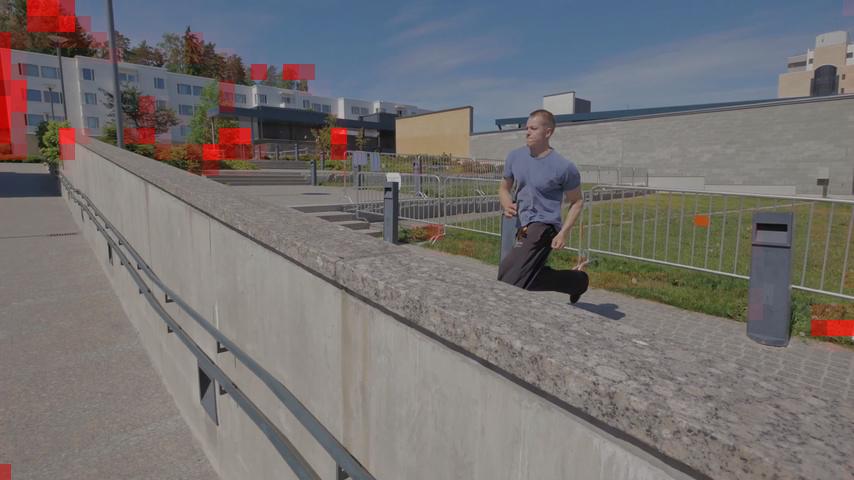} & 
	\includegraphics[width=0.19\linewidth,cfbox=red 1pt 0pt]{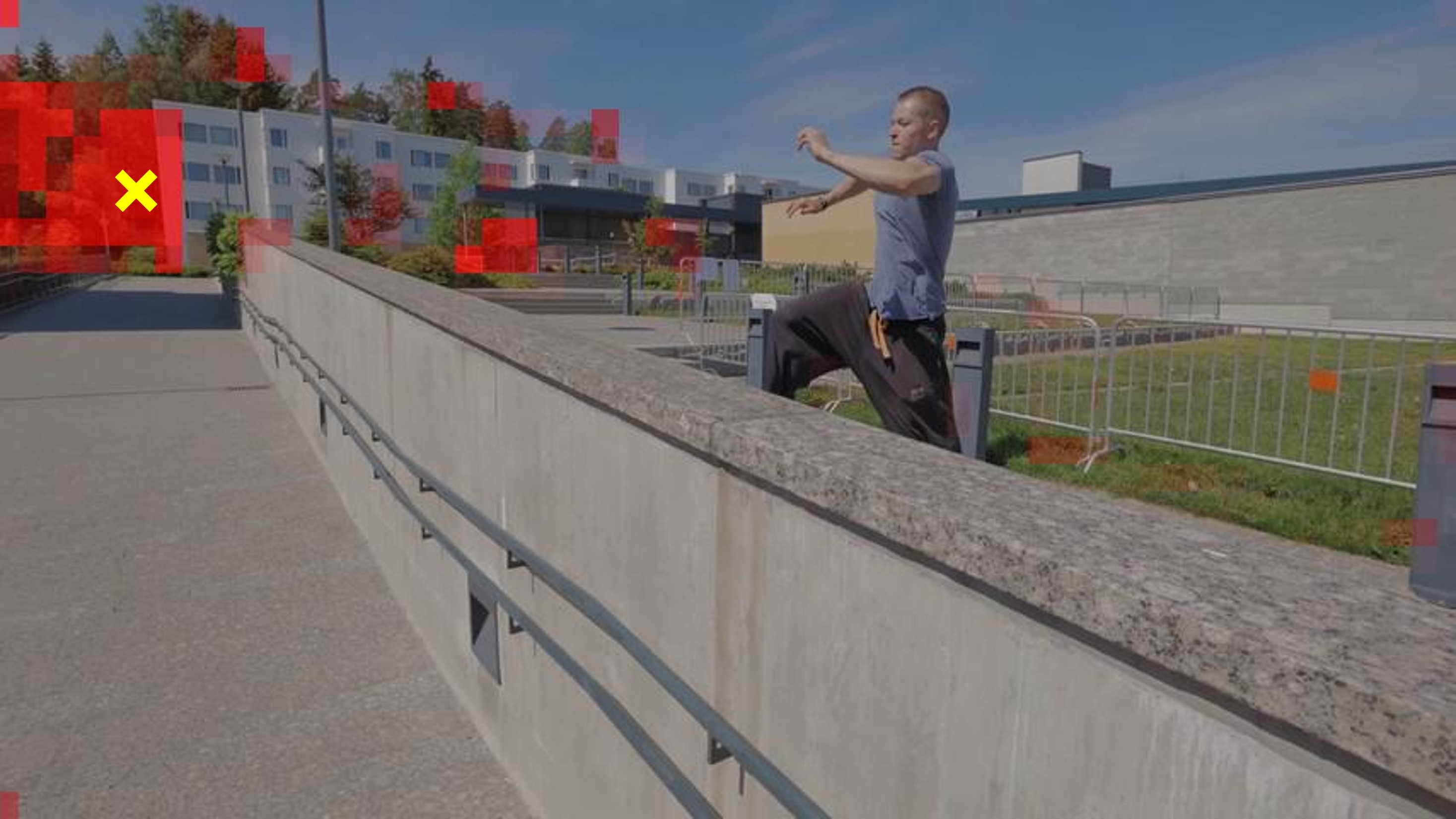} & 
	\includegraphics[width=0.19\linewidth]{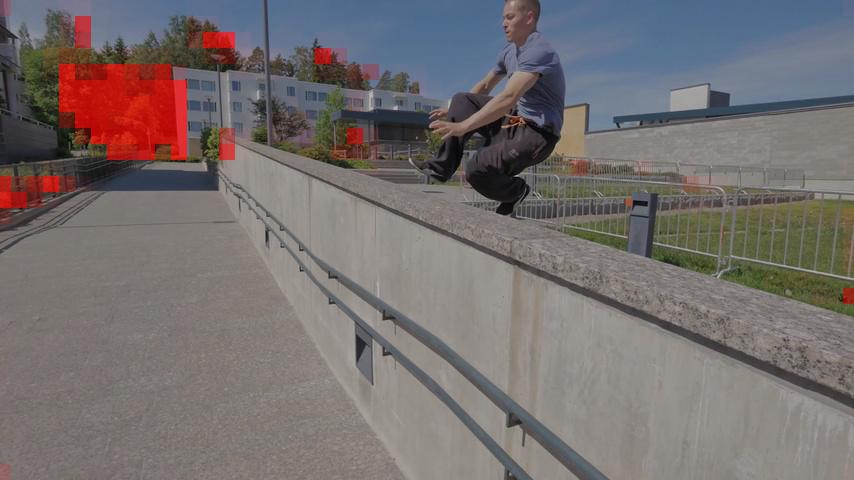} \\
	\vspace{-4.5mm}&&&&\\
	\includegraphics[width=0.19\linewidth]{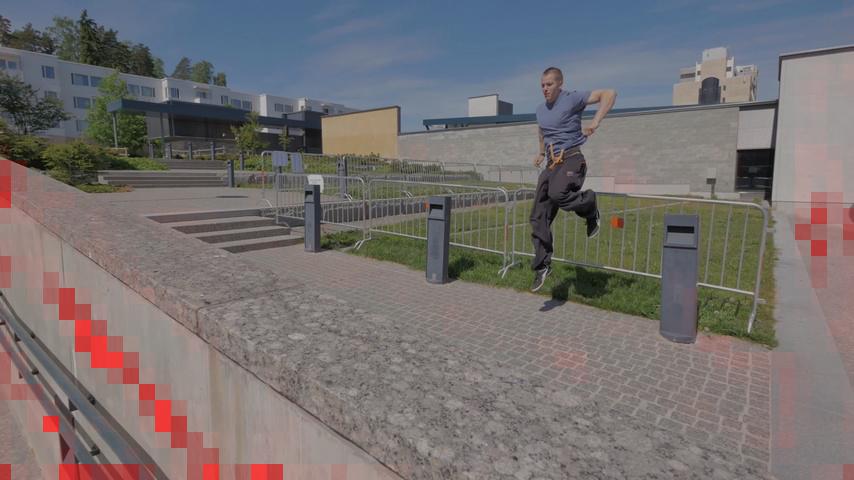} &
	\includegraphics[width=0.19\linewidth]{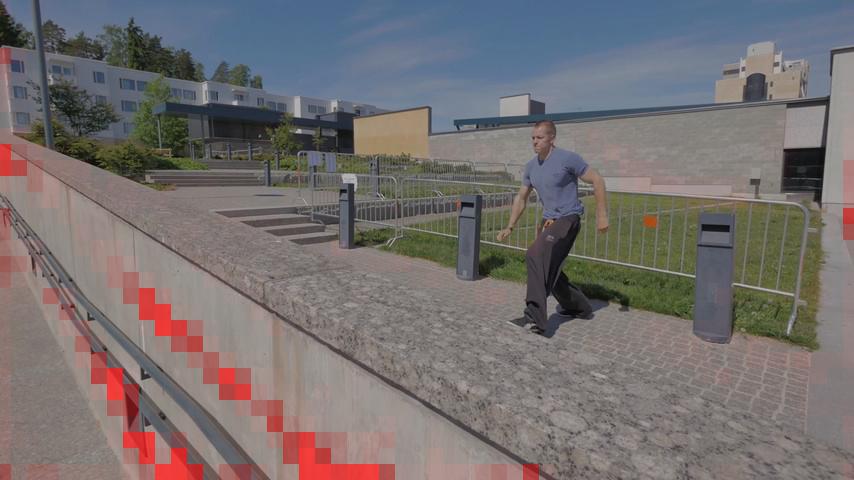} & 
	\includegraphics[width=0.19\linewidth]{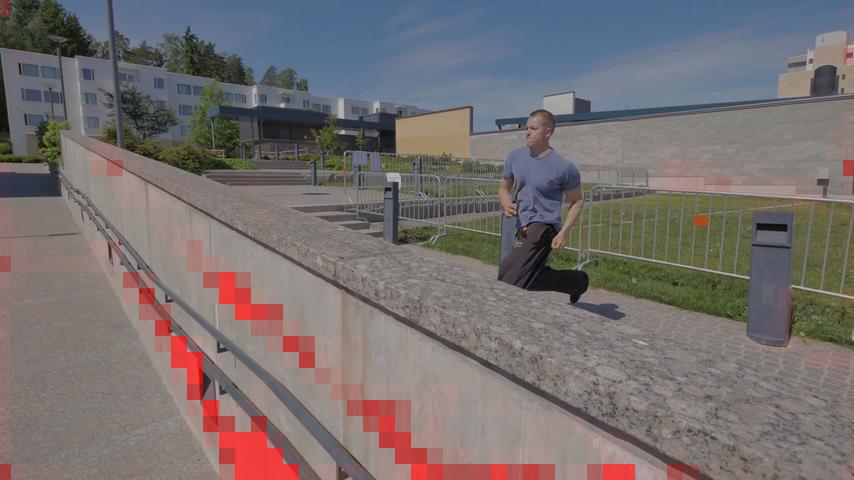} & 
	\includegraphics[width=0.19\linewidth,cfbox=red 1pt 0pt]{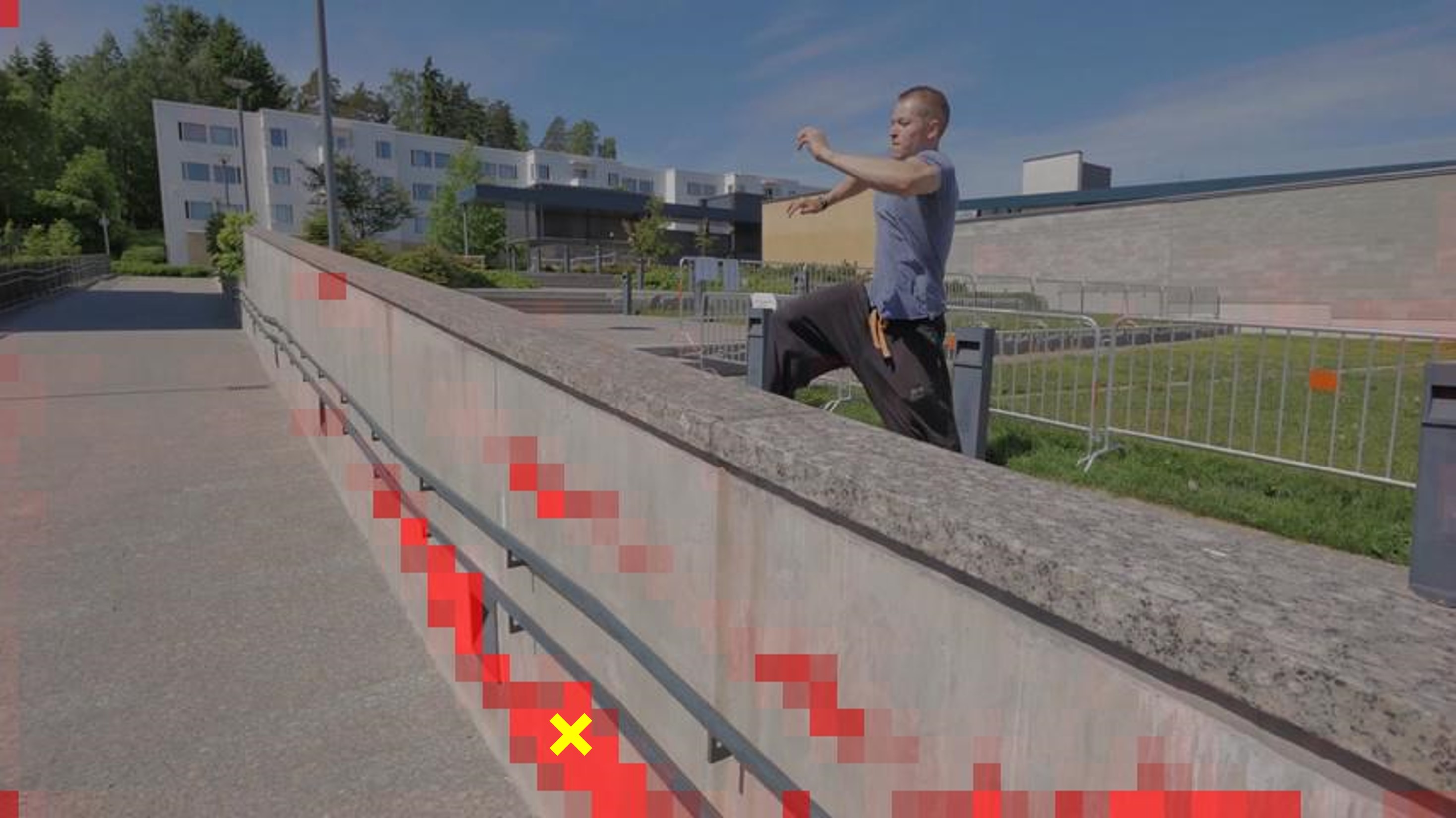} & 
	\includegraphics[width=0.19\linewidth]{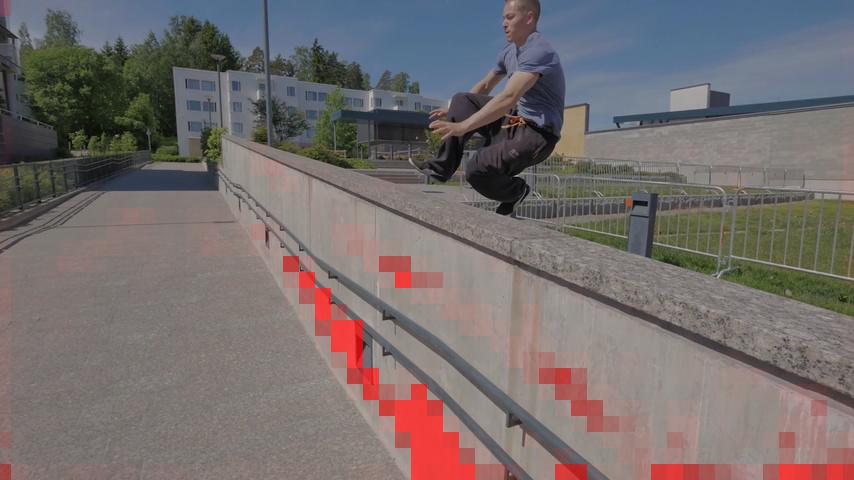} \\
\end{tabular}

	\vspace{-1em}
	\caption{Visualization of memory consolidation. The first row shows the candidate frames to be converted into long-term memory. 
		Each of the following rows show a prototype position  (indicated by a yellow cross), and the corresponding aggregation weights (visualized as a red overlay). 
		Frames that contain a prototype are framed in red.
		The consolidation process aggregates information from semantically meaningful regions (top-to-bottom): torso, legs, arms, trees, and part of the wall.
	}
	\label{fig:vis-mem-b}
	\vspace{-1em}
\end{figure}

Here, we visualize the memory consolidation process (Section 3.3) by showing the candidate frames, some of the selected prototypes, and the corresponding aggregation weights (columns of $\mathbf{W}(\mathbf{k}^\text{c}, \mathbf{k}^\text{p})$, each mapping to a distribution over all the candidates). Recall that $\mathbf{W}(\mathbf{k}^\text{c}, \mathbf{k}^\text{p})$ is used to aggregate candidate values $\mathbf{v}^\text{c}$ into prototype values $\mathbf{v}^\text{p}$.
Figure~\ref{fig:vis-mem-a} and Figure~\ref{fig:vis-mem-b} show two examples. As illustrated in Figure 5 of the main paper we observe semantically meaningful regions to be grouped.

\clearpage

\section{Qualitative Results}
\label{sec:app:qual}
\begin{figure}
\vspace{-1em}
	\centering
	\scriptsize
	\begin{tabular}{c@{\hspace{1mm}}c@{\hspace{.5mm}}c@{\hspace{.5mm}}c@{\hspace{.5mm}}c@{\hspace{.5mm}}c}
	\rotatebox[origin=c]{90}{\scriptsize Images \hspace{1mm}}&
	\raisebox{-0.5\height}{\includegraphics[width=0.18\linewidth]{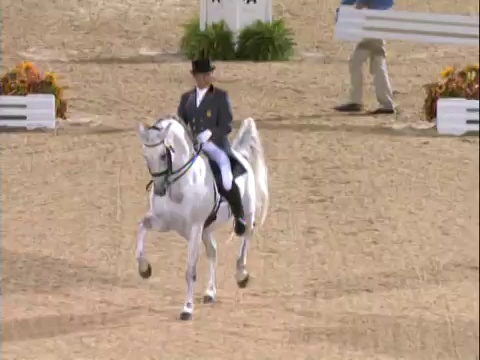}}&
	\raisebox{-0.5\height}{\includegraphics[width=0.18\linewidth]{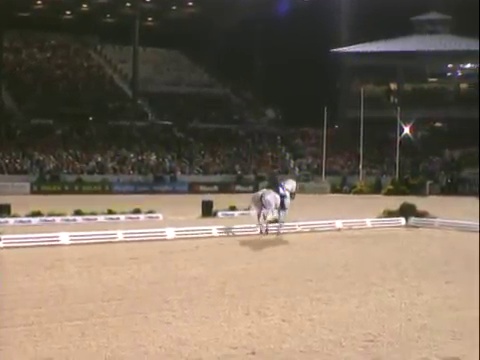}}&
	\raisebox{-0.5\height}{\includegraphics[width=0.18\linewidth]{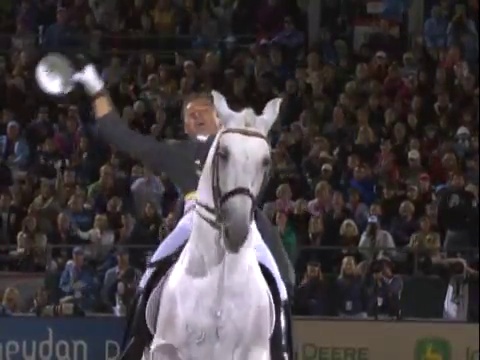}}&
	\raisebox{-0.5\height}{\includegraphics[width=0.18\linewidth]{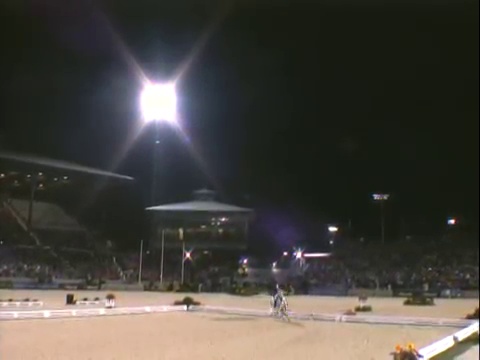}}&
	\raisebox{-0.5\height}{\includegraphics[width=0.18\linewidth]{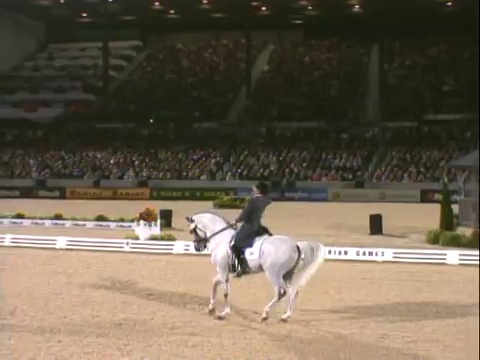}}\\
	\vspace{-2.5mm}&&&&&\\
	\rotatebox[origin=c]{90}{\scriptsize JOINT \hspace{1mm}}&
	\raisebox{-0.5\height}{\includegraphics[width=0.18\linewidth]{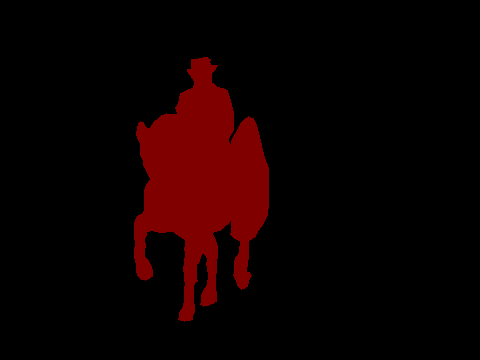}}&
	\raisebox{-0.5\height}{\includegraphics[width=0.18\linewidth]{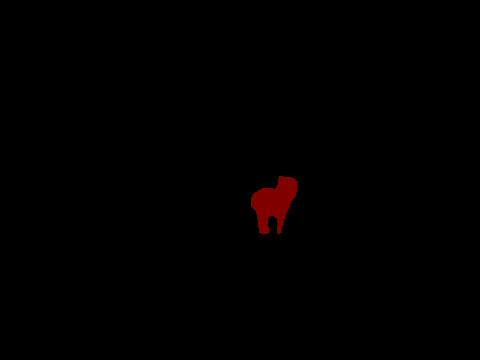}}&
	\raisebox{-0.5\height}{\includegraphics[width=0.18\linewidth]{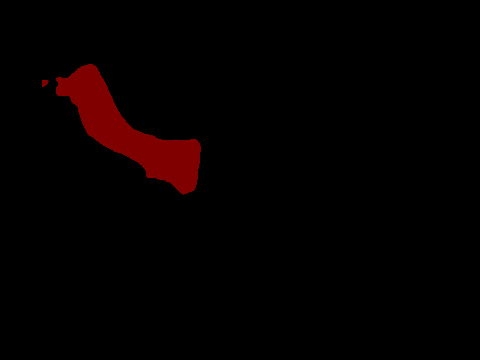}}&
	\raisebox{-0.5\height}{\includegraphics[width=0.18\linewidth]{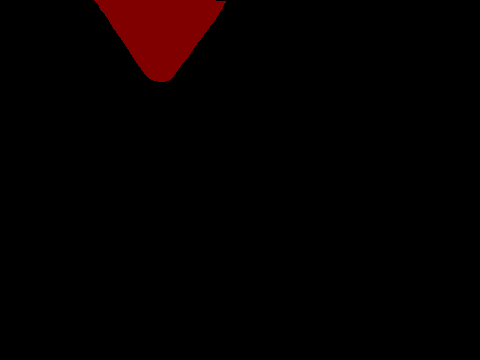}}&
	\raisebox{-0.5\height}{\includegraphics[width=0.18\linewidth]{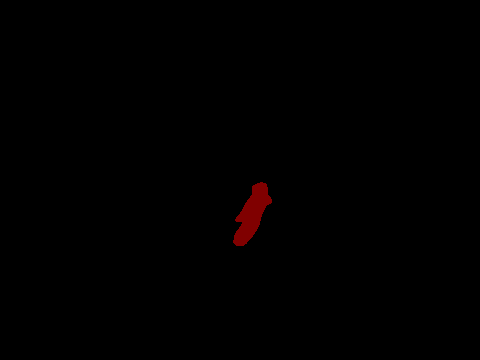}}\\
	\vspace{-2.5mm}&&&&&\\
	\rotatebox[origin=c]{90}{\tiny AFB-URR \hspace{1mm}}&
	\raisebox{-0.5\height}{\includegraphics[width=0.18\linewidth]{appendix/dressage/00000.png}}&
	\raisebox{-0.5\height}{\includegraphics[width=0.18\linewidth]{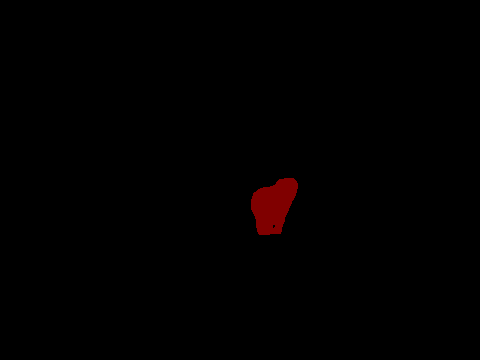}}&
	\raisebox{-0.5\height}{\includegraphics[width=0.18\linewidth]{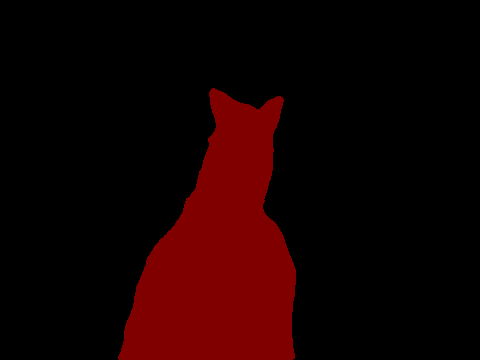}}&
	\raisebox{-0.5\height}{\includegraphics[width=0.18\linewidth]{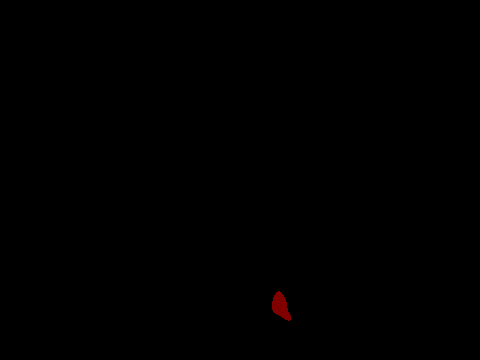}}&
	\raisebox{-0.5\height}{\includegraphics[width=0.18\linewidth]{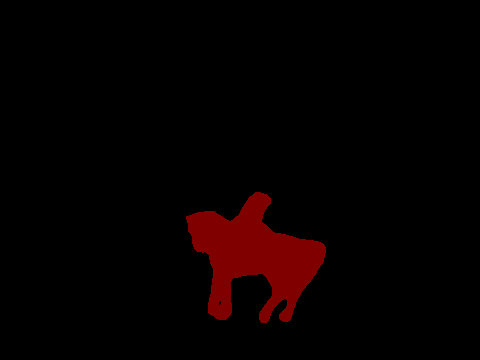}}\\
	\vspace{-2.5mm}&&&&&\\
	\rotatebox[origin=c]{90}{\scriptsize STCN \hspace{1mm}}&
	\raisebox{-0.5\height}{\includegraphics[width=0.18\linewidth]{appendix/dressage/00000.png}}&
	\raisebox{-0.5\height}{\includegraphics[width=0.18\linewidth]{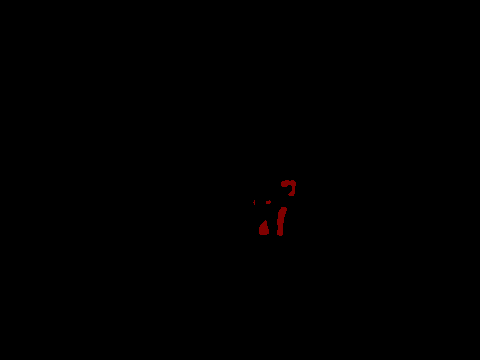}}&
	\raisebox{-0.5\height}{\includegraphics[width=0.18\linewidth]{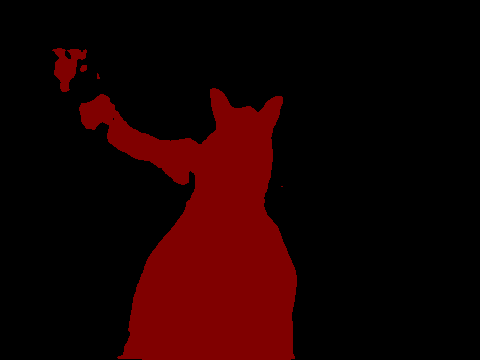}}&
	\raisebox{-0.5\height}{\includegraphics[width=0.18\linewidth]{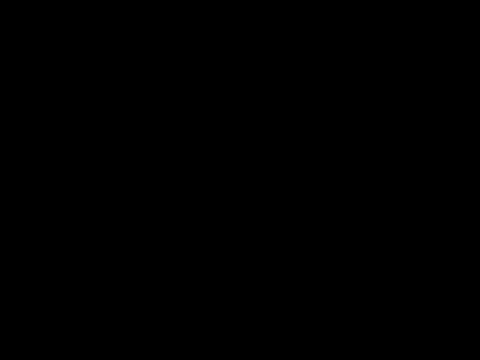}}&
	\raisebox{-0.5\height}{\includegraphics[width=0.18\linewidth]{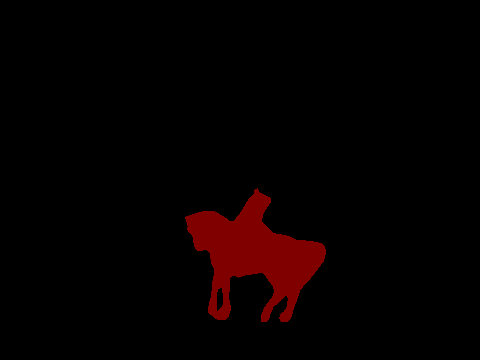}}\\
	\vspace{-2.5mm}&&&&&\\
	\rotatebox[origin=c]{90}{\scriptsize Ours \hspace{1mm}}&
	\raisebox{-0.5\height}{\includegraphics[width=0.18\linewidth]{appendix/dressage/00000.png}}&
	\raisebox{-0.5\height}{\includegraphics[width=0.18\linewidth]{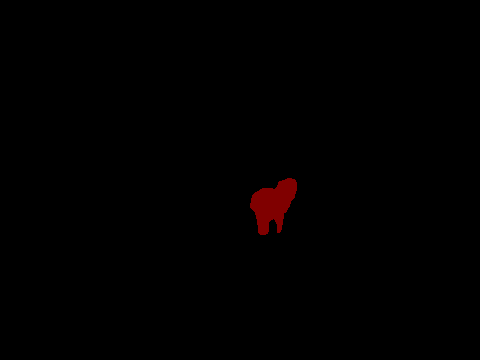}}&
	\raisebox{-0.5\height}{\includegraphics[width=0.18\linewidth]{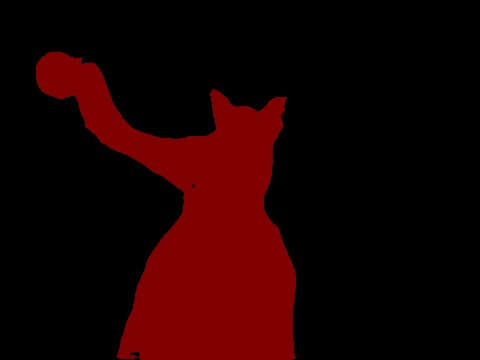}}&
	\raisebox{-0.5\height}{\includegraphics[width=0.18\linewidth]{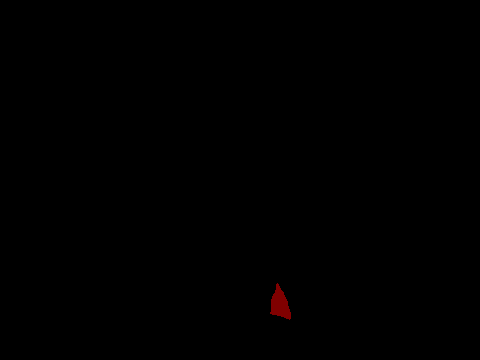}}&
	\raisebox{-0.5\height}{\includegraphics[width=0.18\linewidth]{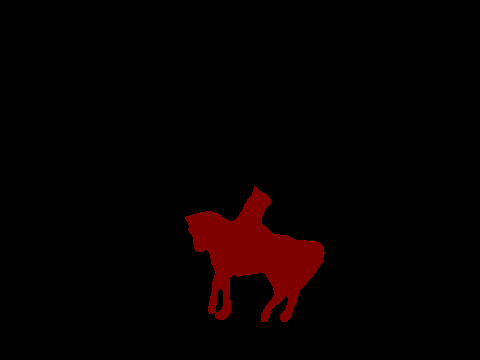}}\\
	\vspace{-2.5mm}&&&&&\\
	\rotatebox[origin=c]{90}{\scriptsize GT \hspace{1mm}}&
	\raisebox{-0.5\height}{\includegraphics[width=0.18\linewidth]{appendix/dressage/00000.png}}&
	\raisebox{-0.5\height}{\includegraphics[width=0.18\linewidth]{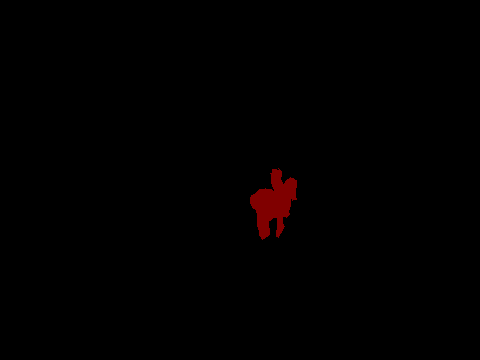}}&
	\raisebox{-0.5\height}{\includegraphics[width=0.18\linewidth]{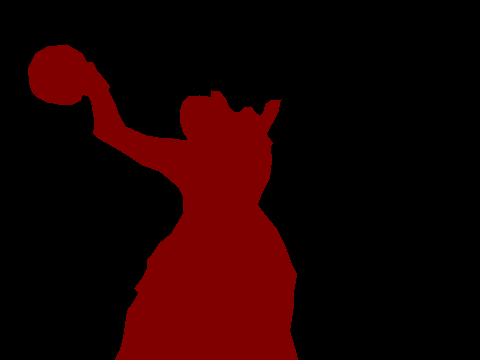}}&
	\raisebox{-0.5\height}{\includegraphics[width=0.18\linewidth]{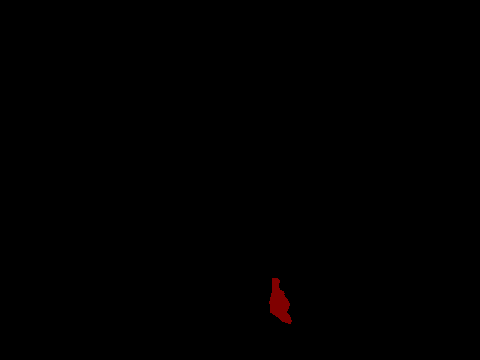}}&
	\raisebox{-0.5\height}{\includegraphics[width=0.18\linewidth]{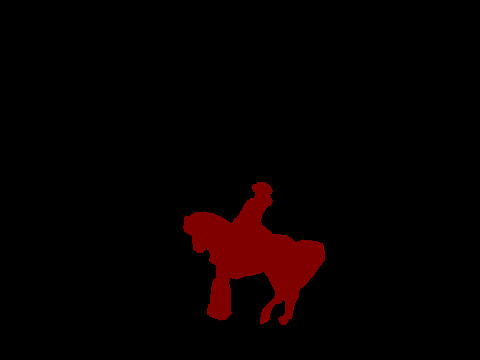}}\\
	\vspace{-2mm}&&&&&\\
	& Frame 0 (input) & Frame 2256 & Frame 3588 & Frame 4168 & Frame 10748 \\
\end{tabular}
	\caption{Results on the {\tt dressage} sequence.
		JOINT~\cite{mao2021joint} uses a temporally local feature window and loses track over time.
		AFB-URR~\cite{Liang2020AFBURR} is stable but produces overall less accurate segmentations.
		STCN~\cite{cheng2021stcn} uses a low memory insertion frequency to avoid memory explosion, and thus misses fast changes (2nd and 4th column).
		Ours is sometimes better than the provided ground-truth (last column, the horse's front legs).}
	\label{fig:vis-dressage}
	\vspace{-1em}
\end{figure}

\begin{figure}[!htb]
	\vspace{-1em}
	\centering
	\begin{tabular}{c@{\hspace{1mm}}c@{\hspace{.5mm}}c@{\hspace{.5mm}}c@{\hspace{.5mm}}c@{\hspace{.5mm}}c}
	\rotatebox[origin=c]{90}{\scriptsize Images \hspace{1mm}}&
	\raisebox{-0.5\height}{\includegraphics[width=0.18\linewidth]{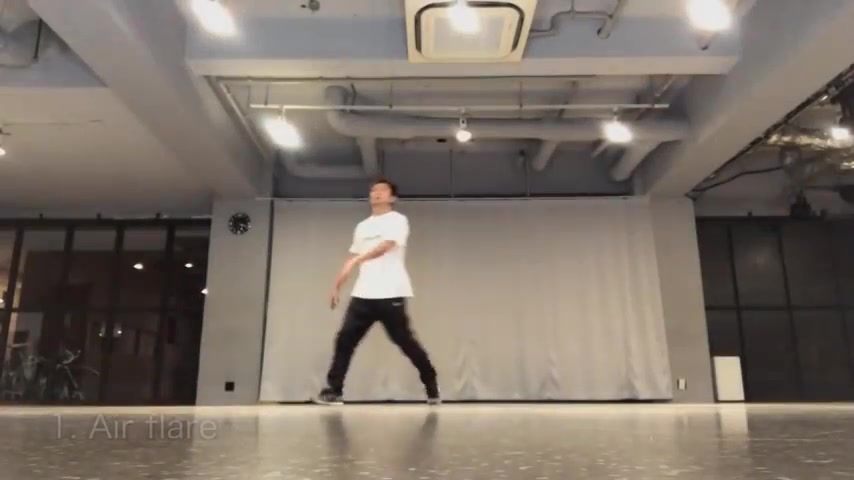}}&
	\raisebox{-0.5\height}{\includegraphics[width=0.18\linewidth]{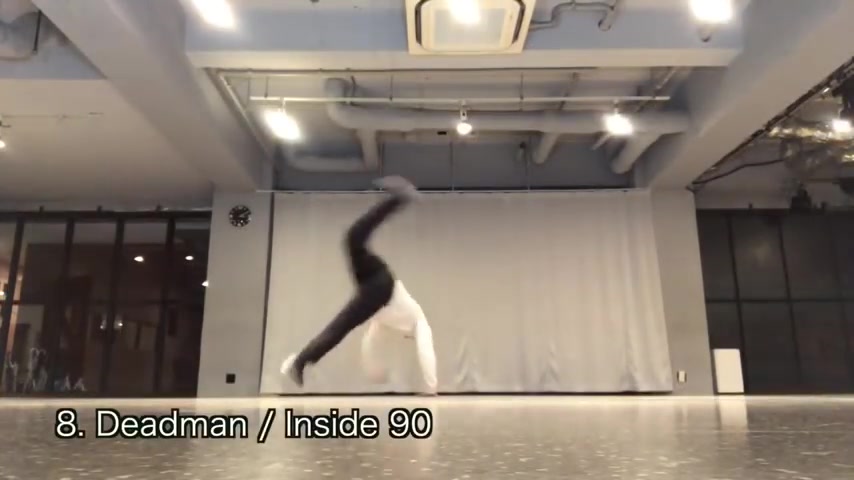}}&
	\raisebox{-0.5\height}{\includegraphics[width=0.18\linewidth]{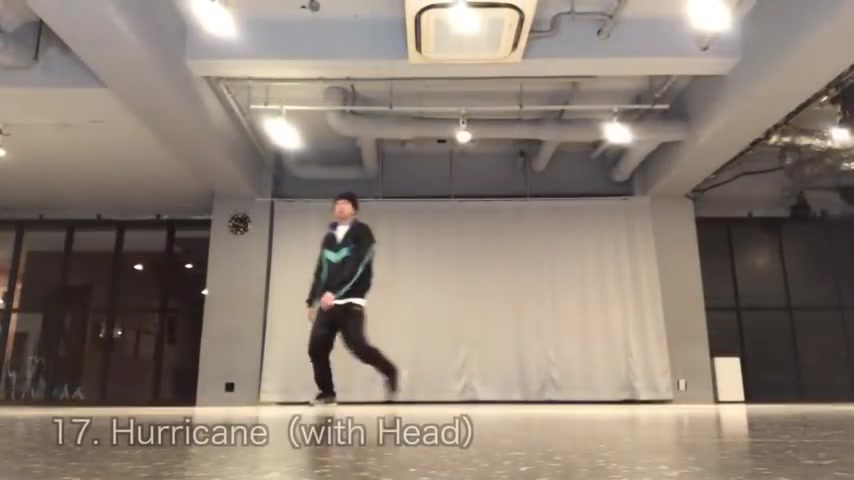}}&
	\raisebox{-0.5\height}{\includegraphics[width=0.18\linewidth]{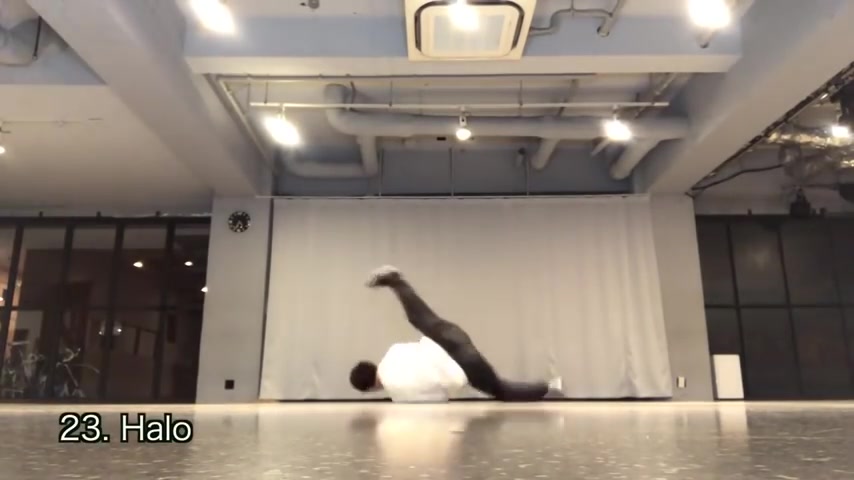}}&
	\raisebox{-0.5\height}{\includegraphics[width=0.18\linewidth]{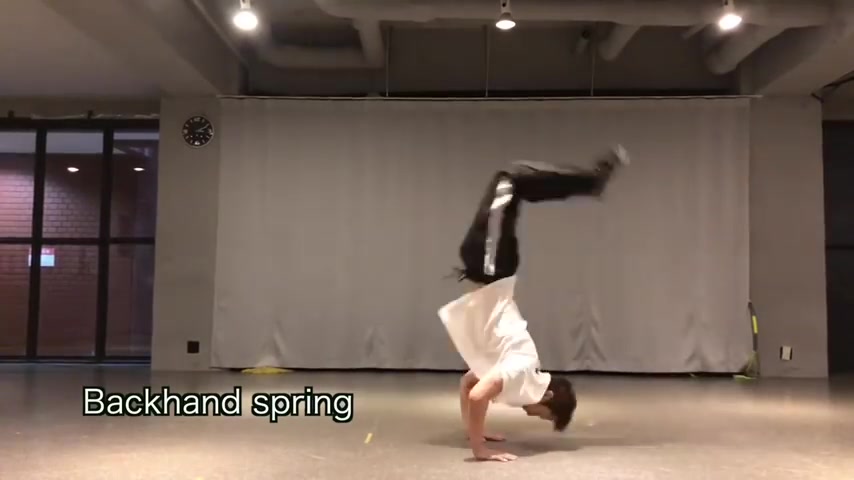}}\\
	\vspace{-3.5mm}&&&&&\\
	\rotatebox[origin=c]{90}{\scriptsize JOINT \hspace{1mm}}&
	\raisebox{-0.5\height}{\includegraphics[width=0.18\linewidth]{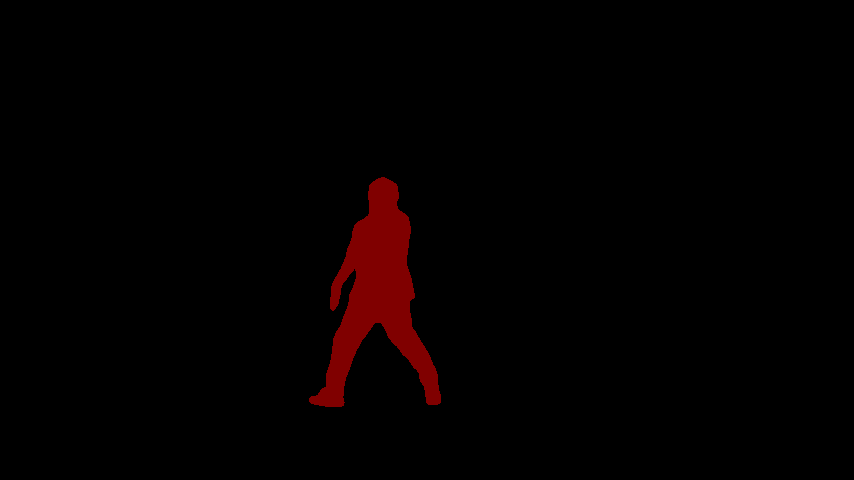}}&
	\raisebox{-0.5\height}{\includegraphics[width=0.18\linewidth]{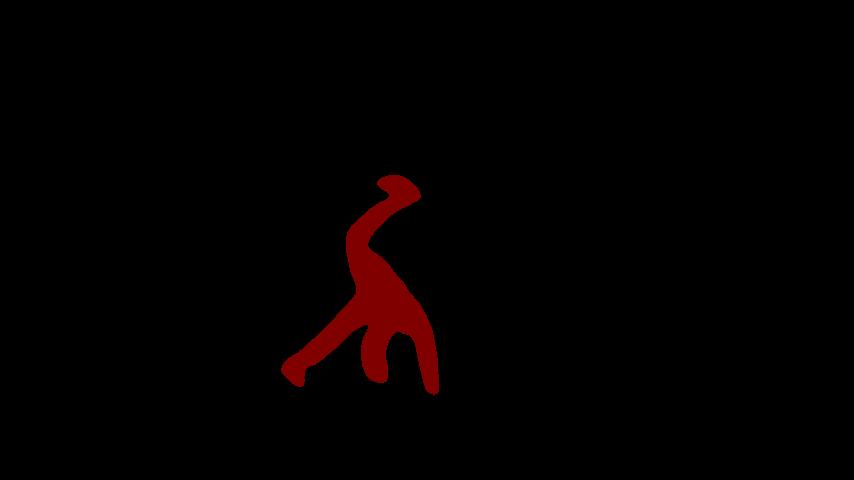}}&
	\raisebox{-0.5\height}{\includegraphics[width=0.18\linewidth]{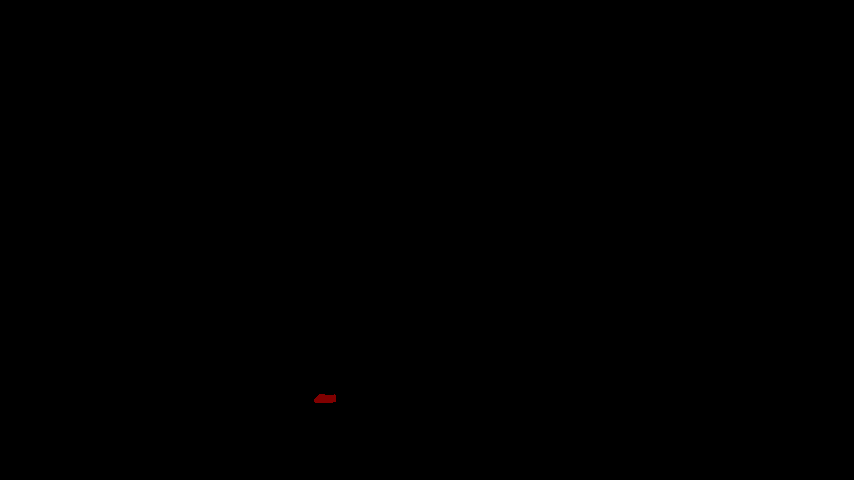}}&
	\raisebox{-0.5\height}{\includegraphics[width=0.18\linewidth]{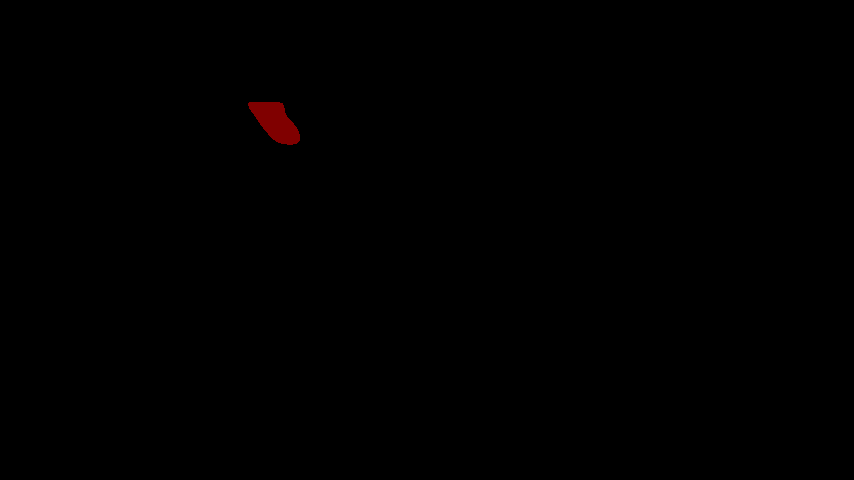}}&
	\raisebox{-0.5\height}{\includegraphics[width=0.18\linewidth]{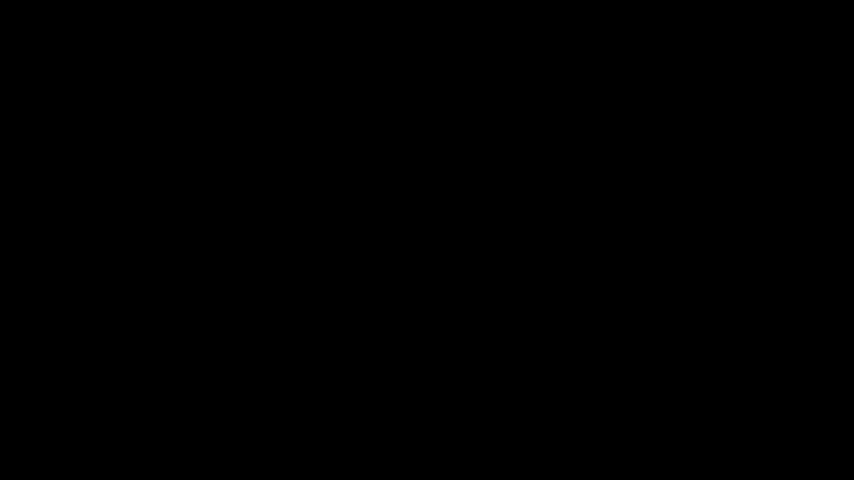}}\\
	\vspace{-4.2mm}&&&&&\\
	\rotatebox[origin=c]{90}{\tiny AFB-URR \hspace{1mm}}&
	\raisebox{-0.5\height}{\includegraphics[width=0.18\linewidth]{appendix/breakdance/00043.png}}&
	\raisebox{-0.5\height}{\includegraphics[width=0.18\linewidth]{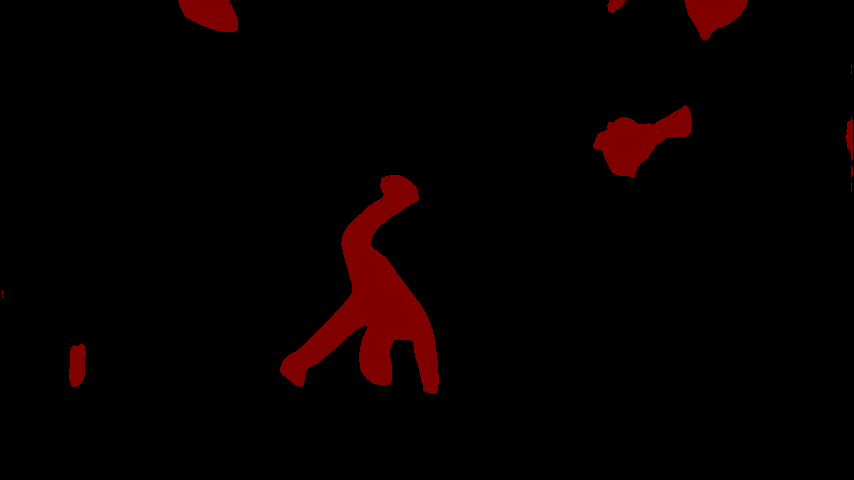}}&
	\raisebox{-0.5\height}{\includegraphics[width=0.18\linewidth]{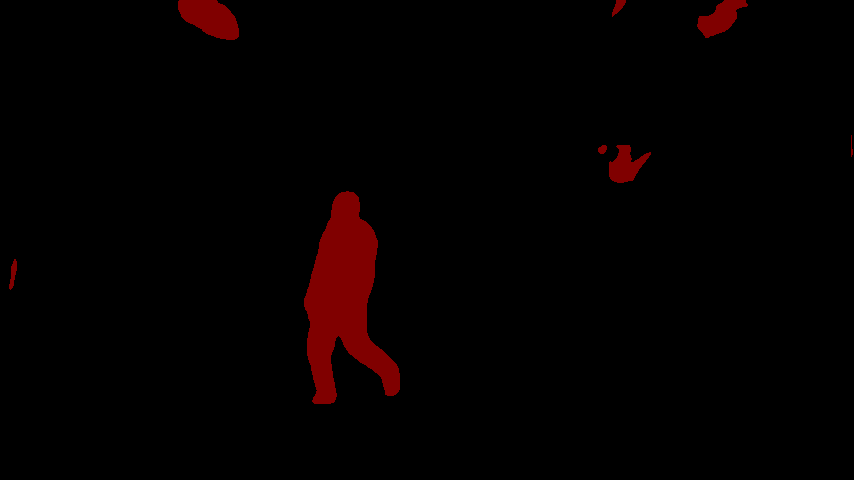}}&
	\raisebox{-0.5\height}{\includegraphics[width=0.18\linewidth]{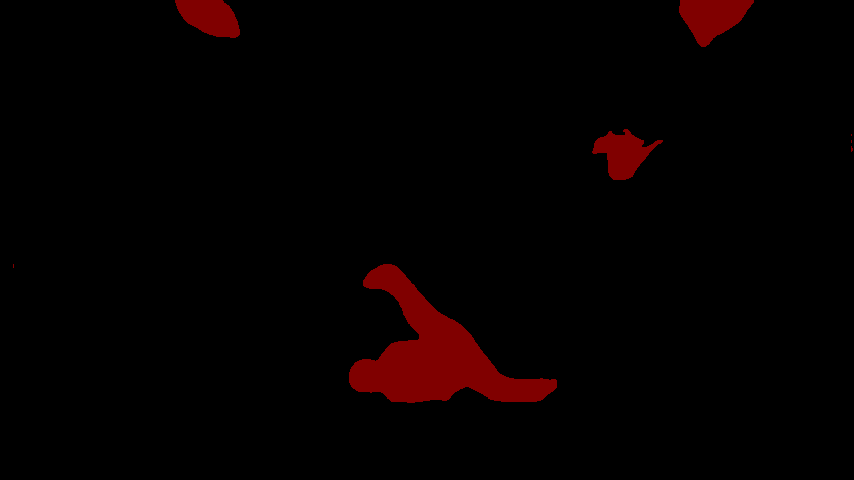}}&
	\raisebox{-0.5\height}{\includegraphics[width=0.18\linewidth]{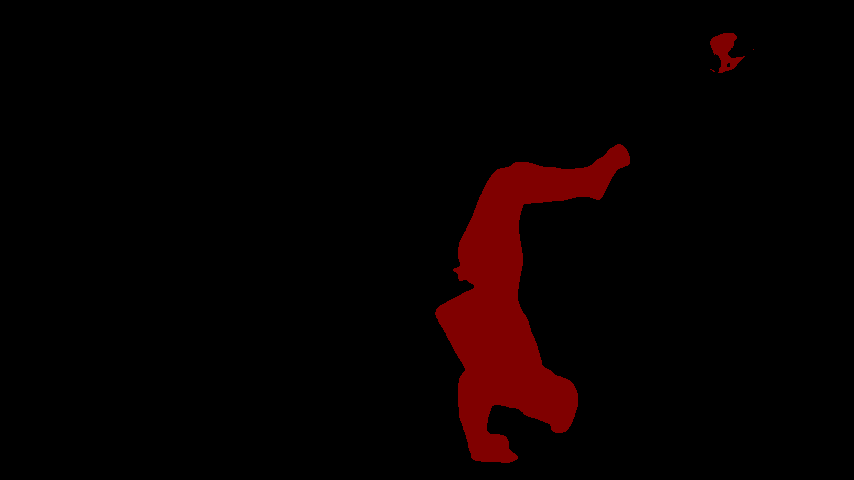}}\\
	\vspace{-3.5mm}&&&&&\\
	\rotatebox[origin=c]{90}{\scriptsize STCN \hspace{1mm}}&
	\raisebox{-0.5\height}{\includegraphics[width=0.18\linewidth]{appendix/breakdance/00043.png}}&
	\raisebox{-0.5\height}{\includegraphics[width=0.18\linewidth]{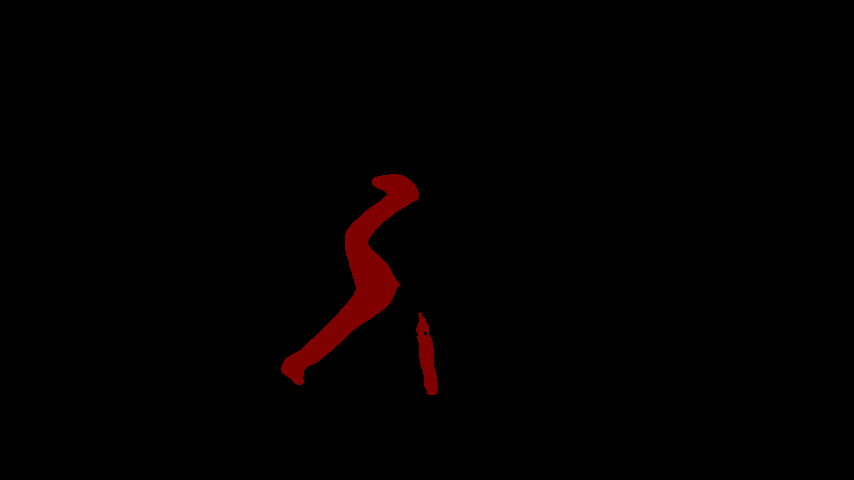}}&
	\raisebox{-0.5\height}{\includegraphics[width=0.18\linewidth]{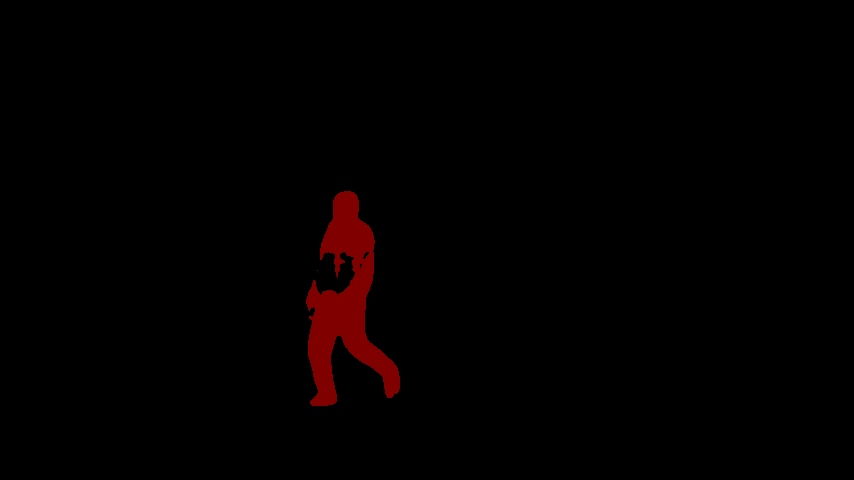}}&
	\raisebox{-0.5\height}{\includegraphics[width=0.18\linewidth]{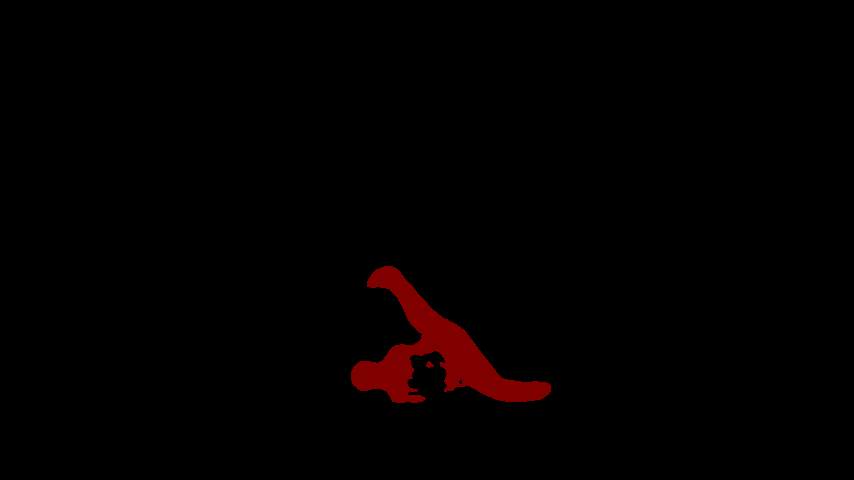}}&
	\raisebox{-0.5\height}{\includegraphics[width=0.18\linewidth]{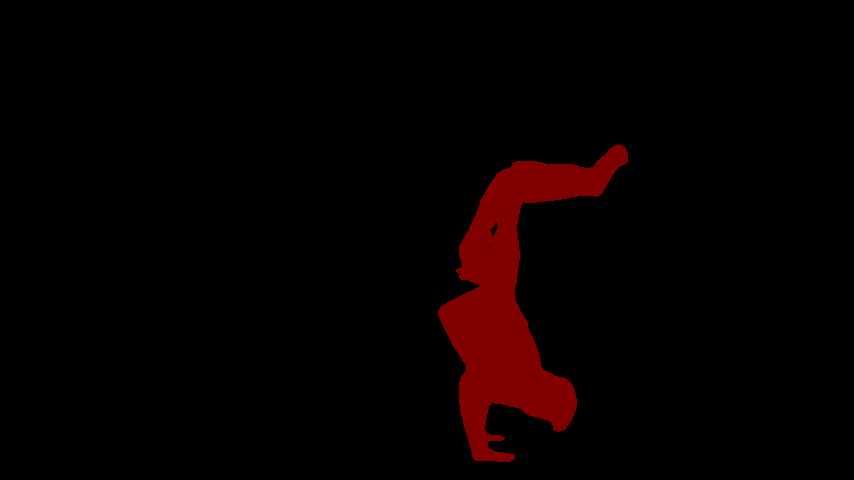}}\\
	\vspace{-3.5mm}&&&&&\\
	\rotatebox[origin=c]{90}{\scriptsize Ours \hspace{1mm}}&
	\raisebox{-0.5\height}{\includegraphics[width=0.18\linewidth]{appendix/breakdance/00043.png}}&
	\raisebox{-0.5\height}{\includegraphics[width=0.18\linewidth]{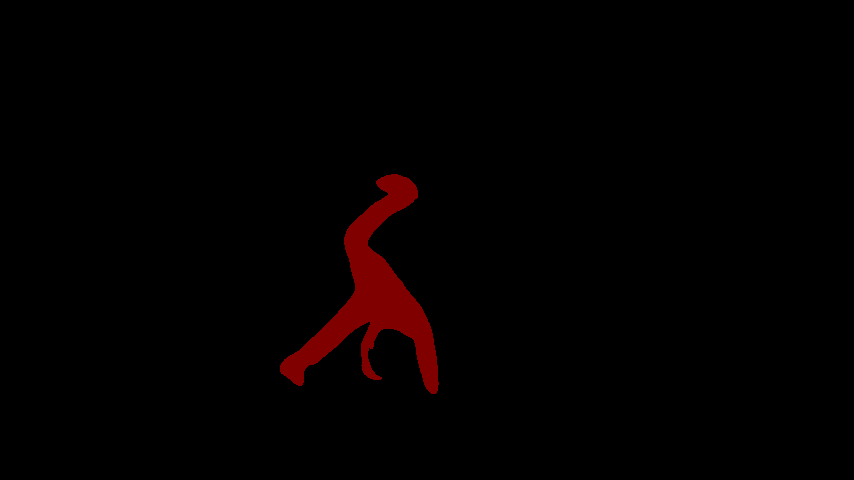}}&
	\raisebox{-0.5\height}{\includegraphics[width=0.18\linewidth]{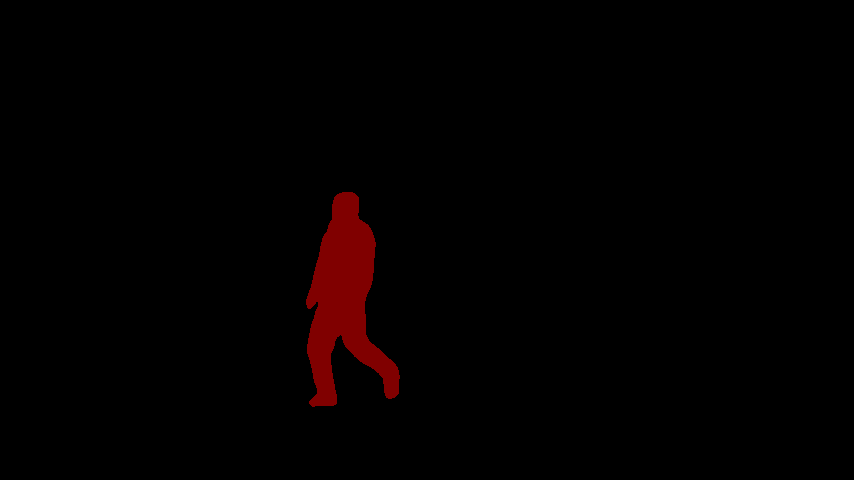}}&
	\raisebox{-0.5\height}{\includegraphics[width=0.18\linewidth]{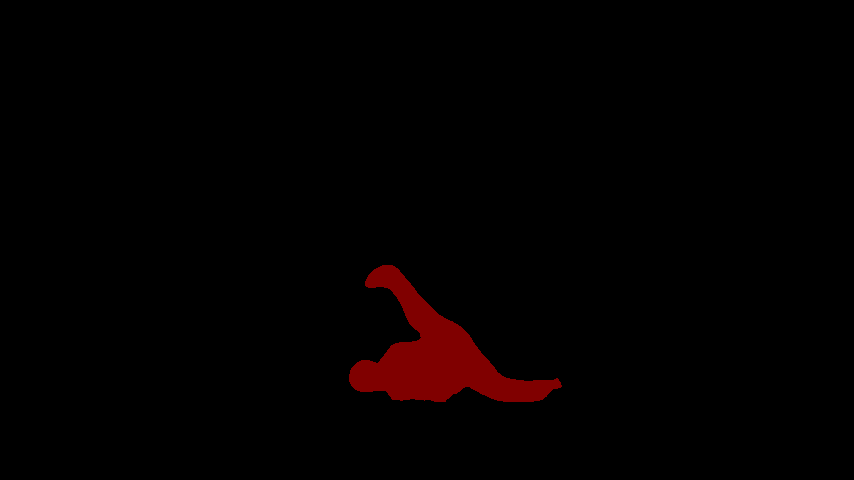}}&
	\raisebox{-0.5\height}{\includegraphics[width=0.18\linewidth]{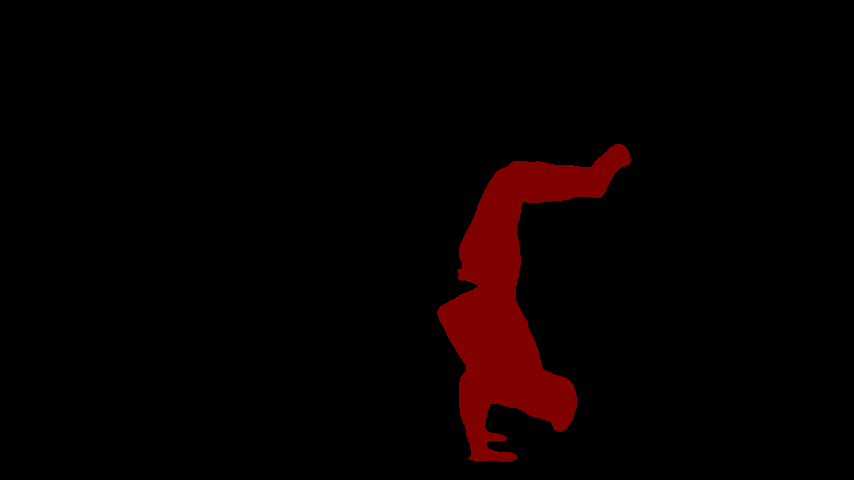}}\\
	\vspace{-3mm}&&&&&\\
	& Frame 0 (input) & Frame 2342 & Frame 4637 & Frame 6644 & Frame 18052 \\
\end{tabular}
	\vspace{-1em}
	\caption{Results on the {\tt breakdance} sequence.
		We manually annotated the first frame as input.
		Similarly to the {\tt dressage} sequence (Figure~\ref{fig:vis-dressage}), JOINT~\cite{mao2021joint} loses track over time, AFB-URR~\cite{Liang2020AFBURR} is overall less accurate, STCN~\cite{cheng2021stcn} struggles with fast motion, and our method performs well on this sequence.}
	\label{fig:vis-breakdance}

	\centering
	\begin{tabular}{c@{\hspace{1mm}}c@{\hspace{.5mm}}c@{\hspace{.5mm}}c@{\hspace{.5mm}}c@{\hspace{.5mm}}c}
	\rotatebox[origin=c]{90}{\scriptsize Images \hspace{1mm}}&
	\raisebox{-0.5\height}{\includegraphics[width=0.18\linewidth]{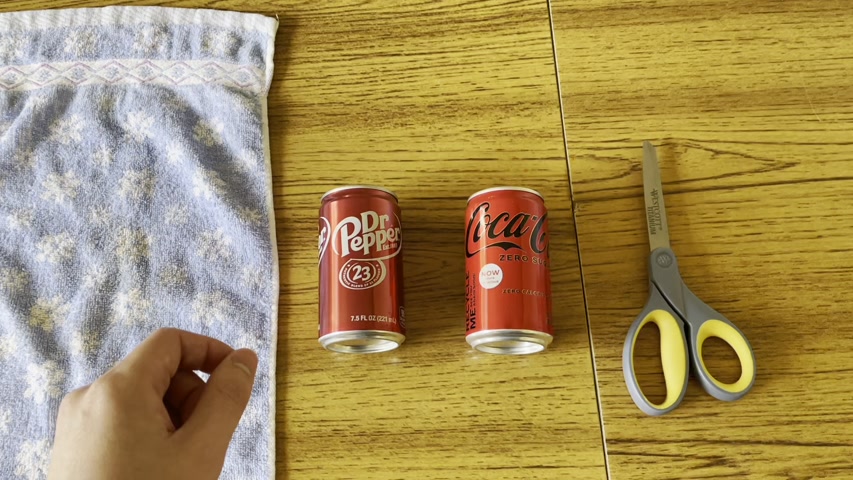}}&
	\raisebox{-0.5\height}{\includegraphics[width=0.18\linewidth]{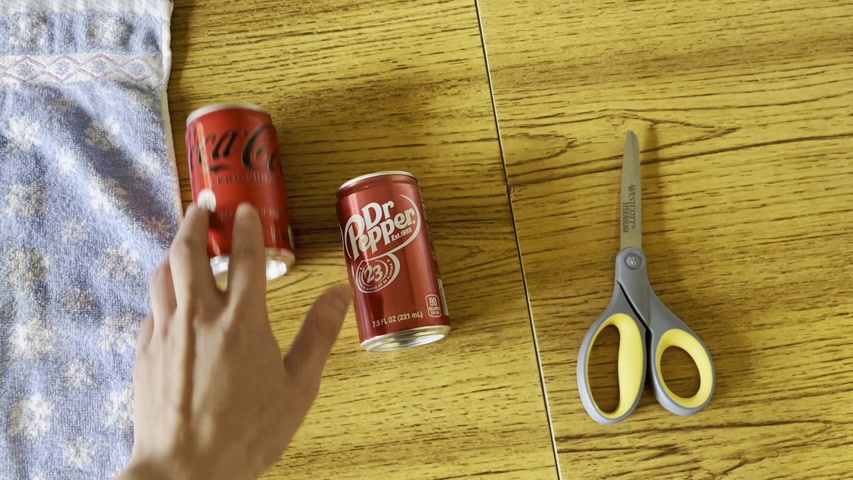}}&
	\raisebox{-0.5\height}{\includegraphics[width=0.18\linewidth]{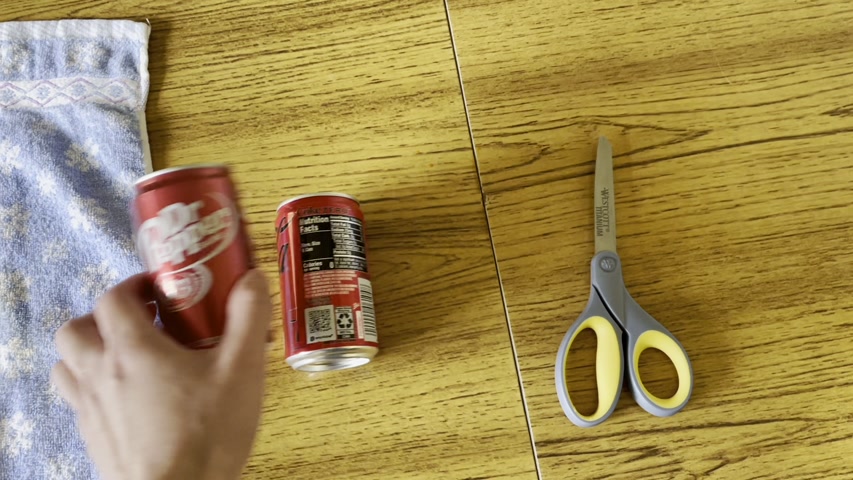}}&
	\raisebox{-0.5\height}{\includegraphics[width=0.18\linewidth]{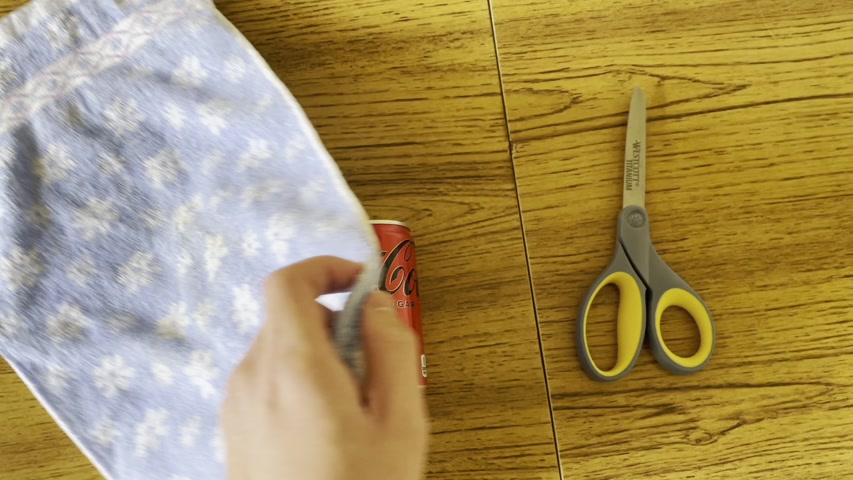}}&
	\raisebox{-0.5\height}{\includegraphics[width=0.18\linewidth]{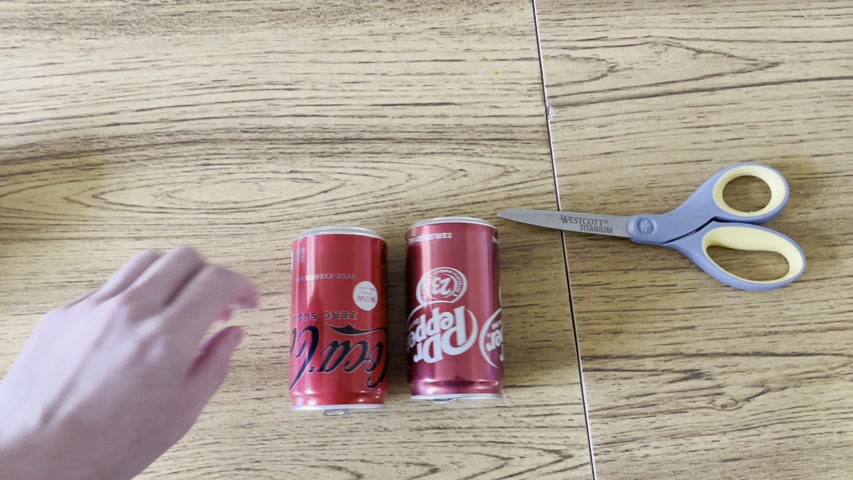}}\\
	\vspace{-3.5mm}&&&&&\\
	\rotatebox[origin=c]{90}{\scriptsize JOINT \hspace{1mm}}&
	\raisebox{-0.5\height}{\includegraphics[width=0.18\linewidth]{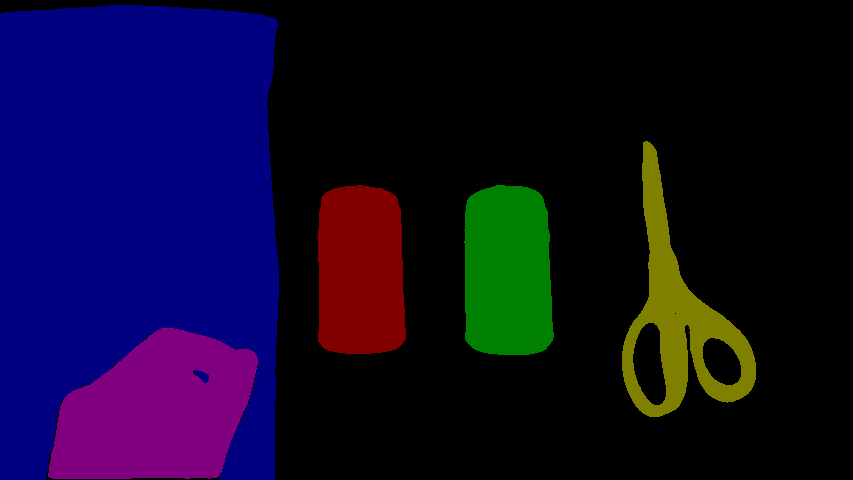}}&
	\raisebox{-0.5\height}{\includegraphics[width=0.18\linewidth]{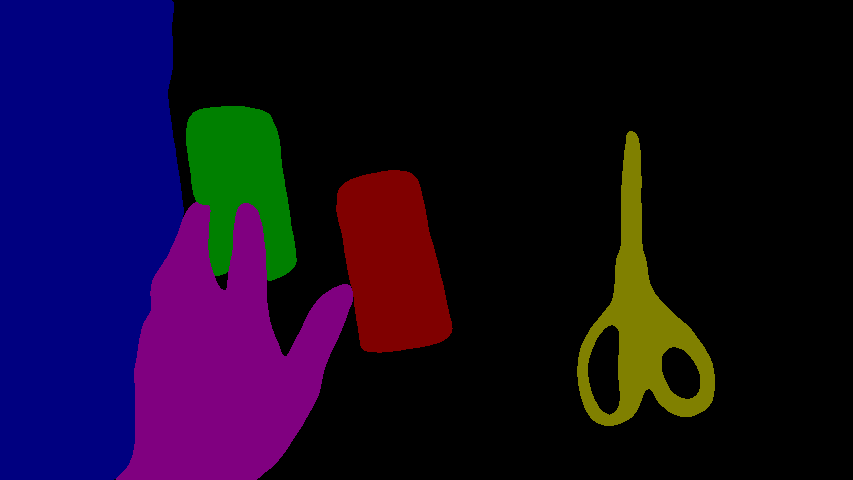}}&
	\raisebox{-0.5\height}{\includegraphics[width=0.18\linewidth]{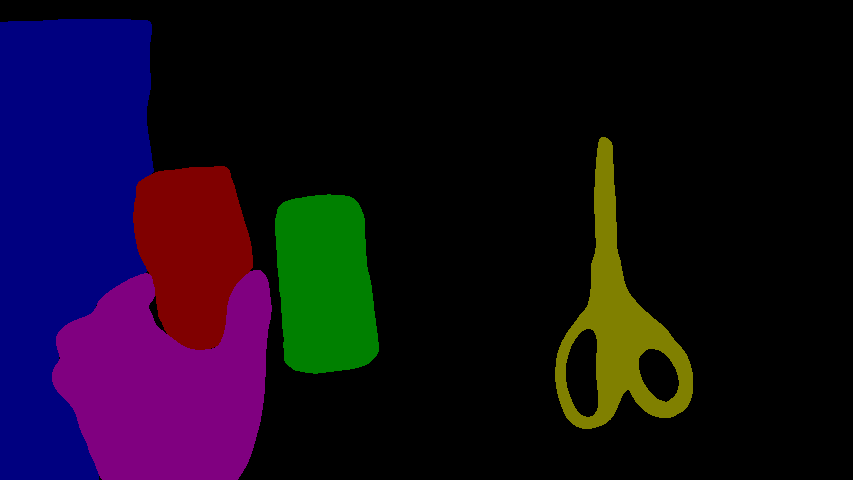}}&
	\raisebox{-0.5\height}{\includegraphics[width=0.18\linewidth]{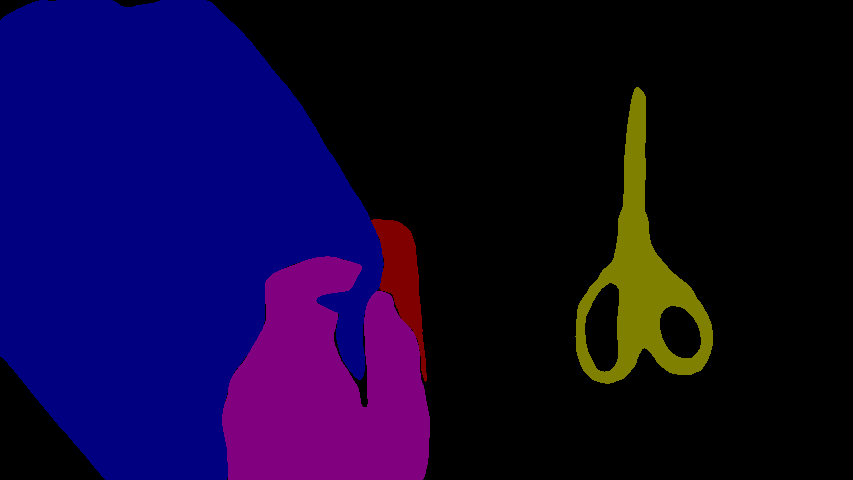}}&
	\raisebox{-0.5\height}{\includegraphics[width=0.18\linewidth]{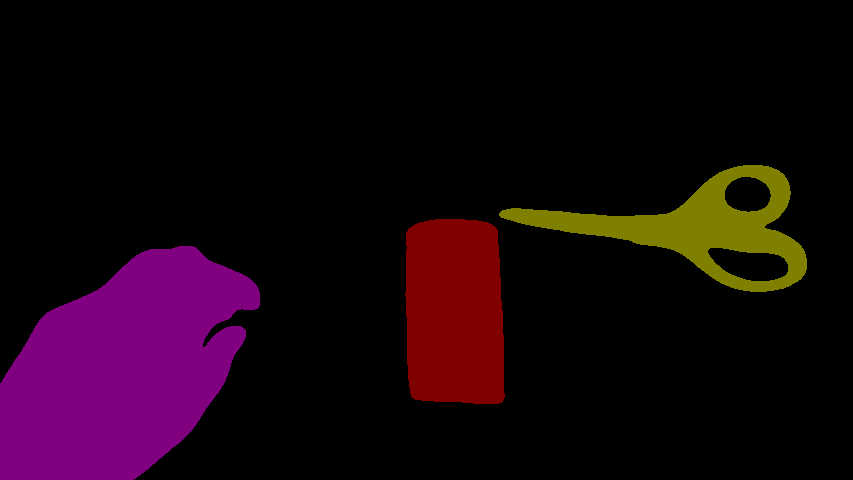}}\\
	\vspace{-4.2mm}&&&&&\\
	\rotatebox[origin=c]{90}{\tiny AFB-URR \hspace{1mm}}&
	\raisebox{-0.5\height}{\includegraphics[width=0.18\linewidth]{appendix/cans/00001.png}}&
	\raisebox{-0.5\height}{\includegraphics[width=0.18\linewidth]{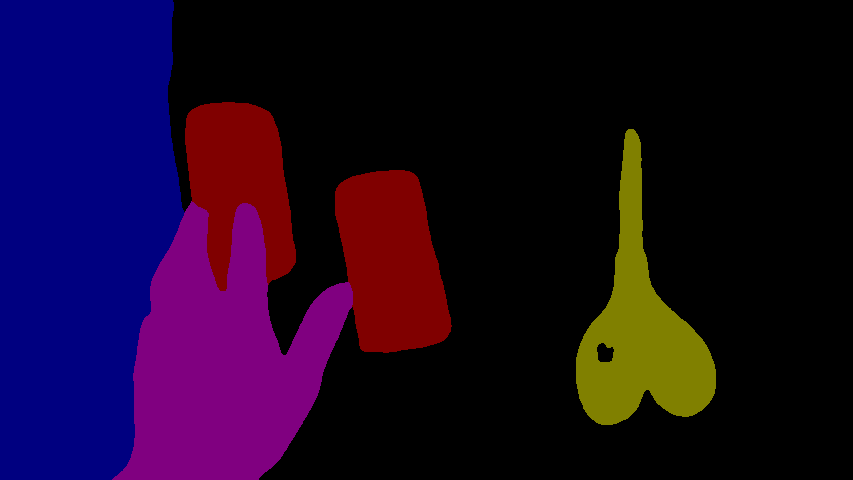}}&
	\raisebox{-0.5\height}{\includegraphics[width=0.18\linewidth]{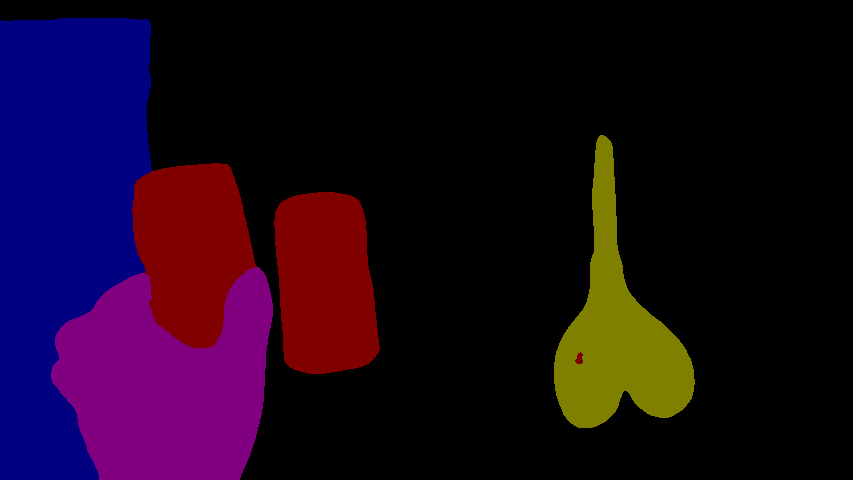}}&
	\raisebox{-0.5\height}{\includegraphics[width=0.18\linewidth]{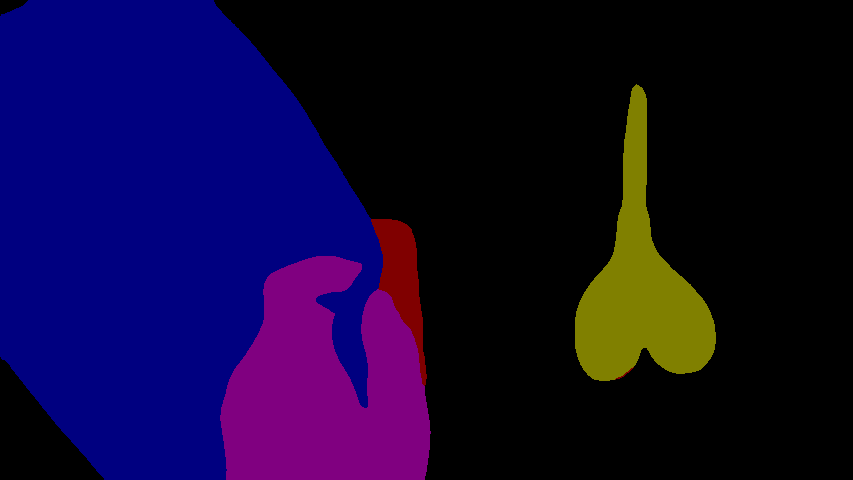}}&
	\raisebox{-0.5\height}{\includegraphics[width=0.18\linewidth]{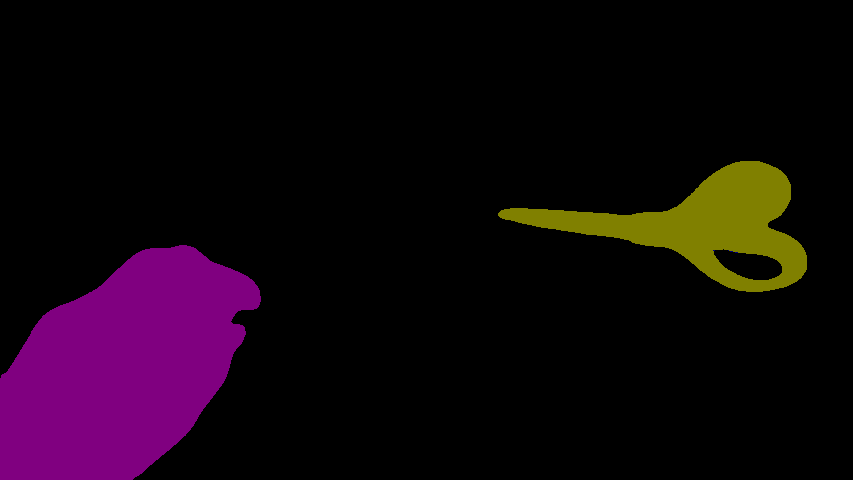}}\\
	\vspace{-3.5mm}&&&&&\\
	\rotatebox[origin=c]{90}{\scriptsize STCN \hspace{1mm}}&
	\raisebox{-0.5\height}{\includegraphics[width=0.18\linewidth]{appendix/cans/00001.png}}&
	\raisebox{-0.5\height}{\includegraphics[width=0.18\linewidth]{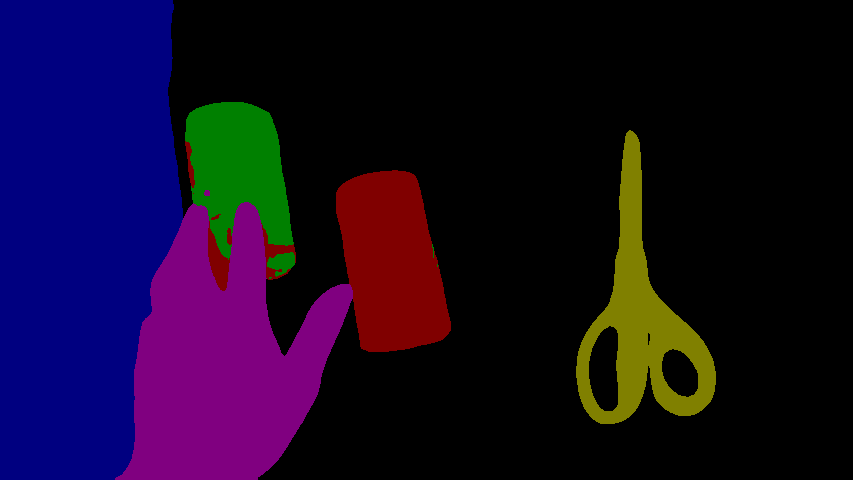}}&
	\raisebox{-0.5\height}{\includegraphics[width=0.18\linewidth]{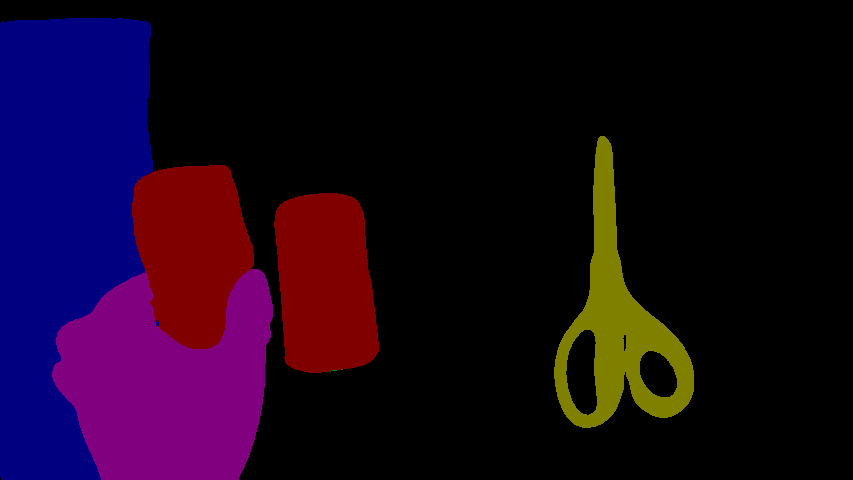}}&
	\raisebox{-0.5\height}{\includegraphics[width=0.18\linewidth]{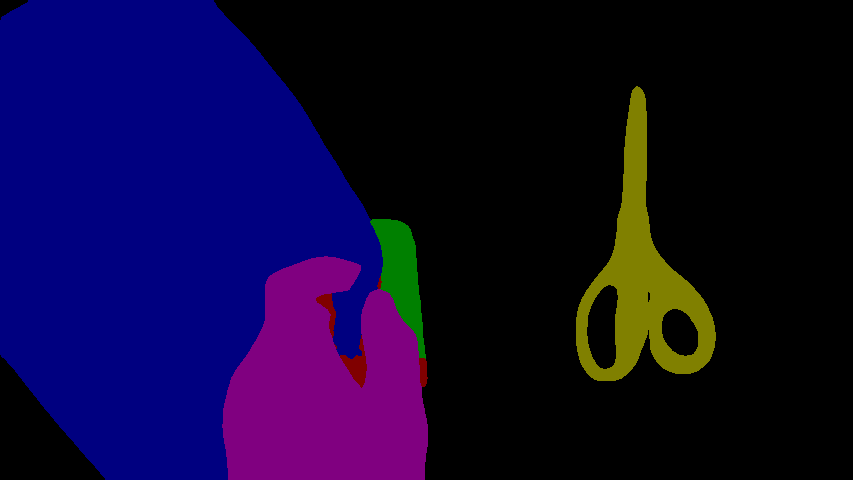}}&
	\raisebox{-0.5\height}{\includegraphics[width=0.18\linewidth]{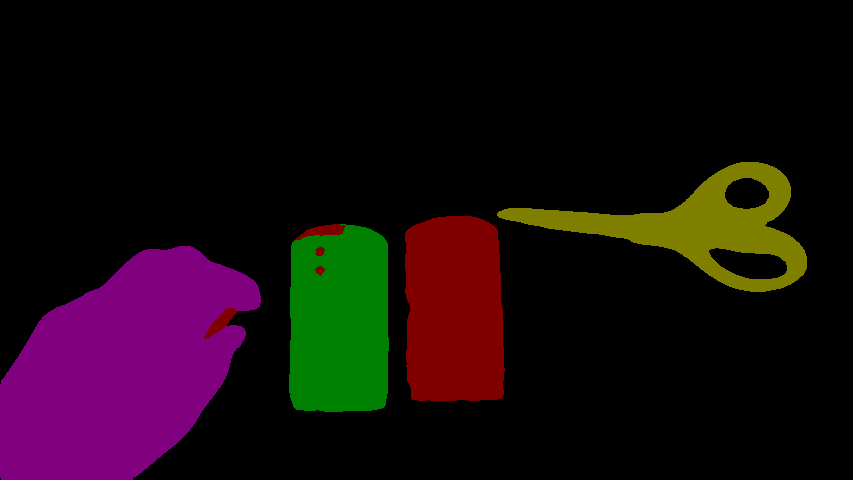}}\\
	\vspace{-3.5mm}&&&&&\\
	\rotatebox[origin=c]{90}{\scriptsize Ours \hspace{1mm}}&
	\raisebox{-0.5\height}{\includegraphics[width=0.18\linewidth]{appendix/cans/00001.png}}&
	\raisebox{-0.5\height}{\includegraphics[width=0.18\linewidth]{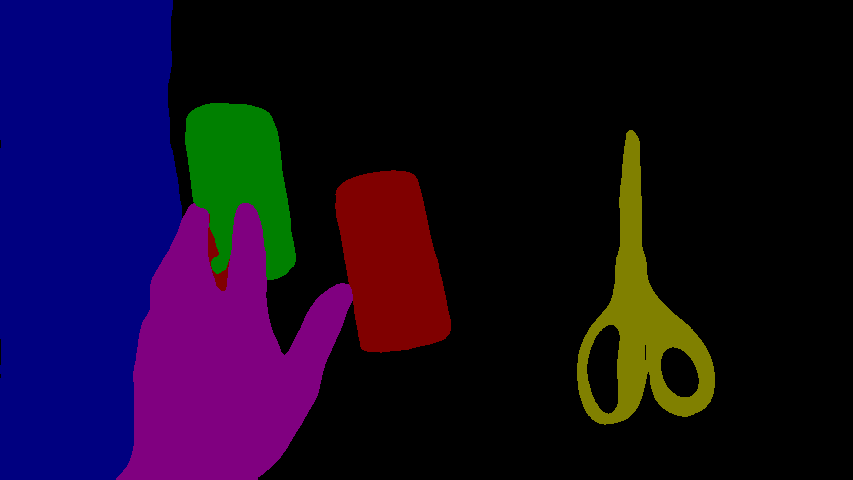}}&
	\raisebox{-0.5\height}{\includegraphics[width=0.18\linewidth]{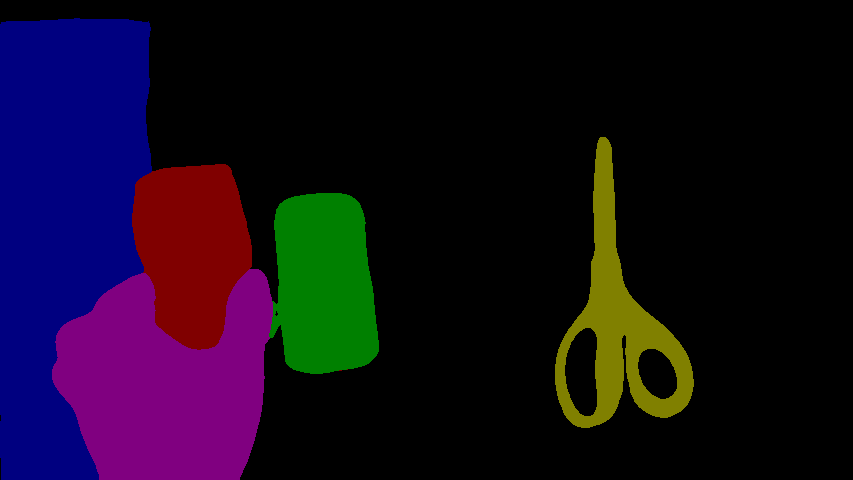}}&
	\raisebox{-0.5\height}{\includegraphics[width=0.18\linewidth]{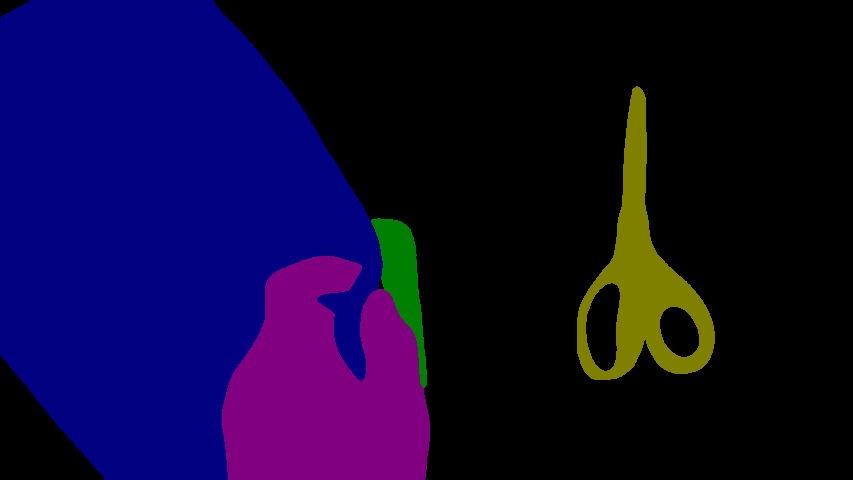}}&
	\raisebox{-0.5\height}{\includegraphics[width=0.18\linewidth]{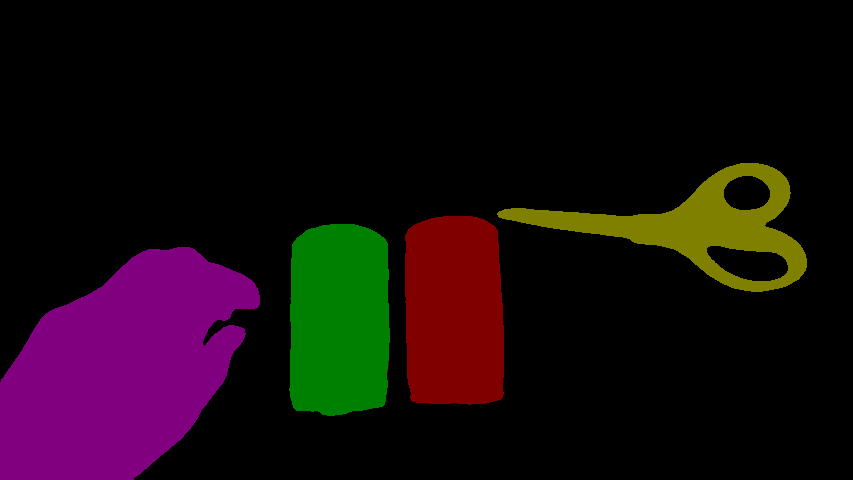}}\\
	\vspace{-3mm}&&&&&\\
	& Frame 0 (input) & Frame 680 & Frame 1062 & Frame 1793 & Frame 3980 \\
\end{tabular}
	\vspace{-1em}
	\caption{Results on the {\tt cans} sequence.
		We manually annotated the first frame as input.
		The Dr.~Pepper can is labeled with red, and the Coca-Cola can is labeled with green. 
		The two cans are completely occluded with a towel after frame 1,793, and reappear about 2,000 frames later.
		The color tone change is due to the camera's auto white balance.
		JOINT~\cite{mao2021joint} misses the Coca-Cola can after occlusion. It still captures the Dr.~Pepper can as  the available first reference frame  helps.
		AFB-URR~\cite{Liang2020AFBURR} mixes up the two cans early on, and fails to capture them after they reappear. This is due to its eager feature compression and thus lower modeling capability.
		STCN~\cite{cheng2021stcn} uses a low memory insertion frequency to avoid memory explosion which causes it to be less accurate when changes happen.
		Our method is the most accurate overall. 
	}
	\label{fig:vis-cans}
	\vspace{-1em}
\end{figure}

Here, we compare qualitatively to JOINT~\cite{mao2021joint}, AFB-URR~\cite{Liang2020AFBURR}, and STCN~\cite{cheng2021stcn} using several long videos and using the same setting as in the paper.
We show results on the {\tt dressage} sequence (10,767 frames) which is part of the Long-time Video (3$\times$) dataset~\cite{Liang2020AFBURR}, and two additional in-the-wild videos. 
{\tt breakdance} contains a single foreground object with large and fast motion, and has 18,187 frames.
{\tt cans} is very challenging, contains five different objects, two of which (Dr.~Pepper and Coca-Cola cans) are similar. 
The two cans are completely occluded for more than 2,000 frames, and our method can successfully capture them when they reappear. Figure~\ref{fig:vis-dressage},~\ref{fig:vis-breakdance}, and~\ref{fig:vis-cans}  compare results on these videos respectively.
We show one potential application where an image layer is inserted between the foreground and the background using the predicted object mask on a snippet of the {\tt breakdance} sequence.


\section{Failure Cases}
\label{sec:app:failure}
\begin{figure}
	\centering
	\begin{tabular}{c@{\hspace{.5mm}}c@{\hspace{.5mm}}c@{\hspace{.5mm}}c}
	\includegraphics[width=0.24\linewidth]{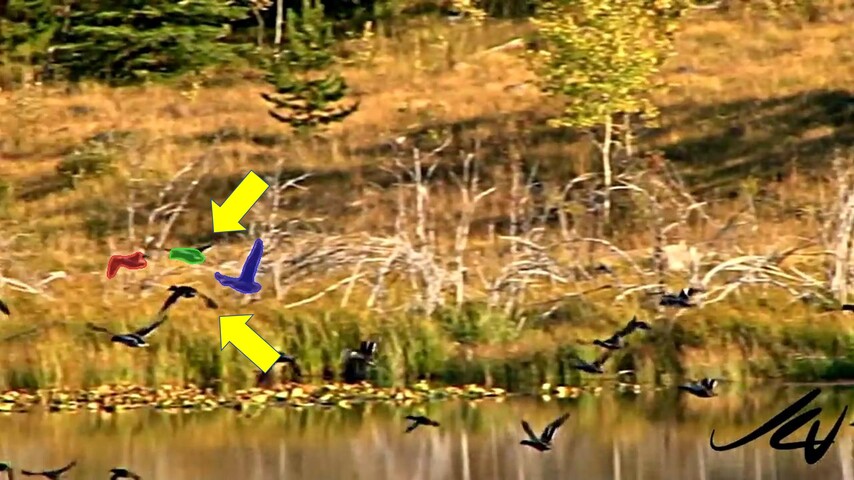} &
	\includegraphics[width=0.24\linewidth]{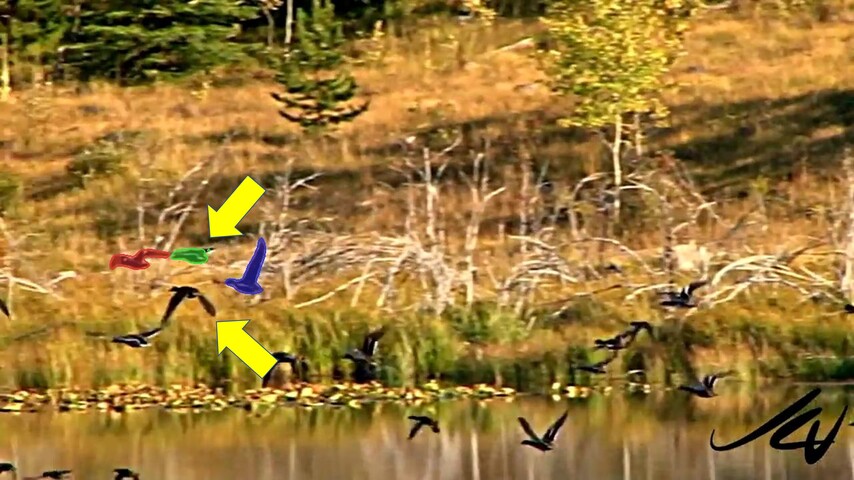} & 
	\includegraphics[width=0.24\linewidth]{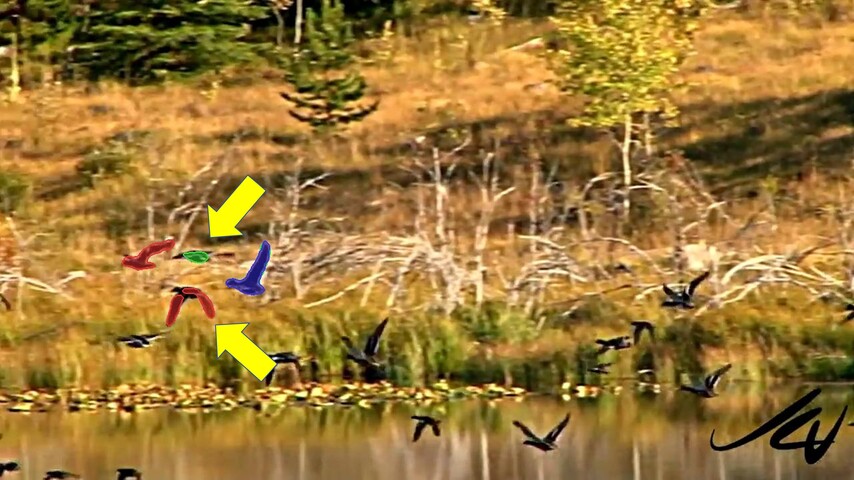} & 
	\includegraphics[width=0.24\linewidth]{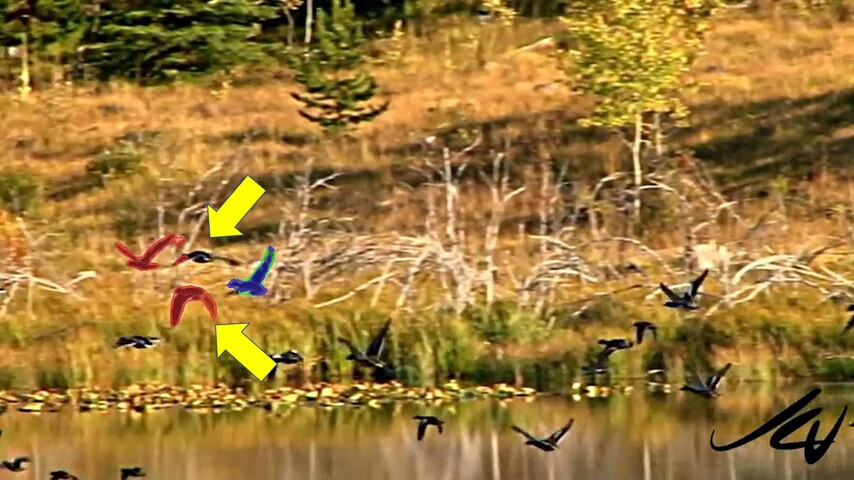} \\
	\vspace{-4.5mm}&&&\\
	\includegraphics[width=0.24\linewidth]{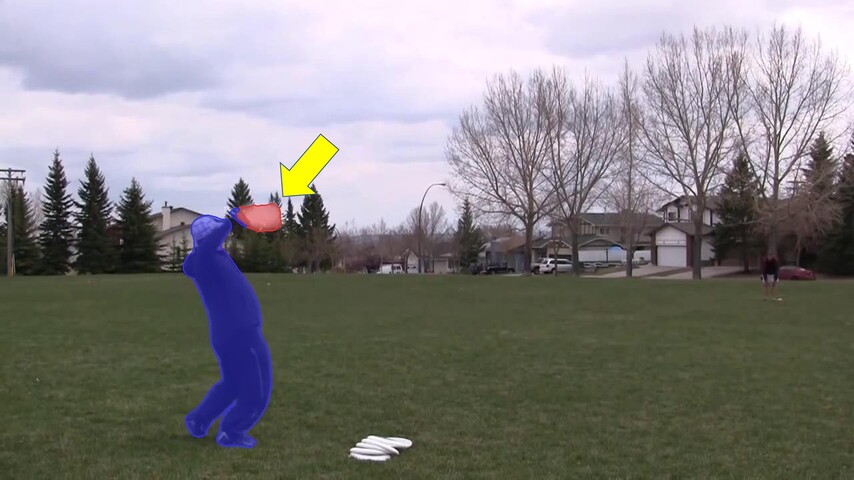} & 
	\includegraphics[width=0.24\linewidth]{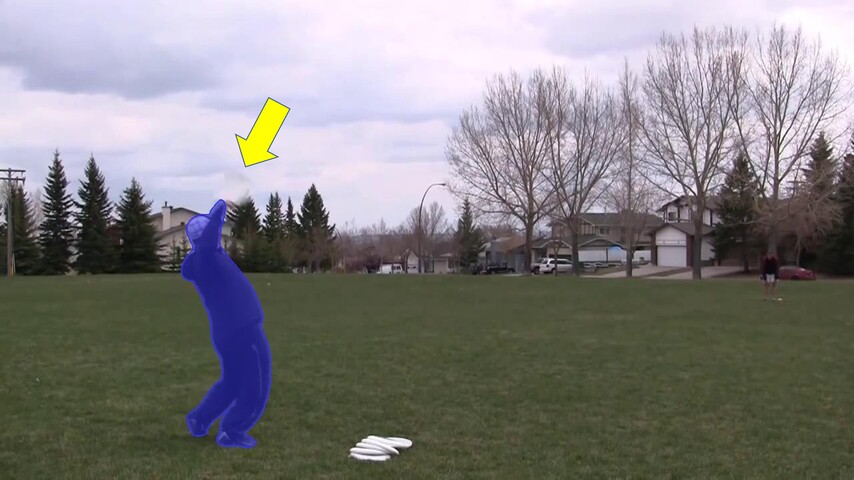} & 
	\includegraphics[width=0.24\linewidth]{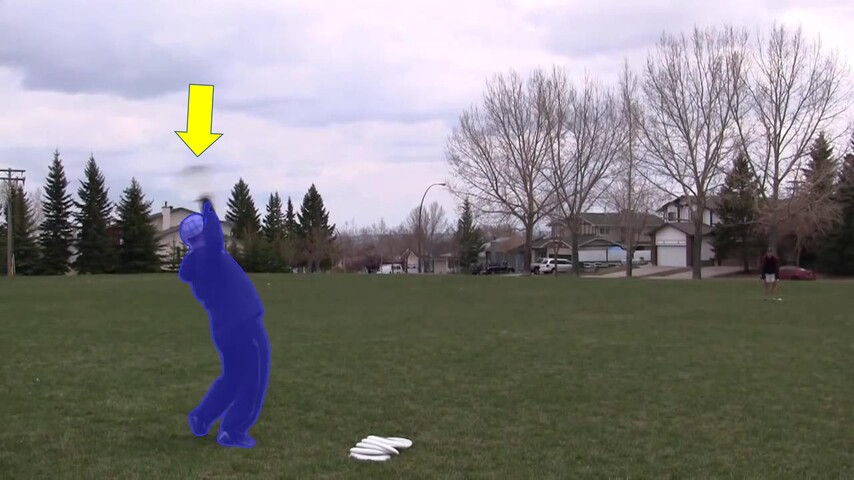} & 
	\includegraphics[width=0.24\linewidth]{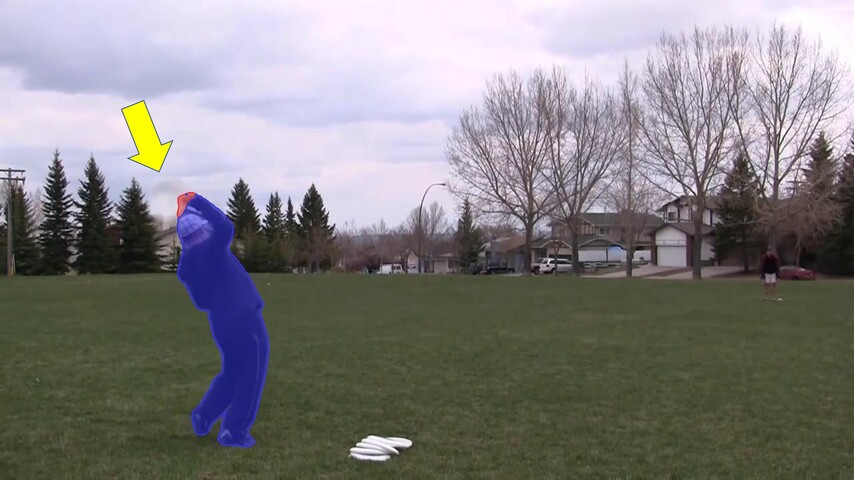} \\
	\vspace{-4.5mm}&&&\\
	\includegraphics[width=0.24\linewidth]{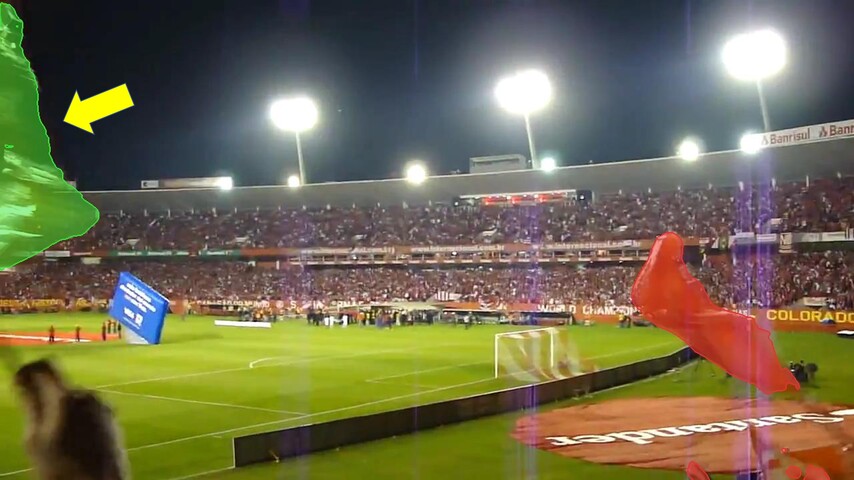} &
	\includegraphics[width=0.24\linewidth]{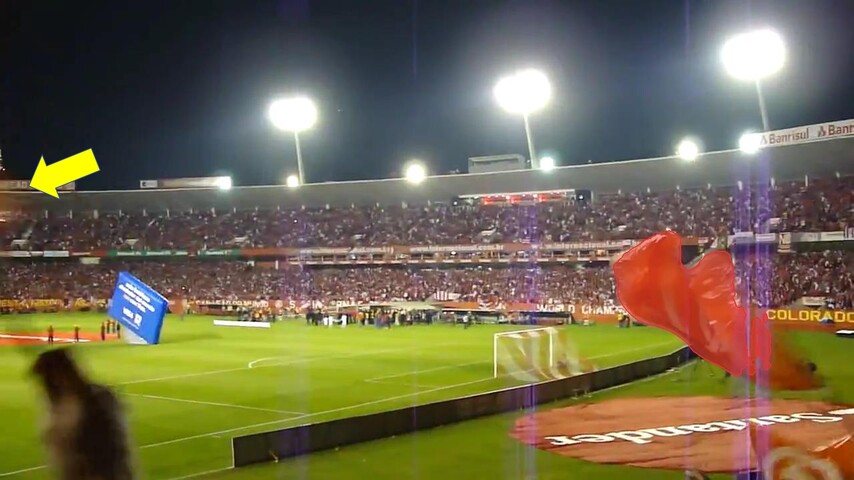} & 
	\includegraphics[width=0.24\linewidth]{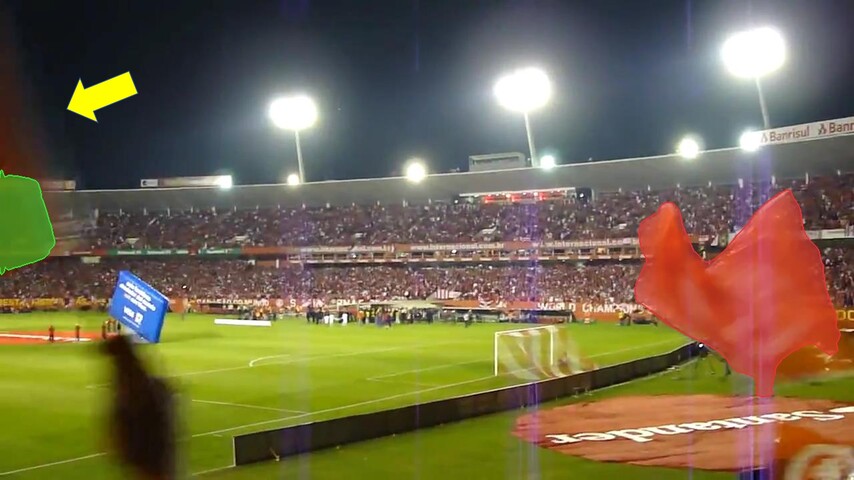} & 
	\includegraphics[width=0.24\linewidth]{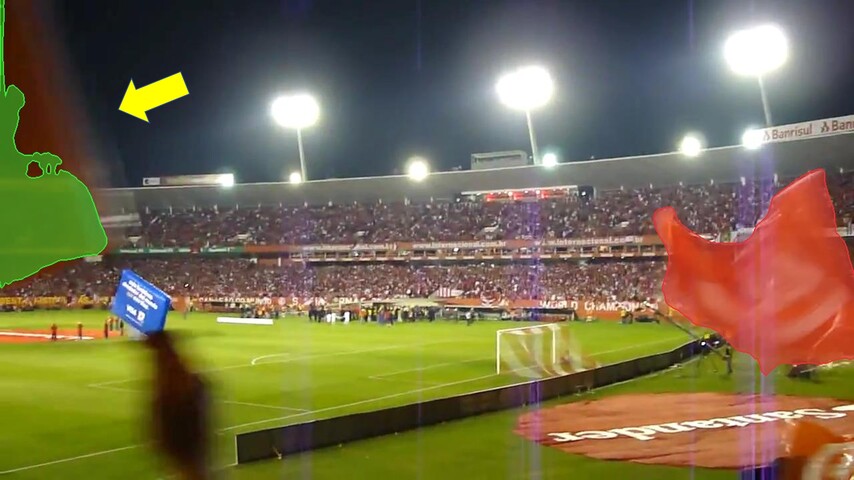} \\
\end{tabular}

	\caption{Failure cases. We point to objects of interest with an arrow.
	First row: multiple birds with similar appearances are flying. We fail to discriminate between some birds that are close to each other.
	Second row: a frisbee is being thrown. We cannot catch up as it is moving quickly with a large motion blur.
	Third row: two flags are being waved quickly. We fail to segment the whole left flag due to fast motion.}
	\label{fig:failure}
\end{figure}

As mentioned in the limitation section of the main text, our method struggles with very fast moving objects.
This is because even the fastest-updating sensory memory fails to track such objects, and the working memory fails to model objects with large motion blur.
Figure~\ref{fig:failure} visualizes some failure cases.

\section{Long-Term Memory Size and FPS Scaling}
\label{sec:app:scaling}
By default, we use a maximum long-term memory size of 10,000 which consumes a small amount of GPU memory and is reasonably capable -- it can store information from around 3,900 frames after memory consolidation ($r=10$).
In practice, users might opt for a different upper limit of the long-term memory (LT$_{\max}$), in consideration of any memory constraints, speed, and the complexity of the video.
Here, we test the performance of different LT$_{\max}$ settings on the Long-time Video (3$\times$) dataset~\cite{Liang2020AFBURR} and show the results in Table~\ref{tab:diff-lt-size}.
There is significant memory saving and speed-up when LT$_{\max}$ is decreased. 
While a smaller LT$_{\max}$ seems to be sufficient for this dataset, we expect using a higher LT$_{\max}$ can benefit more challenging videos with long-term occlusions.

\begin{table}[h]
	\centering
	\caption{Performance of XMem with different upper limits of the long-term memory LT$_{\max}$ on the Long-time Video (3$\times$) dataset~\cite{Liang2020AFBURR}.}
	\begin{tabular}{l@{\hspace{5mm}}c@{\hspace{5mm}}c@{\hspace{5mm}}c}
		\toprule
		LT$_{\max}$ & \mjf & Max. GPU memory & FPS\\
		\midrule
		500 & 87.2$\pm4.7$ & 1168 MB & 35.3 \\
		1,000 & 89.5$\pm0.3$ & 1186 MB & 34.1 \\
		2,500 & 89.8$\pm0.2$ & 1243 MB & 31.1 \\
		5,000 & 89.9$\pm0.2$ & 1332 MB & 27.5 \\
		10,000 & 90.0$\pm0.4$ & 1515 MB & 23.4 \\
		20,000 & 90.0$\pm0.4$ & 1632 MB & 21.1 \\
		30,000 & 90.0$\pm0.4$ & 1632 MB & 20.9 \\
		\midrule
		\bottomrule
	\end{tabular}
	\label{tab:diff-lt-size}
\end{table}

\begin{figure}[h]
	\centering
	\resizebox{!}{0.4\linewidth}{
		\input{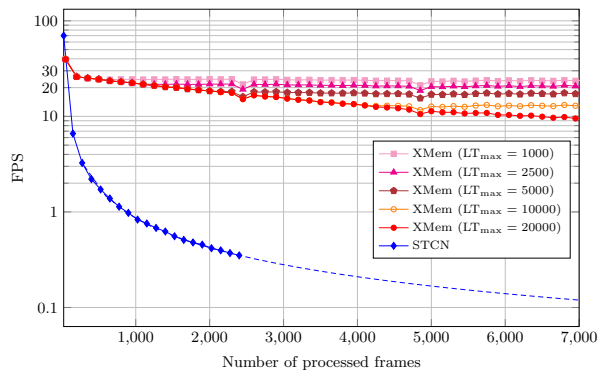}
	}
	\caption{FPS scaling of STCN~\cite{cheng2021stcn} vs.\ variants of XMem.
		STCN starts off faster due to its simpler construction but slows down drastically as its memory bank expands. 
		As STCN soon becomes too slow for practical use, we estimate its FPS by fitting a linear function to its processing time (i.e., inverse linear to FPS). This linear function is illustrated with a dashed line.
		XMem maintains a relatively stable and fast FPS throughout, thanks to our memory consolidation algorithm. 
		}
	\label{fig:fps-scaling}
\end{figure}

We also plot the single-object FPS (1/time required to process a new frame) against the total number of processed frames for STCN~\cite{cheng2021stcn} and different LT$_{\max}$ settings of XMem in Figure~\ref{fig:fps-scaling}.
We use a high-capacity 32GB V100 GPU in this experiment such that STCN can be run without out-of-memory errors.
FPSs for XMem plateaue after reaching LT$_{\max}$.

\section{Re-training STCN}
\label{sec:app:retrain}
We have changed the training schedule (see Section 3.6, Implementation Details) and adjusted parts of the network (including removing some convolutional layers and adding a feature fusion block~\cite{cheng2021stcn} to the decoder for incorporating the sensory memory). For a fair comparison, we re-train the STCN~\cite{cheng2021stcn} baseline under our setting.
This is equivalent to removing the sensory memory, long-term memory, and both scaling terms.
BL30K~\cite{cheng2021mivos} is not used.
Table~\ref{tab:re-train-stcn} tabulates the results on the YouTubeVOS 2018 validation set~\cite{xu2018youtubeVOS} and the DAVIS 2017 validation set~\cite{Pont-Tuset_arXiv_2017}.
The re-trained method achieves a $1.2$ higher~\mjf~on DAVIS and a $1.0$ higher~\mg~on YouTubeVOS.
On average, the change is insignificant, i.e., our training schedule is not a sufficient condition for improved results.

\begin{table}
	\centering
	\caption{We compare the performance of STCN~\cite{cheng2021stcn} to the re-trained version with our training setup.
	On average, there is no significant difference.
	The performance of XMem is provided as a reference.}
	\begin{tabular}{lc@{\hspace{5mm}}c}
		\toprule
		Method & YouTubeVOS 2018 val \mg & DAVIS 2017 val \mjf\\
		\midrule
		STCN~\cite{cheng2021stcn} (original) & 83.0 & 85.4  \\
		STCN~\cite{cheng2021stcn} (re-trained) & 84.0 & 84.2 \\
		XMem (Ours) & \textbf{85.7} & \textbf{86.2} \\
		\midrule
		\bottomrule
	\end{tabular}
	\label{tab:re-train-stcn}
\end{table}

\section{Results on YouTubeVOS 2019 validation}
\label{sec:app:yv-2019}

Table~\ref{tab:app:y19} tabulates our results on the YouTubeVOS~\cite{xu2018youtubeVOS} 2019 validation set. We compare the measured FPS on the 2018 version. The FPS on these two versions are highly correlated as their average number of objects and video length are similar.

\begin{table}
	\centering
	\caption{Quantitative comparisons on YouTubeVOS 2019 validation.}
	\begin{tabular}{l@{\hspace{7pt}}c@{\hspace{7pt}}c@{\hspace{7pt}}c@{\hspace{7pt}}c@{\hspace{7pt}}c@{\hspace{7pt}}c}
	\toprule
	& \multicolumn{6}{c}{YouTubeVOS 2019 val~\cite{xu2018youtubeVOS}} \\
	\cmidrule(lr{\dimexpr 4\tabcolsep+5pt}){2-6} 
	Method & \mg & \mjs & \mfs & \mju & \mfu & FPS$_{\text{Y18}}$ \\
	\midrule
	CFBI~\cite{yang2020collaborativeCFBI} & 81.0 & 80.6 & 85.1 & 75.2 & 83.0 & 3.4 \\
	SST~\cite{duke2021sstvos} & 81.8 & 80.9 & - & 76.7 & - & - \\
	MiVOS$^{\ast}$~\cite{cheng2021mivos} & 82.4 & 80.6 & 84.7 & 78.1 & 86.4 & - \\
	HMMN~\cite{seong2021hierarchical} & 82.5 & 81.7 & 86.1 & 77.3 & 85.0 & - \\
	CFBI+~\cite{yang2020CFBIP} & 82.6 & 81.7 & 86.2 & 77.1 & 85.2 & 4.0 \\
	STCN~\cite{cheng2021stcn} & 82.7 & 81.1 & 85.4 & 78.2 & 85.9 & 13.2 \\
	JOINT~\cite{mao2021joint} & 82.8 & 80.8 & 84.8 & 79.0 & 86.6 & - \\
	STCN$^{\ast}$~\cite{cheng2021stcn} & 84.2 & 82.6 & 87.0 & 79.4 & 87.7 & 13.2 \\
	AOT~\cite{yang2021associating} & 85.3 & 83.9 & 88.8 & 79.9 & 88.5 & 6.4 \\
	XMem~(Ours) & 85.5 & 84.3 & 88.6 & \textbf{80.3} & 88.6 & \textbf{22.6} \\
	XMem$^{\ast}$~(Ours) & \textbf{85.8} & \textbf{84.8} & \textbf{89.2} & \textbf{80.3} & \textbf{88.8} & \textbf{22.6} \\
	\midrule
	\bottomrule
\end{tabular}
	\label{tab:app:y19}
\end{table}

\section{Results with Different Training Datasets}
\label{sec:app:no-pretrain}

Following prior works~\cite{oh2019videoSTM}, we first pretrain our network on static images. As in the implementation of~\cite{cheng2021mivos,cheng2021stcn}, we use a mix of single object datasets~\cite{shi2015hierarchicalECSSD,wang2017DUTS,FSS1000,zeng2019towardsHRSOD,cheng2020cascadepsp}.
We compare with prior works that do not use pretraining in Table~\ref{tab:app:no-pretrain}.
We additionally present detailed results of our method when it is trained on 1) DAVIS 2017~\cite{perazzi2016benchmark} only, 2) YouTubeVOS 2019~\cite{xu2018youtubeVOS} only, and 3) a mix of both, in the followings tables.

\begin{table}
	\centering
	\caption{Comparisons with methods without static image pretraining. }
	\begin{tabular}{lc@{\hspace{5pt}}c@{\hspace{5pt}}c@{\hspace{5pt}}c@{\hspace{5pt}}c@{}c@{}}
	\toprule
	Method & Y$_{\text{18}}$ & Y$_{\text{19}}$ & D$_{\text{16}}$ & D$_{\text{17}}$ & D$_{\text{17td}}$ & FPS$_{\text{D17}}$\\
	\midrule
	LWL~\cite{bhat2020learningLWL} & 81.5 & 81.0 & - & 81.6 & - & - \\
	SST~\cite{duke2021sstvos} & 81.7 & 81.8 & - & 82.5 & - & - \\
	CFBI+~\cite{yang2020CFBIP} & 82.0 & 82.6 & 89.9 & 82.9 & 78.0 & 5.6 \\
	JOINT~\cite{mao2021joint} & 83.1 & 82.8 & - & 83.5 & - & 6.8 \\
	Ours$^{-}$ & \textbf{84.3} & \textbf{84.2} & \textbf{90.8} & \textbf{84.5}& \textbf{79.8} & \textbf{20.2} \\
	\midrule
	\bottomrule
\end{tabular}
	\label{tab:app:no-pretrain}
\end{table}

\begin{table}
	\centering
	\caption{Performance of XMem on DAVIS 2016 with different training data.}
	\begin{tabular}{lc@{\hspace{5pt}}c@{\hspace{5pt}}c}
		\toprule
		Training data & \mjf & \mj & \mf \\
		\midrule
		DAVIS only & 87.8 & 86.7 & 88.9 \\
		DAVIS+YouTubeVOS only & 90.8 & 89.6 & 91.9 \\
		Static+DAVIS+YouTubeVOS & 91.5 & 90.4 & 92.7 \\
		Static+BL30K+DAVIS+YouTubeVOS & 92.0 & 90.7 & 93.2 \\
		\midrule
		\bottomrule
	\end{tabular}
	\label{tab:app:training-data-davis16}
\end{table}

\begin{table}
	\centering
	\caption{Performance of XMem on DAVIS 2017 validation with different training data.}
	\begin{tabular}{lc@{\hspace{5pt}}c@{\hspace{5pt}}c}
		\toprule
		Training data & \mjf & \mj & \mf \\
		\midrule
		DAVIS only & 76.7 & 74.1 & 79.3 \\
		DAVIS+YouTubeVOS only & 84.5 & 81.4 & 87.6 \\
		Static+DAVIS+YouTubeVOS & 86.2 & 82.9 & 89.5 \\
		Static+BL30K+DAVIS+YouTubeVOS & 87.7 & 84.0 & 91.4 \\
		\midrule
		\bottomrule
	\end{tabular}
	\label{tab:app:training-data-davis17}
\end{table}

\begin{table}
	\centering
	\caption{Performance of XMem on DAVIS 2017 test-dev with different training data.}
	\begin{tabular}{lc@{\hspace{5pt}}c@{\hspace{5pt}}c}
		\toprule
		Training data & \mjf & \mj & \mf \\
		\midrule
		DAVIS only & 64.8 & 61.4 & 68.1 \\
		DAVIS+YouTubeVOS only & 79.8 & 76.3 & 83.4 \\
		Static+DAVIS+YouTubeVOS & 81.0 & 77.4 & 84.5 \\
		Static+BL30K+DAVIS+YouTubeVOS & 81.2 & 77.6 & 84.7 \\
		\midrule
		\bottomrule
	\end{tabular}
	\label{tab:app:training-data-davis17td}
\end{table}

\begin{table}
	\centering
	\caption{Performance of XMem on YouTubeVOS 2018 validation with different training data.}
	\begin{tabular}{lc@{\hspace{5pt}}c@{\hspace{5pt}}c@{\hspace{5pt}}c@{\hspace{5pt}}c}
		\toprule
		Training data & \mg & \mjs & \mfs & \mju & \mfu \\
		\midrule
		YouTubeVOS only & 84.4 & 83.7 & 88.5 & 78.2 & 87.2 \\
		DAVIS+YouTubeVOS only & 84.3 & 83.9 & 88.8 & 77.7 & 86.7 \\
		Static+DAVIS+YouTubeVOS & 85.7 & 84.6 & 89.3 & 80.2 & 88.7 \\
		Static+BL30K+DAVIS+YouTubeVOS & 86.1 & 85.1 & 89.8 & 80.3 & 89.2 \\
		\midrule
		\bottomrule
	\end{tabular}
	\label{tab:app:training-data-yv18}
\end{table}

\begin{table}
	\centering
	\caption{Performance of XMem on YouTubeVOS 2019 validation with different training data.}
	\begin{tabular}{lc@{\hspace{5pt}}c@{\hspace{5pt}}c@{\hspace{5pt}}c@{\hspace{5pt}}c}
		\toprule
		Training data & \mg & \mjs & \mfs & \mju & \mfu \\
		\midrule
		YouTubeVOS only & 84.3 & 83.6 & 88.0 & 78.5 & 87.1 \\
		DAVIS+YouTubeVOS only & 84.2 & 83.8 & 88.3 & 78.1 & 86.7 \\
		Static+DAVIS+YouTubeVOS & 85.5 & 84.3 & 88.6 & 80.3 & 88.6 \\
		Static+BL30K+DAVIS+YouTubeVOS & 85.8 & 84.8 & 89.2 & 80.3 & 88.8 \\
		\midrule
		\bottomrule
	\end{tabular}
	\label{tab:app:training-data-yv19}
\end{table}

\section{Multi-scale Evaluation}
\label{sec:app:multiscale}
Multi-scale evaluation is a general trick used in segmentation tasks to boost accuracy by combining results from augmented inputs. Common augmentations include scale-change or vertical mirroring.
Here, we show XMem's results with multi-scale evaluation as an attempt to achieve the best performance with a single model without retraining or using a better backbone.
For these results, we use $P=512$ for a relaxed compression. Vertical mirroring is used. Different augmentations are processed independently and the output probability maps are simply averaged.

On DAVIS, we note that a single large scale (720p) is better than merging multiple smaller scales. 
We use $r=3$ for better results which has also been noted in STCN~\cite{cheng2021stcn}.
In the test-dev set, we additionally include results with $r=5$ (i.e., multi-temporal-scale) in the merge. Table~\ref{tab:app:ms-davis} tabulates our results.

\begin{table}
	\centering
	\caption{XMem with/without multi-scale evaluation on DAVIS. $\ddagger$: 600p evaluation.}
	\begin{tabular}{lccccccccc}
		\toprule
		 & \multicolumn{3}{c}{\hspace{0.5em}DAVIS 2017 val\hspace{0.5em}} & \multicolumn{3}{c}{\hspace{0.5em}DAVIS 2016 val\hspace{0.5em}} & \multicolumn{3}{c}{\hspace{0.5em}DAVIS 2017 test-dev\hspace{0.5em}}\\
		\cmidrule(lr){2-4} \cmidrule(lr){5-7} \cmidrule(lr){8-10}
		Method & \mjf & \mj & \mf & \mjf & \mj & \mf & \mjf & \mj & \mf \\
		\midrule
		XMem & 86.2 & 82.9 & 89.5 & 91.5 & 90.4 & 92.7 & 81.0 & 77.4 & 84.5 \\
		XMem$^{\ast}$ & 87.7 & 84.0 & 91.4 & 92.0 & 90.7 & 93.2 & 81.2 & 77.6 & 84.7 \\
		XMem$^{\ast}$$\ddagger$ & - & - & - & - & - & - & 82.5 & 79.1 & 85.8 \\
		XMem (MS) & 88.2 & 85.4 & 91.0 & 92.7 & 92.0 & 93.5 & 83.1 & 79.7 & 86.4 \\
		XMem$^{\ast}$ (MS) & \textbf{89.5} & \textbf{86.3} & \textbf{92.6} & \textbf{93.3} & \textbf{92.2} & \textbf{94.4} & \textbf{83.7} & \textbf{80.5} & \textbf{87.0} \\
		\midrule
		\bottomrule
	\end{tabular}
	\label{tab:app:ms-davis}
\end{table}

On YouTubeVOS, we adopt multiple scales: $\{480, 528, 576, 624\}$. Unlike on DAVIS, we find larger scales to be unhelpful -- which might be due to the overall less accurate annotation of YouTubeVOS. 
We do not adopt multiple temporal scales here. 
Table~\ref{tab:app:ms-youtube} tabulates our results.

\begin{table}
	\centering
	\caption{XMem with/without multi-scale evaluation on YouTubeVOS.}
	\begin{tabular}{l@{\hspace{7pt}}c@{\hspace{7pt}}c@{\hspace{7pt}}c@{\hspace{7pt}}c@{\hspace{7pt}}c@{\hspace{7pt}}c@{\hspace{7pt}}c@{\hspace{7pt}}c@{\hspace{7pt}}c@{\hspace{7pt}}c}
		\toprule
		& \multicolumn{5}{c}{\hspace{0.5em}YouTubeVOS 2018 val\hspace{0.5em}} & \multicolumn{5}{c}{\hspace{0.5em}YouTubeVOS 2019 val\hspace{0.5em}} \\
		\cmidrule(lr){2-6} \cmidrule(lr){7-11}
		Method & \mg & \mjs & \mfs & \mju & \mfu & \mg & \mjs & \mfs & \mju & \mfu \\
		\midrule
		XMem & 85.7 & 84.6 & 89.3 & 80.2 & 88.7 & 85.5 & 84.3 & 88.6 & 80.3 & 88.6 \\
		XMem$^{\ast}$ & 86.1 & 85.1 & 89.8 & 80.3 & 89.2 & 85.8 & 84.8 & 89.2 & 80.3 & 88.8 \\
		XMem (MS) & 86.7 & 85.3 & 89.9 & \textbf{81.7} & 89.9 & 86.4 & 84.9 & 89.2 & \textbf{81.8} & 89.8 \\
		XMem$^{\ast}$ (MS) & \textbf{86.9} & \textbf{85.6} & \textbf{90.3} & \textbf{81.7} & \textbf{90.2} & \textbf{86.8} & \textbf{85.5} & \textbf{89.8} & \textbf{81.8} & \textbf{89.9} \\
		\midrule
		\bottomrule
	\end{tabular}
	\label{tab:app:ms-youtube}
\end{table}

\newpage
\section{Implementation of the Anisotropic L2 Similarity}
\label{sec:app:similarity}

STCN~\cite{cheng2021stcn} decomposes the L2 similarity into a sequence of tensor operations for a memory- and compute-efficient implementation. For the proposed similarity function to be practical, a similar decomposition is required. Here, we derive and outline our implementation. Recall the definition of the anisotropic L2 similarity:

We are given key $\mathbf{k}\in\mathbb{R}^{\C{k}\times N}$, value $\mathbf{v}\in\mathbb{R}^{\C{v}\times N}$, and query $\mathbf{q}\in\mathbb{R}^{\C{k}\times HW}$.
The key is associated with a \textbf{\underline{s}}hrinkage term $\mathbf{s}\in[1, \infty)^N$ and the query is associated with a s\textbf{\underline{e}}lection term $\mathbf{e}\in[0, 1]^{\C{k}\times HW}$.
Then, the similarity between the $i$-th key element and the $j$-th query element is computed via 
\vspace{-0.2em}
\begin{equation}
	\mathbf{S}(\mathbf{k}, \mathbf{q})_{ij} = -\mathbf{s}_i \sum_{c}^{\C{k}}{ \mathbf{e}_{cj} \left( \mathbf{k}_{ci} - \mathbf{q}_{cj} \right)^2 }, 
	\label{eq:app:similarity}
	\vspace{-0.2em}
\end{equation}
which equates to the original L2 similarity~\cite{cheng2021stcn} if $\mathbf{s}_i=\mathbf{e}_{cj}=1$ for all $i$, $j$, and $c$. 

We use ``$:$'' to denote all the elements in a dimension and ``$@$'' to denote a singleton dimension to be broadcasted.\footnote{Boardcasting as in numpy.} 
$\odot$ denotes the Hadamard (element-wise) product.
$\mathbf{1}$ is an all-ones row vector with length $\C{k}$.
\begin{align}
	\mathbf{S}(\mathbf{k}, \mathbf{q})_{ij} &= -\mathbf{s}_i \sum_{c}^{\C{k}}{ \mathbf{e}_{cj} \left( \mathbf{k}_{ci} - \mathbf{q}_{cj} \right)^2 } \nonumber \\
	&= -\mathbf{s}_i \left( 
	\sum_{c}^{\C{k}}{\mathbf{e}_{cj} \mathbf{k}_{ci}^2} 
	- \sum_{c}^{\C{k}}{2\mathbf{e}_{cj} \mathbf{k}_{ci}\mathbf{q}_{cj}}
	+ \sum_{c}^{\C{k}}{\mathbf{e}_{cj} \mathbf{q}_{cj}^2} 
	\right) \nonumber\\
	&= -\mathbf{s}_i \left( 
	 (\mathbf{k}_{:i} \odot \mathbf{k}_{:i})^T \mathbf{e}_{:j}
	- 2\mathbf{k}_{:i}^T (\mathbf{e}_{:j} \odot \mathbf{q}_{:j})
	+ \mathbf{1} {(\mathbf{e}_{:j} \odot \mathbf{q}_{:j} \odot \mathbf{q}_{:j})} 
	\right) \nonumber\\
	\Rightarrow \mathbf{S}(\mathbf{k}, \mathbf{q}) &= 
	\mathbf{s}_{:@} \left( 
	- (\mathbf{k} \odot \mathbf{k})^T \mathbf{e}
	+ 2\mathbf{k}^T (\mathbf{e} \odot \mathbf{q})
	- \mathbf{1} {(\mathbf{e} \odot \mathbf{q} \odot \mathbf{q})} 
	\right).
\end{align}
This gives a fully vectorized implementation consisting of only element-wise operations and matrix multiplications with broadcasting.


\end{document}